\documentclass{article}

    \usepackage[preprint]{neurips_data_2022}

\usepackage[utf8]{inputenc} 
\usepackage[T1]{fontenc}    
\usepackage{hyperref}       
\usepackage{url}            
\usepackage{booktabs}       
\usepackage{amsfonts}       
\usepackage{nicefrac}       
\usepackage{microtype}      
\usepackage{xcolor}         
\usepackage{graphicx}
\usepackage{multirow}
\usepackage[tbtags]{mathtools}
\usepackage{amsmath}
\usepackage{amssymb}
\usepackage{wrapfig}
\usepackage{gensymb}        
\setcitestyle{numbers}
\setcitestyle{square}

\usepackage{caption}
\usepackage{subcaption}

\usepackage{amsmath,amsfonts,bm}

\def\eqref#1{equation~\ref{#1}}

\def\1{\bm{1}}

\def\vw{{\bm{w}}}
\def\vx{{\bm{x}}}
\def\vy{{\bm{y}}}

\DeclareMathAlphabet{\mathsfit}{\encodingdefault}{\sfdefault}{m}{sl}
\SetMathAlphabet{\mathsfit}{bold}{\encodingdefault}{\sfdefault}{bx}{n}

\title{Benchmark Dataset for Precipitation Forecasting by Post-Processing the Numerical Weather Prediction}

\author{%
  Taehyeon Kim\thanks{Equally contributed.}  ,\quad Namgyu Ho$^*$,\quad Donggyu Kim\thanks{Mainly contributed to our open-source Python code development.},\quad Se-Young Yun \\
  KAIST AI\\
  Seoul, South Korea \\
  \texttt{\{potter32, itsnamgyu, eaststar, yunseyoung\}@kaist.ac.kr} \\
}

\begin{document}

\maketitle

\begin{abstract}
Precipitation forecasting is an important scientific challenge that has wide-reaching impacts on society.
%
Historically, this challenge has been tackled using numerical weather prediction (NWP) models, grounded on physics-based simulations.
Recently, many works have proposed an alternative approach, using end-to-end deep learning (DL) models to replace physics-based NWP models.
While these DL methods show improved performance and computational efficiency, they exhibit limitations in long-term forecasting and lack the explainability. 
%
In this work, we present a hybrid NWP-DL workflow to fill the gap between standalone NWP and DL approaches.
Under this workflow, the outputs of NWP models are fed into a deep neural network, which post-processes the data to yield a refined precipitation forecast.
The deep model is trained with supervision, using Automatic Weather Station\,(AWS) observations as ground-truth labels.
This can achieve the best of both worlds, and can even benefit from future improvements in NWP technology.
To facilitate study in this direction, we present a novel dataset focused on the Korean Peninsula, termed \textbf{KoMet}\,({\underline{Ko}rea \underline{Met}eorological Dataset}), comprised of NWP outputs and AWS observations. For the NWP model, the Global Data Assimilation and Prediction Systems-Korea Integrated Model\,(GDAPS-KIM) is utilized.
%
We provide analysis on a comprehensive set of baseline methods aimed at addressing the challenges of KoMet, including the sparsity of AWS observations and class imbalance.
To lower the barrier to entry and encourage further study, we also provide an extensive open-source Python package for data processing and model development.
Our benchmark data and code are available at \url{https://github.com/osilab-kaist/KoMet-Benchmark-Dataset}.
 \end{abstract}

\vspace{-20pt}
\section{Introduction}
\label{intro}
 
 
\vspace{-10pt}
Precipitation forecasting is an ardous problem of forecasting the specific range of region rainfall based on current observations, from sources including radar\,\cite{ravuri2021skilful, shehu2021relevance}, satellites\,\cite{lebedev2019precipitation}, and rain gauges\,\cite{shehu2021relevance}.
Forecasting plays a vital role in society, aiding in a wide range of weather-dependant decision-making, including large-scale crop management, marine services, and air traffic control\,\cite{wilson2010nowcasting, book, nurmi2013expected, hoogenboom2000contribution}.
A {Numerical Weather Prediction\,(NWP)} model\,\cite{sun2014use} is a general tool for weather forecasting, which involves calculating complex physical processes pertaining to interactions across large time and space scales, spanning the Earth's atmosphere, ocean, land, and ice\,\cite{navon2009data, kalnay1990global, bauer2015quiet}.
Despite continuous efforts to enhance NWP models \cite{shin2022overview}, erroneous predictions can occur due to the delicacy of the models with respect to the noise in the initially observed state.

The weather and climate research community is becoming more aware of contemporary deep learning (DL) technologies, and numerous attempts have been made to apply them to the task of precipitation forecasting.
Most DL-based methods directly utilize radar observations to predict future precipitation. Despite the lack of expert knowledge or explicit physics-based modeling, these methods achieve notable performance while being computationally efficient\,\cite{shi2017deep, ayzel2020rainnet}.
Ravuri et al.\,\cite{ravuri2021skilful} facilitate the performance of observation-driven methods with a deep generative model that learns probability distributions of the data and allow for easy generation of samples from their learned distributions.
Espeholt et al.\,\cite{espeholt2021skillful} extend the capability of precipitation forecasting up to twelve hours ahead, surpassing the performance of existing physics-based models widely used in the Continental United States. 
Despite its significance for instantaneous forecasting, it is shown that pure observation-driven DL models underperform those that are supplemented by NWP outputs when lead time is larger than 6 hours, revealing the limitations of DL in long-term forecasting.
Furthermore, it still lacks \textit{explainability} and explicit consideration for \textit{physics constraints} in the end-to-end DL framework\,\cite{schultz2021can}.

This paper presents a novel benchmark dataset for a hybrid workflow that combines the strengths of both NWP and DL approaches.
Our new benchmark, termed \textbf{KoMet}\,(\textit{\underline{Ko}rea \underline{Met}eorological dataset}), consists of two parts: predictions from an NWP model, specifically, the Global Data Assimilation and Prediction Systems-Korea Integrated Model\,(GDAPS-KIM), and Automatic Weather Station\,(AWS) observations, \textcolor{black}{whose rainfall amounts are hourly monitored in areas covering 13km around the center.}
The data is provided by the Korean Meteorological Administration\,(KMA), for July and August of 2020 and 2021--two months in which most precipitation in the Korean Peninsula occurs.
Our dataset takes into account of the characteristics of the Korean Peninsula whose precipitation is \textcolor{black}{difficult to predict due to} congested spatio-temporal variability owing to complex terrain and the Asian monsoon season\,\cite{qian2002distribution, chen2004variation}.
\textcolor{black}{For example, its complex topography leads to large variation in annual precipitation in the southern part\,(1000 mm to 1800 mm) than the central part\,(1100 mm to 1400 mm)\,\cite{azam2018spatial}.}
Under the proposed workflow, GDAPS-KIM predictions are used as inputs to a DL model which is trained to output refined precipitation forecasts. AWS observations are used as ground-truth targets to train the deep model.
In nutshell, the overall goal is to post-process the predictions from GDAPS-KIM using deep models, under the supervision of AWS observations.

To date, many benchmark datasets have been introduced to facilitate existing approaches in precipitation forecasting–those based on NWP and end-to-end DL\,\cite{https://doi.org/10.1029/2018MS001597, rasp2020weatherbench}.
We advocate an alternative approach, in which NWP output is post-processed using DL models for a more refined prediction. This enables the prediction to respect physics-related constraints and is partially explainable by the injected NWP predictions. To catalyze future work, we also provide extensive evaluation of several baseline models and learning methods over our KoMet dataset. Our key contributions include:
\vspace{-7pt}
\begin{itemize}
\item \textbf{KoMet} is the most comprehensive public dataset \textcolor{black}{for DL post-processing of NWP outputs. The spatial size of GDAPS-KIM is 50 x 65 covering the terrain and surrounding sea whose each pixel covers 12km$^2$ and 484 AWS\,(ground-truth pixels) are installed in such resolution.}

\item \textbf{Hybrid NWP-DL Baselines}. We present and analyze a comprehensive set of baseline methods for our proposed hybrid approach, and provide a suite of Python modules designed to facilitate further research from the ML community.

\end{itemize}

\begin{figure}[t]
\begin{center}
   \includegraphics[width=1.0\linewidth]{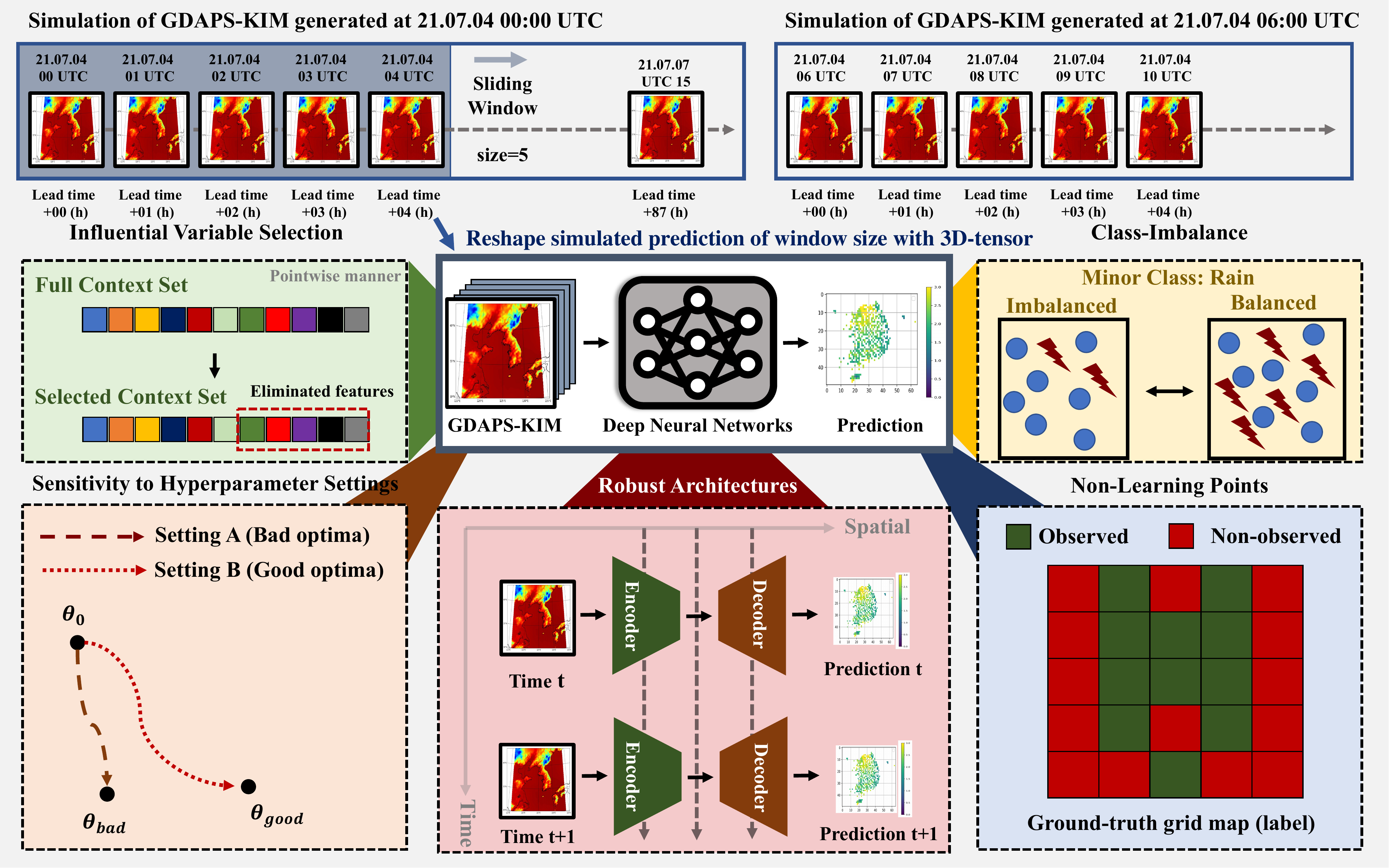}
\end{center}
\caption{\textbf{Overview of challenges for KoMet}. The central part shows our main workflow; as GDAPS-KIM simulations are released timely, a deep neural network aims to post-process the model output for rain forecast. Challenges can be categorized into five folds: (1) influential variable selection that selects the key features for data pre-processing, (2) sensitivity to hyperparameter settings, (3) robust architectures, (4) non-learning points on the ground-truth labels, and (5) class-imbalance issues where the number of precipitation areas are fewer than the number of other cases.}
\label{overview}
\vspace{-10pt}
\end{figure}

\vspace{-10pt}
\paragraph{Post-NWP Optimization.} We refer to the optimization problem implicit in the post-processing of NWP model ouput for precipitation forecasting as {post-NWP optimization}, drawing a connection to skillful precipitation optimization. Post-NWP optimization has several key properties that differentiates it from a typical weather-forecast optimization problem\,(\autoref{overview}):

\vspace{-7pt}
\begin{itemize}
    \item \textbf{Influential Variable Selection.} In NWP, each pixel on grid has a list of variables expressing the state regarding atmospheric feature\,(\autoref{tab:variable}).
    
    \item \textbf{Sensitivity to Hyperparameter Settings.} NWP data can be regarded as tabular data. In this sense, as Kadra et al.\,\cite{kadra2021well} observe, finding the optimal cocktail with tuning the modern regularization techniques can boost the performance comparable to the state-of-the-art models in end-to-end DL models.
    
    \item \textbf{Robust Architectures. } We expect a new dedicated architecture by considering the spatial and temporal characteristics of our benchmark.
    
    \item \textbf{Non-Learning Points of AWS. } AWS is not installed at all locations due to space and cost limitations. Therefore, its value for some regions are sparse, which needs to be overcome for the supervision of NWP.
    
    \item \textbf{Class-Imbalance. } The distribution of precipitation amounts are highly asymmetric. Despite rare occurrences of heavier rain, it can bring significant real-world damages. 
\end{itemize}


\vspace{-10pt}



\paragraph{Organization.} The remainder of this paper is organized as follows. In Section\,\ref{sec2}, we discuss the related literature on traditional NWP approaches, data assimilation, and deep learning-based weather forecasting. In Section\,\ref{sec3}, we describe our benchmark \textbf{KoMet} thoroughly. In Section\,\ref{sec4}, we formulate the problem settings for the use of KoMet, and Section\,\ref{sec5} exhibits the baseline experimental results. Finally, Section\,\ref{sec6} concludes the paper.

\vspace{-10pt}
\section{Related Works}
\label{sec2}
\vspace{-10pt}
\paragraph{Numerical Weather Prediction\,(NWP).}
NWP models explain and predict the future changes of the atmospheric conditions through solving dynamics and physics partial differential equations. Since the advent of the NWP models, weather forecasting highly relies on them according to their steady progress over time\,\cite{lynch2008origins,bauer2015quiet}. NWP applications have been involved into air quality\,\cite{zhang2012real}, solar\,\cite{perez2013comparison,zhang2022solar,verbois2022statistical}, wind power\,\cite{de2011assessment,chen2013wind}, wildfire\,\cite{di2016potential}, anomalies (i.e. severe weather events)\,\cite{donaldson1975objective,mills1998objective,xue2003advanced,rodwell2007using}, and hydrological forecasting. The last of these, hydrological forecasting, includes precipitation, humidity\,\cite{drusch2007initializing}, drought\,\cite{palmer19891988,vstvepanek2018drought}, and  evapotranspiration\,\cite{medina2020comparison} forecasting. 
Despite its raving utility in real world, NWP faces some endemic challenges\,\cite{sun2014use,bauer2015quiet}. Firstly, the quality of NWP models can be determined with how accurate the given observations of present atmospheric states\,(initial condition) is\,\cite{kalnay2003atmospheric,slingo2011uncertainty,wahl2015uncertainty,olafsson2020uncertainties}. In addition, the computational and power demands of NWP grow in lockstep with the resolution of the forecast, creating a trade-off between forecast accuracy, which necessitates higher levels of resolution, and forecasting time. Lastly, some models utilize the machine learning approach for prediction while there remains a lack of considering two types of uncertainty: \textit{aleatoric} uncertainty and \textit{epistemic} uncertainty\,\cite{der2009aleatory}. 




\vspace{-5pt}
\paragraph{Data Assimilation\,(DA).}
DA is the process of combining every possible source of information for the accurate prediction of each status. 
Generally, DA serves as an initial condition for computing the NWP in weather forecasting\,\cite{kalnay2003atmospheric,navon2009data}.
Various observation platforms have been evolving through weather station, radar, and satellite\,\cite{xue2003advanced,sun2005convective,drusch2007initializing,boussetta2015assimilation,buehner2020non} whose observation equipment is also becoming more sophisticated. To reinforce model uncertainty, variational data assimilation methods have been emerged as a line of works\,\cite{talagrand1987variational,parrish1992national}.
Data assimilation methods\,\cite{talagrand1987variational,parrish1992national} have been done on such observations. Three-dimensional variational methods\,(3DVar)\,\cite{courtier1998ecmwf,gauthier1999implementation,lorenc2000met}, four dimentional approaches\,(4DVar)\,\cite{rabier2000ecmwf}, ensemble Kalman filter\,\cite{hamill2000hybrid,lorenc2003potential} are general examples of DA techniques used in global NWP workflow.

\vspace{-5pt}
\paragraph{Deep Learning-based Weather Forecasting.}

Deep learning rises as the one way to mitigate the shortcomings of NWP.
DL enables its success by conjugating vast amount of available data as well as geometrically increased computational power of tensor processing units\,(e.g. GPUs or TPUs)\,\cite{jouppi2017datacenter}. 
Because deluge of meteorological data has been accumulated since the establishment of weather stations, there are an increasing demand of efforts to adopt modern DL techniques for weather forecasting\,\cite{shi2017deep,schultz2021can,ren2021deep,rolnick2022tackling}. DL-based weather forecast approach can be categorized into two folds: end-to-end DL approach and NWP-DL hybrid approach. End-to-end DL approaches\,\cite{liu2016application,rasp2018deep,rodrigues2018deepdownscale,weyn2019can,zhang2021rn,ravuri2021skilful} learn the representation from meteorological observations using DNN without the access of physics knowledge. 
On the other side, NWP-DL hybrid approaches\,\cite{rasp2018neural,weyn2019can,gronquist2019predicting,taillardat2020research,gronquist2021deep,haupt2021towards,gronquist2021deep} integrate theoretical models\,(NWP) and data-driven models\,(DL), and thus encompass the explainability as well as accurate expressivity. 



\begin{table}[t]
\centering\footnotesize
\caption{List of variables contained in the benchmark dataset. Pres data contains atmospheric features at different pressure levels, and Unis data contains some other features as well as those in Pres data but does not at certain pressure levels. $^\dagger$ indicates the variable with integer type. }
\addtolength{\tabcolsep}{-3pt}
\begin{tabular}{@{}llclc@{}}
\toprule
Type          & Long name                              & Short name & Description                                    & Unit          \\ \midrule
\multirow{5}{*}{Pres}  & U-component of wind           & u    & Wind in x/longitude-direction               & ($ms^{-1}$) \\
                       & V-component of wind           & v    & Wind in y/latitude direction          & ($ms^{-1}$) \\
                       & Temperature                   & T          & Temperature                                    & (K)           \\
                       & Relative humidity             & rh liq     & Humidity relative to saturation                & (\%)          \\
                       & Geopotential                  & hgt        & Proportional to the height of a pressure level & ($m^2s^{-2}$) \\\midrule
\multirow{12}{*}{Unis} & Rain                                   & rain       & Rain                                           & ($mm/h$)\\
                       & Snow                                   & snow       & Snow                                           & ($cm/h$)\\
                       & Height of PBL                          & hpbl      & Height of planetary boundary layer             & (km)               \\
                       & Type of PBL$^\dagger$                         & pbltype    & Type of planetary boundary layer               &   - \\
                       & 2m specfic humidity                  & q2m        & Specific humidity at 2m height                 & (g/kg) \\
                       & 2m relative humidity                 & rh2m       & Humidity relative to saturation at 2m height   & (\%)          \\
                       & 2m temperature                        & t2m        & Temperature at 2m height above surface         & (K)           \\
                       & Surface temperature                   & tsfc       & Surface temperature                            & (K)           \\
                      & 10m u component of wind & u10m & Wind in x/longitude-direction at 10m height & ($ms^{-1}$) \\
                      & 10m v component of wind & v10m & Wind in y/latitude-direction at 10m height  & ($ms^{-1}$) \\
                       & Topography                             & topo       & Topography                                     &   (m)           \\
                       & Pressure of surface                    & ps         & Pressure of surface                            & (Pa)               \\
\bottomrule                       
\end{tabular}
\label{tab:variable}
\vspace{-15pt}
\end{table}

\vspace{-10pt}
\section{KoMet Dataset}\label{sec3}
\vspace{-8pt}
\textcolor{black}{The KoMet dataset is comprised of predictions from GDAPS-KIM, a global numerical weather prediction model operated by the Korea Meterological Administration\,(KMA\footnote{\url{https://www.kma.go.kr/eng/index.jsp}}), as well as AWS observations which serve as ground-truth precipitation data.}
\vspace{-10pt}
\subsection{\textcolor{black}{Input: GDAPS-KIM}}
\vspace{-5pt}
\textcolor{black}{GDAPS-KIM is a global numerical weather prediction model designed to improve the prediction accuracy of weather phenomena with a particular focus on the Korean Peninsula\,\cite{shin2022overview}. GDAPS-KIM provides hourly predictions of various atmospheric variables. This is achieved by assimilating observations from all over the globe to compute the initial state, and subsequently solving dynamics and physics equations to predict future states.}

\textcolor{black}{GDAPS-KIM prediction data is provided in the following format: $\mathbf{X} \in \mathbb{R}^{T \times L \times C \times W \times H}$ where each dimension corresponds to the origin time\,(T) at which the simulation took place, lead time\,(L) between the origin time and the target prediction time, the variable index\,(C), and the spatial dimensions of the prediction map\,(W, H). Each GDAPS-KIM instance in KoMet dataset is a daily simulation executed at 00 UTC. We provide predictions made between July 1st and August 31st (inclusive) of 2020 and 2021, i.e., $T=124$ days, when rainfall is most intensive due to the seasonal characteristics of the region. We provide predictions with lead times ranging from 0 to 89 hours, i.e., $L=90$. We provide data on $C=122$ atmospheric variables for the Korean Peninsula and the surrounding region lying within [32.94\degree N, 39.06\degree N] and [124.00\degree E, 132.00\degree E]. GDAPS-KIM has a resolution of 12 km $\times$ 12 km, and thus its spatial dimension is 65$\times$50.}

\textcolor{black}{GDAPS-KIM variables can be categorized into two types: Pres variables and Unis variables. While Pres variables include values for various altitudes corresponding to specific air pressure levels\,(i.e., different isobaric surfaces), Unis data does not.
Each GDAPS-KIM instance has 5 Pres variable types at 22 isobaric surfaces\,(22$\times$5=110 variables) and 12 Unis variables (\autoref{tab:variable}), thus it has 122 variables at each pixel.
Despite its rich variables, using all variables may not always be a silver bullet owing to the high correlation among themselves and noisiness.
Furthermore, as the assimilated initial condition used for numerical prediction has significant spatial- and temporal-variability, it is more prone to inaccuracies compared to other regions, leading to cascading errors.
In this paper, for the sake of benchmark experiments, we use 12 variables\footnote{Unis 6 variables: rain, q2m, rh2m, t2m, tsfc, ps; Pres 6 variables: T and rh\_liq  at 500, 700, 850 hpa.} out of 122 through the advice of KMA experts.}



\vspace{-10pt}
\subsection{\textcolor{black}{Ground-Truth: AWS Observations}}
\vspace{-5pt}

\textcolor{black}{Surface observations from AWS provide ground-truth data on hourly precipitation. While AWS observation data includes various weather measurements such as temperature, wind speed, and solar radiation, the precipitation is only utilized for supervision. Unlike GDAPS-KIM predictions,
only several pixels have AWS observations as weather stations are sparsely installed.
Even for such stations, observations are occasionally omitted due to miscellaneous reasons, leading to NaN values.}

\textcolor{black}{Raw AWS observation data is provided in the following format: $\mathbf{Y}_{raw} \in \mathbb{R}^{h \times S}$ where each dimension corresponds to the hour\,(h) at which the measurement of accumulated rainfall (from the preceding hour) is taken and the weather station index\,(S). AWS observations are recorded every hours at which GDAPS-KIM predictions are provided, with some additional leeway. More precisely, hourly observations are ranging from July 1st, 00:00 UTC to September 3rd, 23:00 UTC, totaling $h=3120$ hours, from 484 stations\,($S=484$). To facilitate use of AWS data to train GDAPS-KIM post-processing models, we use the location metadata of each weather station to convert AWS data into the spatial grid format used in GDAPS-KIM. The format is $\mathbf{Y} \in \mathbb{R}^{h \times W \times H}$ where each dimension corresponds to the hour\,(h) at which the measurement of accumulated rainfall is taken, and the spatial dimensions of the prediction map (W, H; identical to GDAPS-KIM).}

\begin{figure}[t]
\begin{center}
   \begin{subfigure}[b]{0.45\textwidth}
         \centering
         \includegraphics[width=\textwidth]{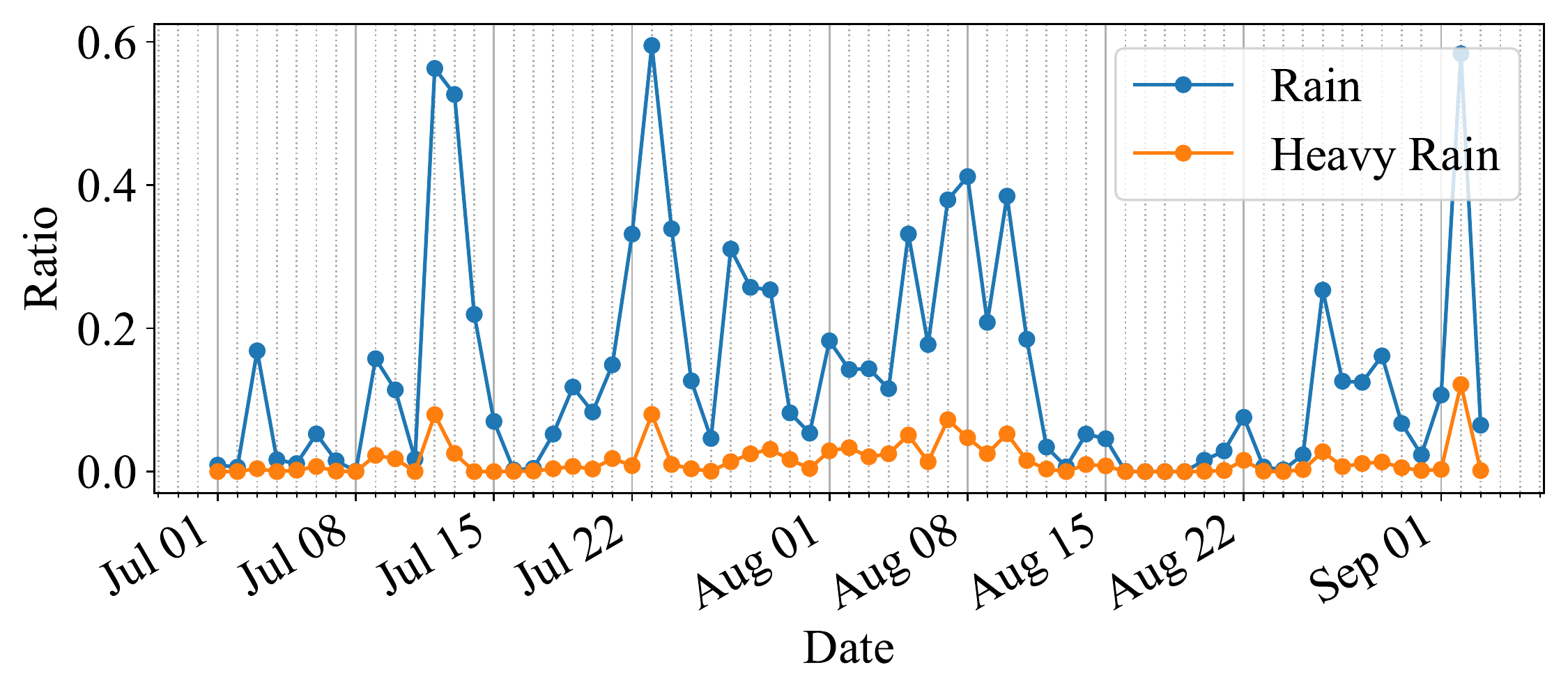}
         \vspace{-20pt}
         \caption{Precipitation Ratio by Date (2020)}
         \label{fig:precipitation_ratio_by_date_2020}
     \end{subfigure}
   \begin{subfigure}[b]{0.45\textwidth}
         \centering
         \includegraphics[width=\textwidth]{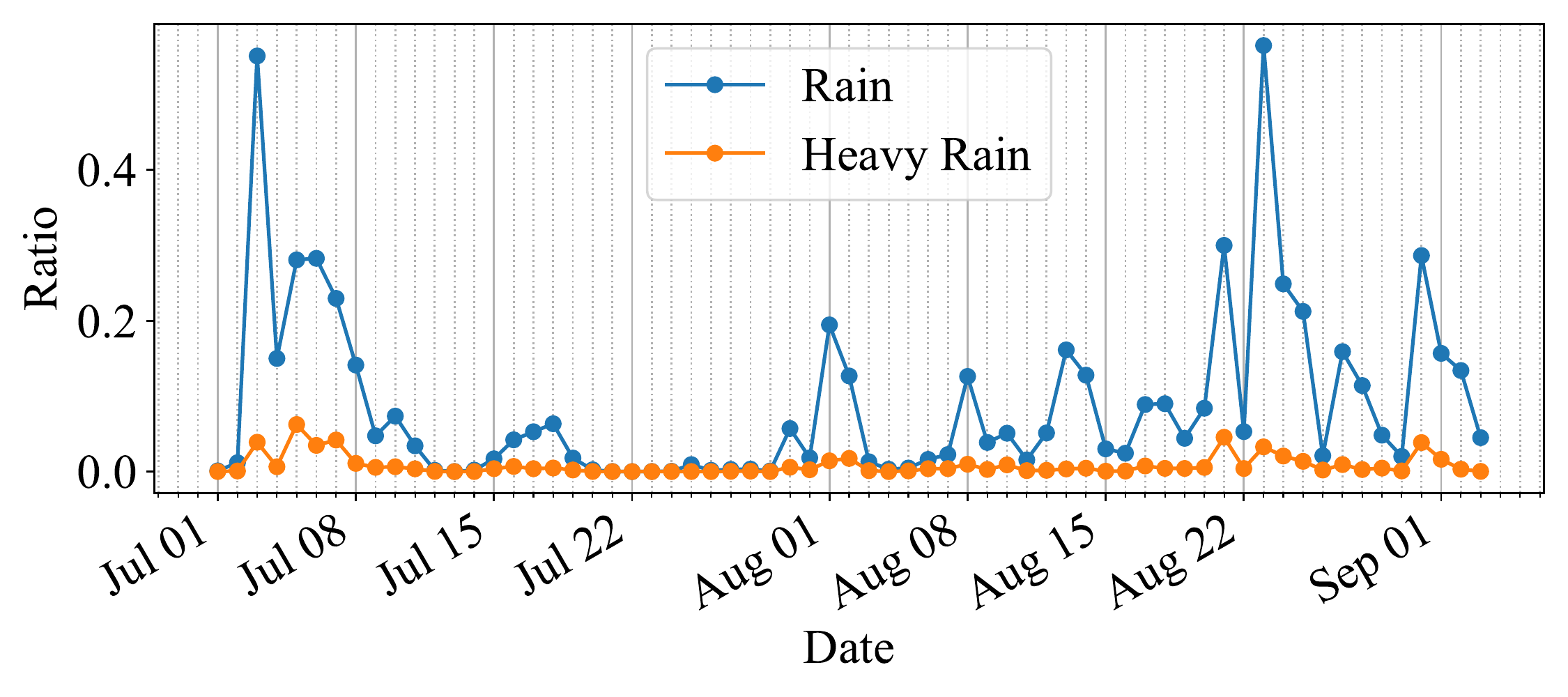}
         \vspace{-20pt}
         \caption{Precipitation Ratio by Date (2021)}
         \label{fig:precipitation_ratio_by_date_2021}
     \end{subfigure}
   \begin{subfigure}[b]{0.45\textwidth}
         \centering
         \includegraphics[width=\textwidth]{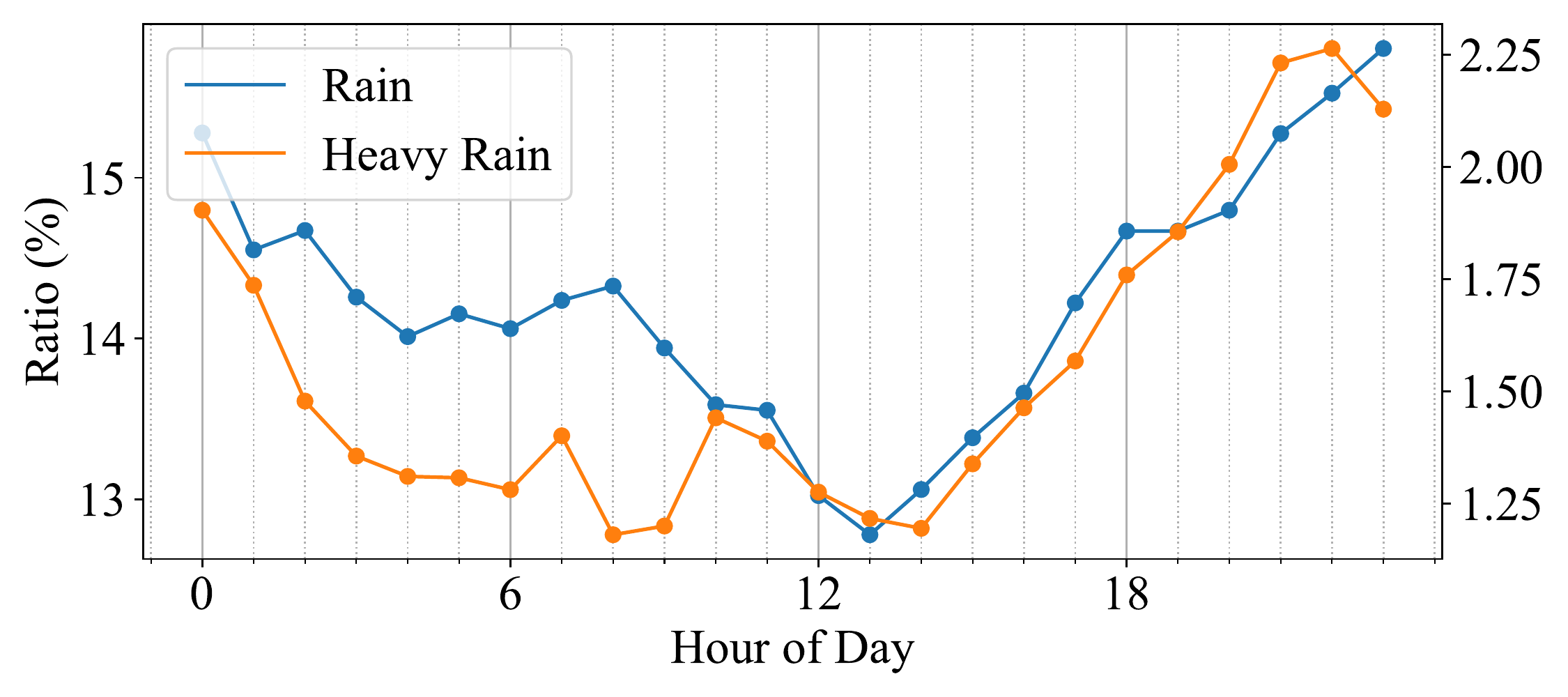}
         \vspace{-20pt}
         \caption{Precipitation Ratio by Hour (2020)}
         \label{fig:precipitation_ratio_by_hour_2020}
     \end{subfigure}
   \begin{subfigure}[b]{0.45\textwidth}
         \centering
         \includegraphics[width=\textwidth]{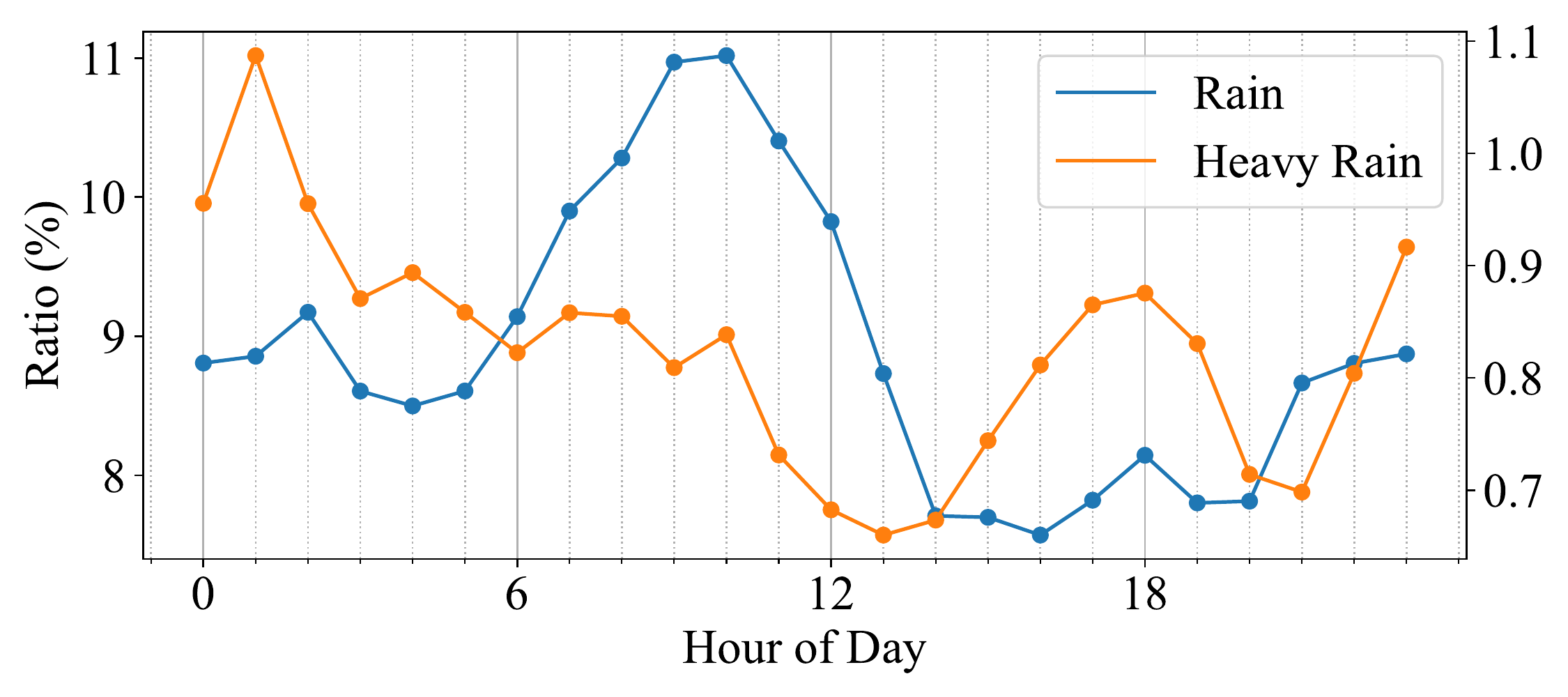}
         \vspace{-20pt}
         \caption{Precipitation Ratio by Hour (2021)}
         \label{fig:precipitation_ratio_by_hour_2021}
     \end{subfigure}
\end{center}
\vspace{-5pt}
\caption{Temporal distribution of \textcolor{black}{the rain ratio among all AWS observations}. `Rain' refers to hourly precipitation between 0.1mm and 10mm, while `heavy rain' refers to that above 10mm. \textcolor{black}{In (c), (d), the left axis refers to the ratio of `rain' and the right axis is the ratio of `heavy rain', in percentages.}}
\label{temperal_sampling}
\vspace{-15pt}
\end{figure}
\begin{figure}[t]
\begin{center}
   \begin{subfigure}[b]{0.3\textwidth}
         \centering
         \includegraphics[width=\textwidth]{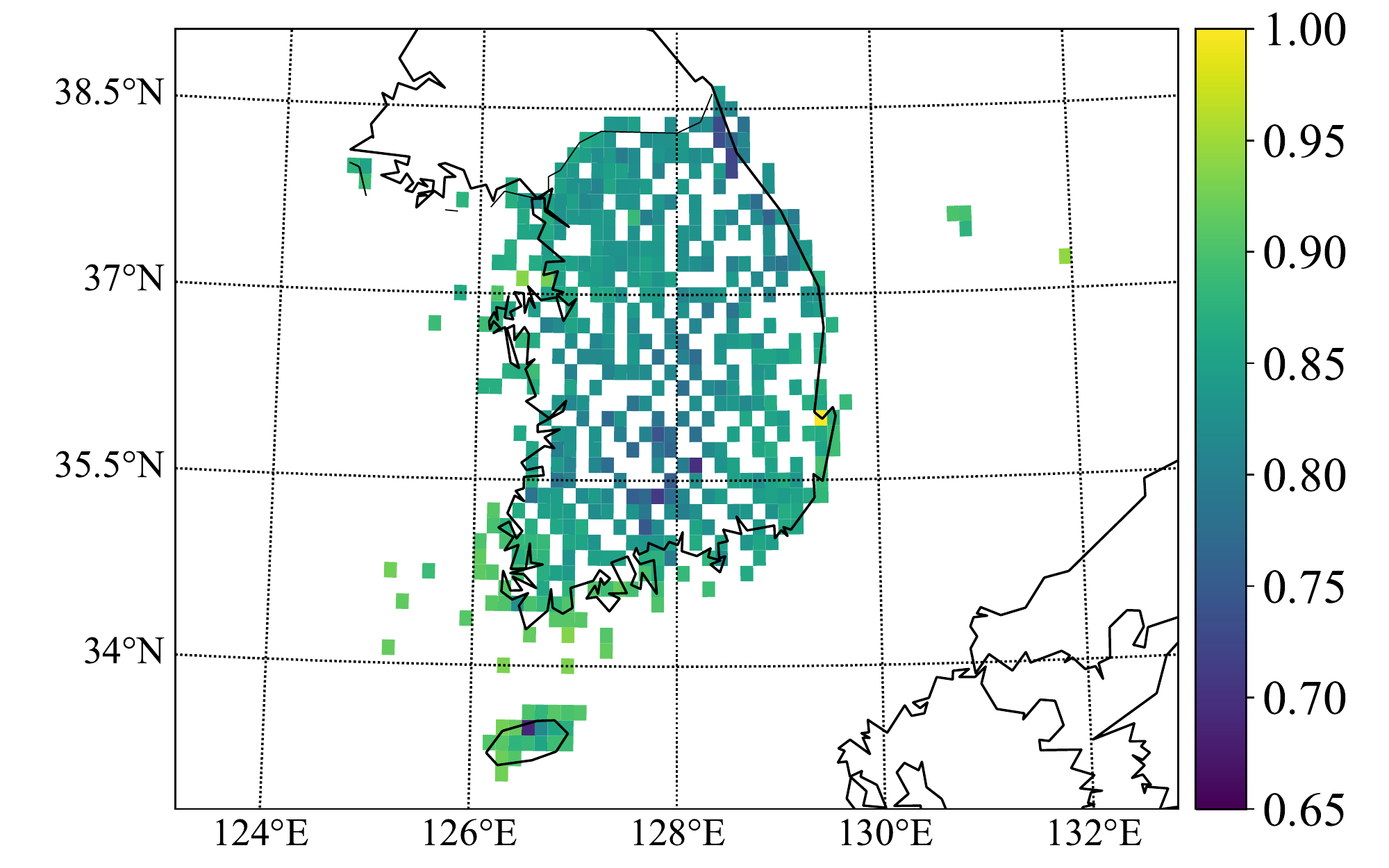}
         \caption{No Rain Ratio (2020)}
         \label{fig2:spatial_no_rain_ratio_2020}
     \end{subfigure}
   \begin{subfigure}[b]{0.3\textwidth}
         \centering
         \includegraphics[width=\textwidth]{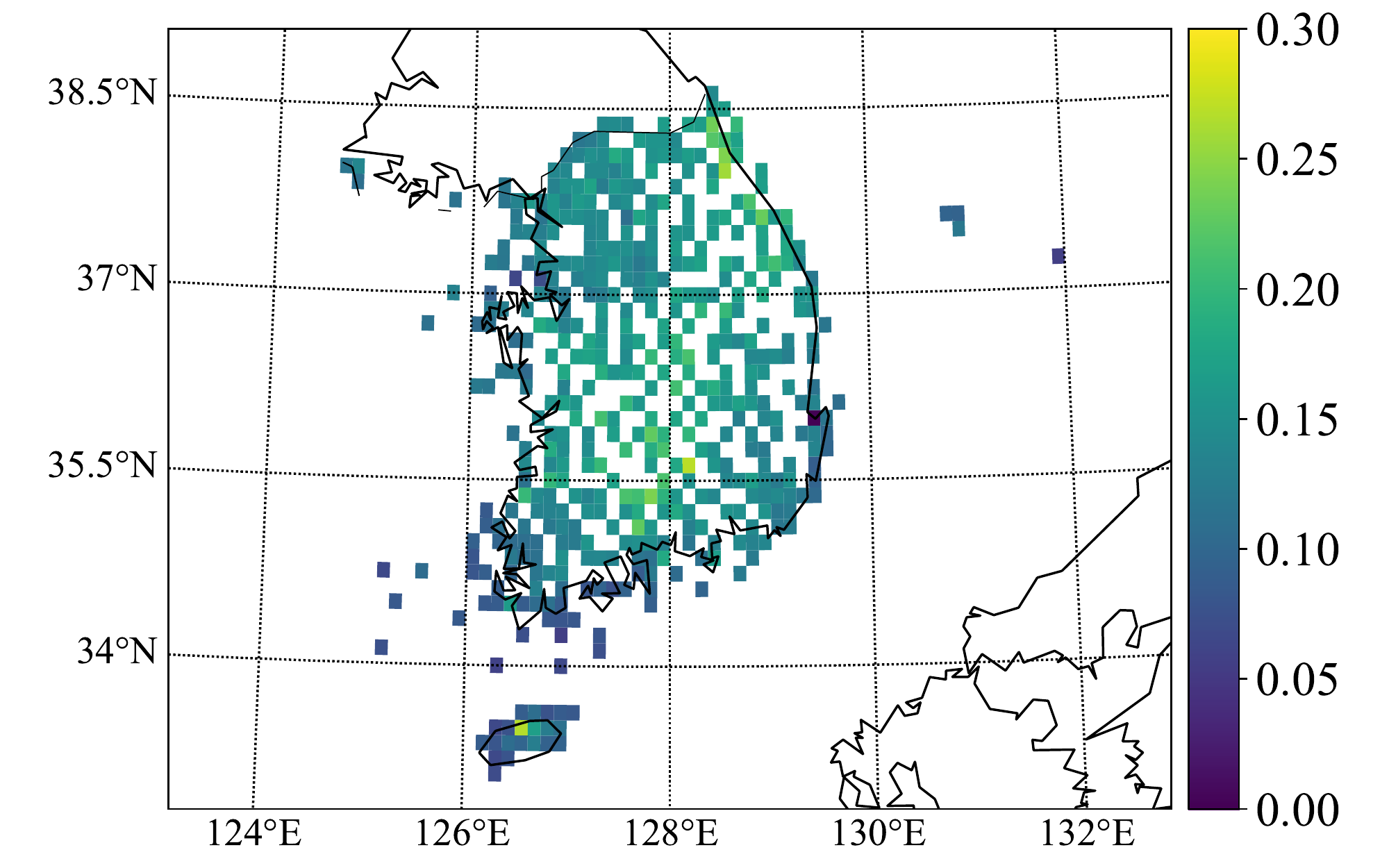}
         \caption{Rain Ratio (2020)}
         \label{fig2:spatial_rain_ratio_2020}
     \end{subfigure}
   \begin{subfigure}[b]{0.3\textwidth}
         \centering
         \includegraphics[width=\textwidth]{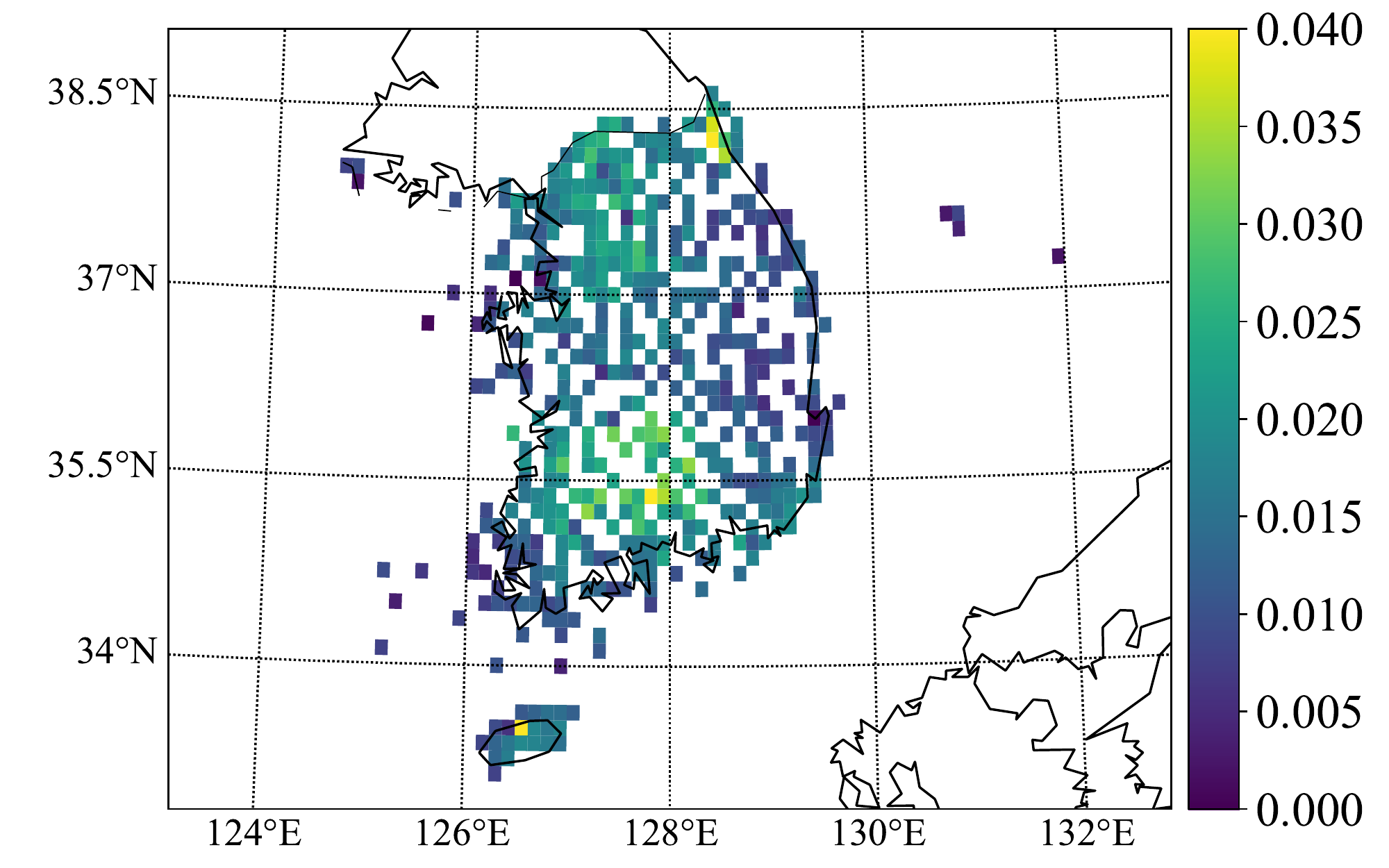}
         \caption{Heavy Rain Ratio (2020)}
         \label{fig2:spatial_heavy_rain_ratio_2020}
     \end{subfigure}
     
   \begin{subfigure}[b]{0.3\textwidth}
         \centering
         \includegraphics[width=\textwidth]{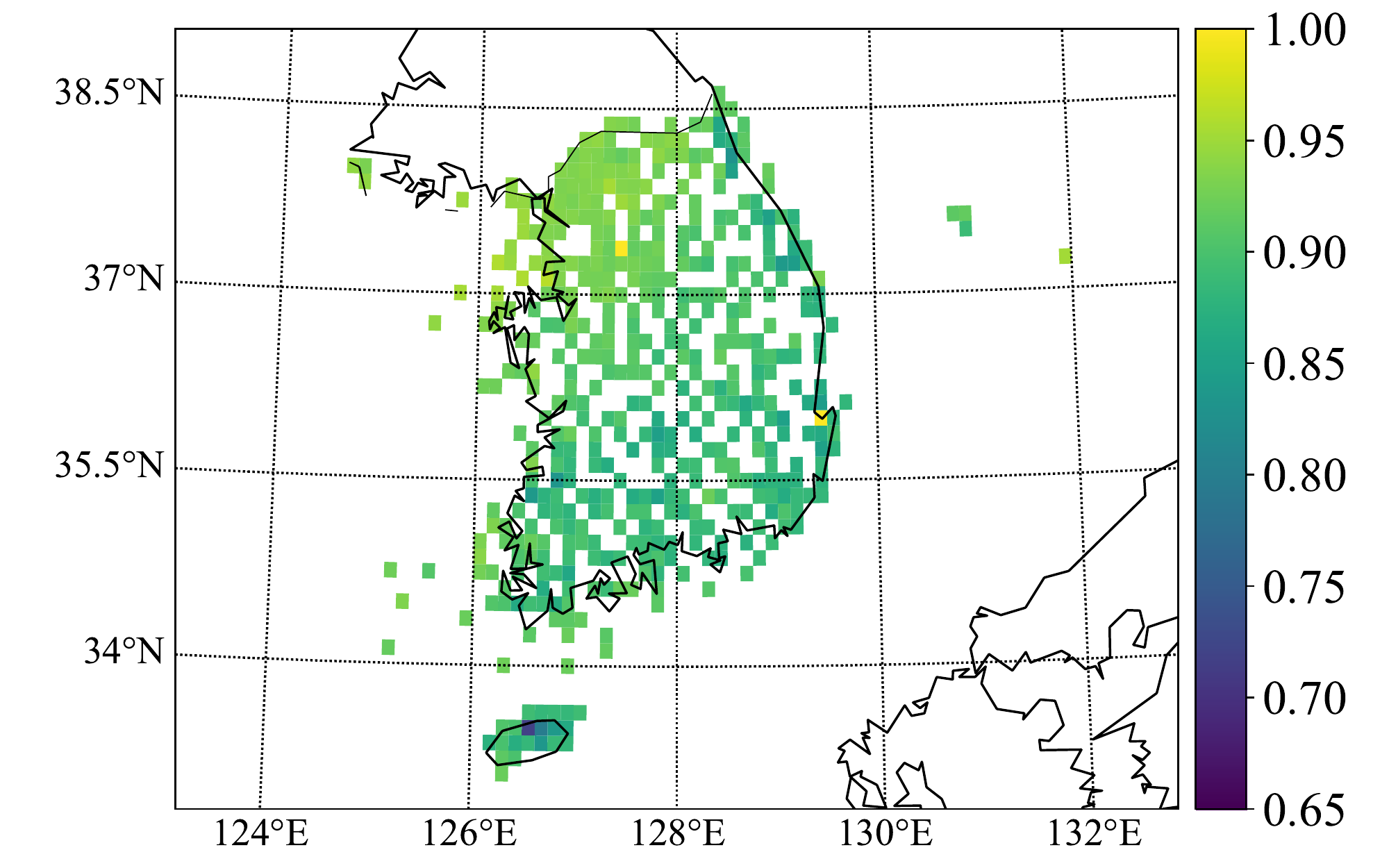}
         \caption{No Rain Ratio (2021)}
         \label{fig2:spatial_no_rain_ratio_2021}
     \end{subfigure}
   \begin{subfigure}[b]{0.3\textwidth}
         \centering
         \includegraphics[width=\textwidth]{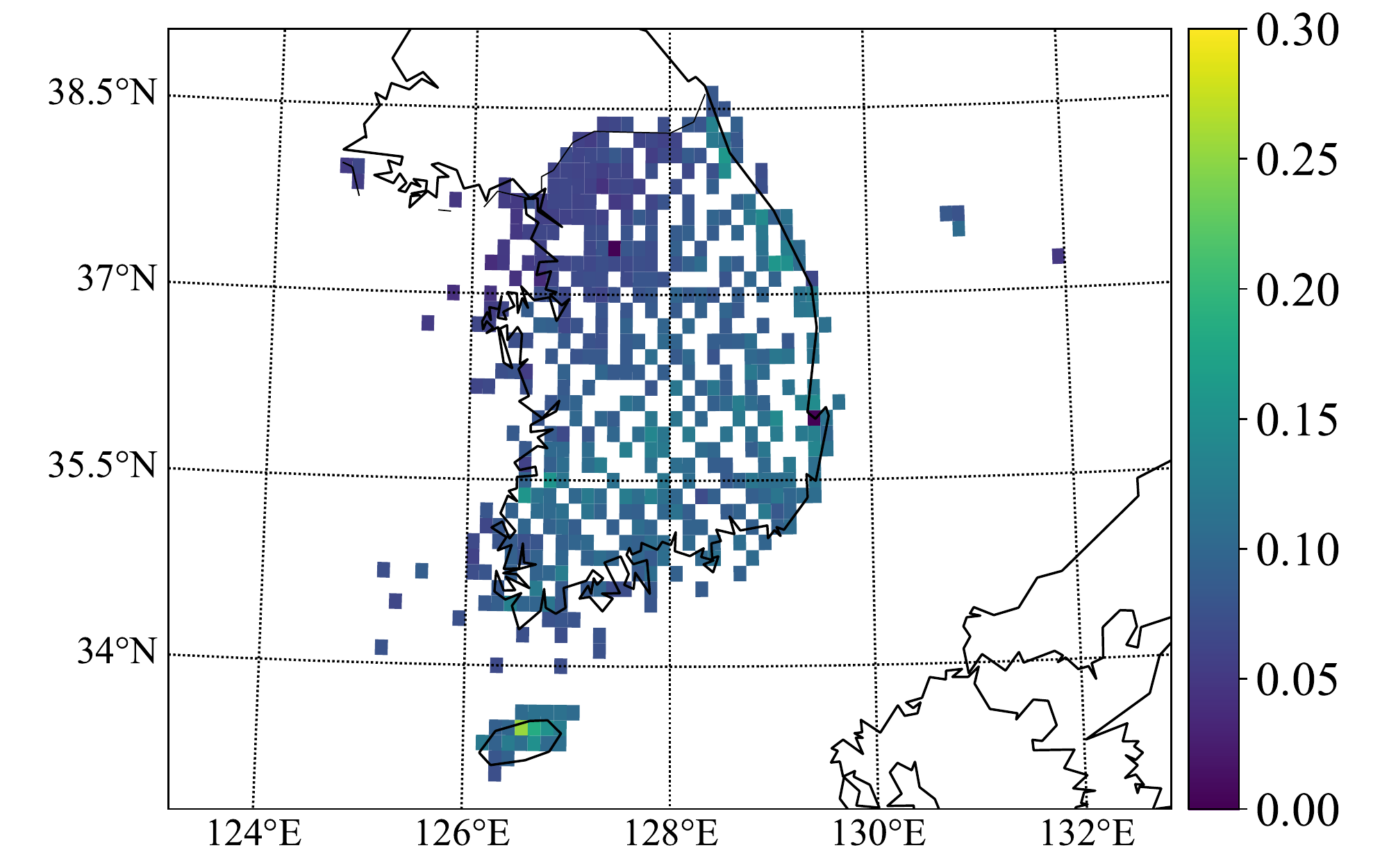}
         \caption{Rain Ratio (2021)}
         \label{fig2:spatial_rain_ratio_2021}
     \end{subfigure}
   \begin{subfigure}[b]{0.3\textwidth}
         \centering
         \includegraphics[width=\textwidth]{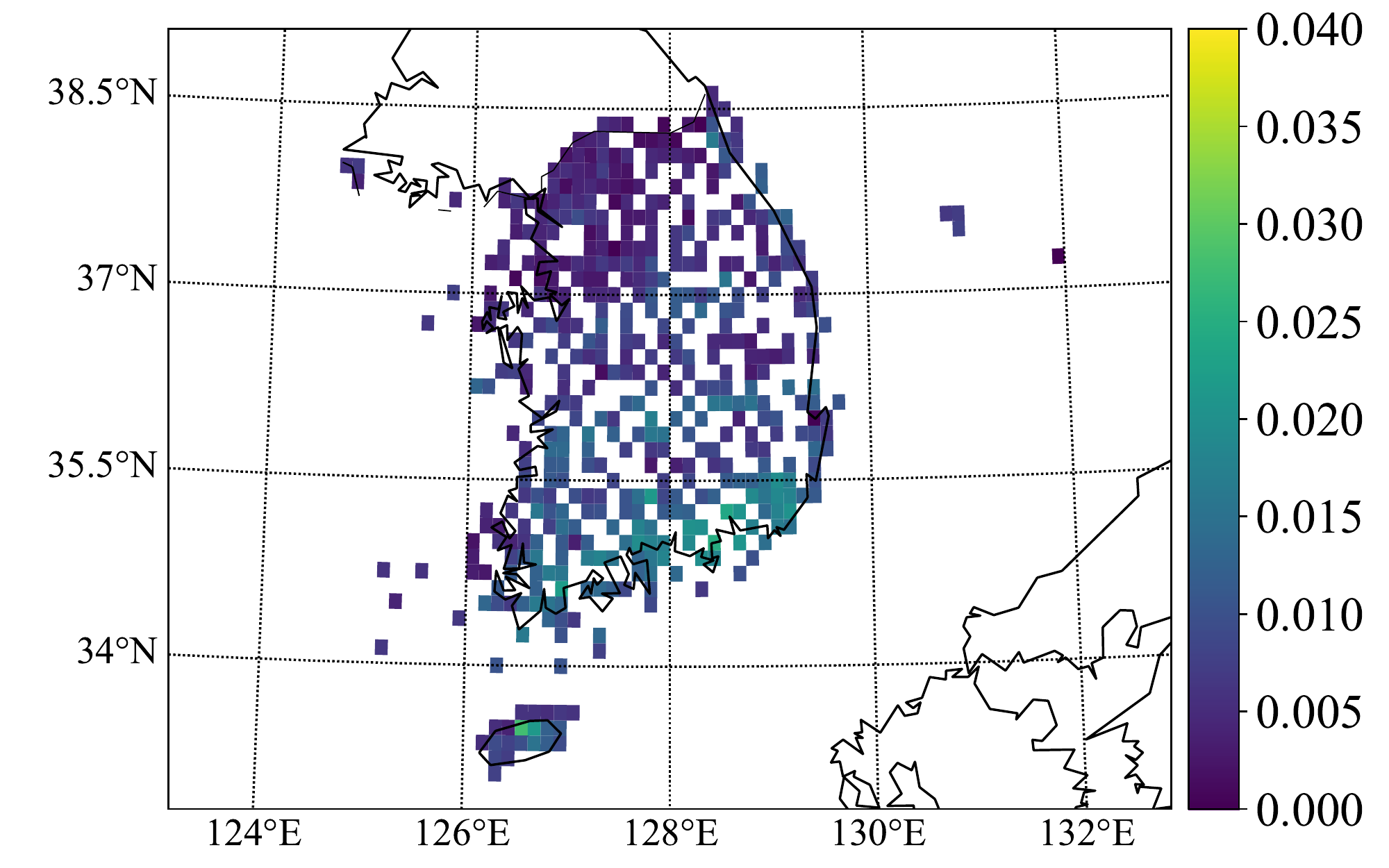}
         \caption{Heavy Rain Ratio (2021)}
         \label{fig2:spatial_heavy_rain_ratio_2021}
     \end{subfigure}
\end{center}
\vspace{-5pt}
\caption{Spatial distribution of precipitation from AWS observations in South Korea. `Rain' refers to hourly precipitation between 0.1mm and 10mm, while `heavy rain' refers to that above 10mm.}
\vspace{-20pt}
\label{spatial_sampling}
\end{figure}

\textcolor{black}{We investigate the characteristics of AWS  considering the sparsity of AWS installation, temporal and spatial rainfall distributions. \autoref{temperal_sampling} provides the distributions of hourly and daily average of rainfall amounts on each year. As \autoref{temperal_sampling} shows, hourly rainfall events occur sporadically, and these characteristics vary greatly from year to year. For the spatial distribution, among the $65 \times 50 = 3250$ pixels comprising the GDAPS-KIM grid, there are only 484 pixels where ground-truth data is available, and even many of those are concentrated to certain locations such as metropolitan areas\,(\autoref{spatial_sampling}).  These challenges can pose a significant challenge for training calibration models, as the supervisory signal from ground-truth values is limited to specific locations and time.}




\vspace{-10pt}
\subsection{Dataset Interface}
\vspace{-5pt}


\begin{wraptable}{r}{6.5cm}
\vspace{-25pt} 
\centering \footnotesize
\addtolength{\tabcolsep}{-3.5pt}
\caption{Statistics in the KoMet benchmark.\label{tab:rain_fall_stat}}%
\begin{tabular}{@{}ccc@{}}
\toprule
Rain rate (mm/h) & Proportion (\%) & Rainfall Level \\ \midrule
$[0.0, 0.1)$& 87.24 & No Rain \\
$[0.1, 10.0)$& 11.57 & Rain \\
$[10.0, \infty)$& 1.19 & Heavy Rain \\ \bottomrule
\end{tabular}
\vspace{-10pt}
\end{wraptable}

\paragraph{Inputs.}

In our dataset, GDAPS-KIM predictions are provided in NumPy format. These predictions are propagated to deep models, following normalization; \textcolor{black}{each variable is linearly scaled based on min-max values derived from the entire dataset}.

\vspace{-5pt}
\paragraph{Outputs.}

We formulate the precipitation forecast problem as a pointwise classification task pertaining to three categorical classes: non-rain, rain, and heavy rain\,(\autoref{tab:rain_fall_stat}). 
Following this, the AWS observation data is pre-processed into 2D array format according to the grid used in GDAPS-KIM, respectively. The location of each station is determined within each grid based on the location metadata of AWS stations and grid specifications for GDAPS-KIM. \textcolor{black}{Of course, our benchmark also supports predicting rainfall as a regression task or classification tasks having more categories. However, the evaluation should be based on the \autoref{tab:rain_fall_stat}, which is generally used in South Korea\,\cite{sohn2005statistical}.}

\vspace{-10pt}
\subsection{Dataset Split}
\vspace{-5pt}

We split the data temporally into three non-overlapping datasets by repeatedly using approximately 4 days for training followed by 2 days for validation and 2 days for testing. Separation by month or year could cause a shift in distribution between datasets. With reference to S{\o}nderby et al.\,\cite{sonderby2020metnet}, this category of temporal split is utilized. Because the data scale is small, more proportions are devoted to validation and testing compared to S{\o}nderby et al.\,\cite{sonderby2020metnet}.

Due to the overlap between the predicted instances at the same time from different GDAPS-KIM simulations, it is not straightforward to perform this split. For example, there are multiple predictions for April 21st, 2021 2:00PM, including a prediction ran at midnight the same day with lead time of 14 hours, and one ran at midnight of April 20th with 38 hour lead time, to name a few. In this paper, we follow the strategy that divides among prediction simulations, based on the point in time in which the predictions were ran. Since predictions from GDAPS-KIM simulations are highly conditioned on the initial state, our benchmark uses the aforementioned scheme to prevent predictions from the same initial state to leak into different data splits.


\vspace{-10pt}
\section{Optimization for Post-Processing the NWP Outputs}
\label{sec4}
\vspace{-5pt}


\paragraph{Problem Formulation.} In this work, we consider the following optimization model:
\vspace{-5pt}
\begin{gather}
    \min_{\vw} \Big\{ \mathcal{L}(\vw; \mathcal{D}) \triangleq \mathbb{E}_{(\mathbb{X}_t, \mathbb{Y}_t) \sim \mathcal{D}} [\ell(\mathbb{X}_t,\mathbb{Y}_t; \vw)] \Big\}
    \vspace{-10pt}
\end{gather}

where $\mathcal{L}$ is the objective function parameterized by $\vw$ on the dataset $\mathcal{D}$ having the input as the NWP outputs $\mathbb{X}_t$, the corresponding rainy output $\mathbb{Y}_t$ at time $t$, and $\ell$ is the classification loss between the output of the neural network and the ground-truth. $\mathbb{X}_t$ is a sequence of NWP predictions printed at time $t$, $\vx_{(t,0)}, \vx_{(t,1)}, \cdots, \vx_{(t, L-1)}$ where $\vx_{(t,i)}$ is the prediction at time $t+i$ based on a simulated NWP output at time $t$, $L$ is the length of lead time which means the prediction of additional hour per simulation, $\mathbb{Y}_t$ is a sequence of AWS observations $\vy_{t}, \vy_{t+1}, \cdots, \vy_{t + L-1}$. In this approach, the prediction of the neural network at time $t$ is defined as the following probabilistic forecast:
\vspace{-5pt}
\begin{gather}
    f(\tau; \mathbb{X}_{t_0}, \vw, ws) = \mathbb{P} (\vy_\tau | \vx_{(t, \Delta -ws+1)}, \cdots, \vx_{(t, \Delta - 1)}, \vx_{(t, \Delta)}) 
    \vspace{-10pt}
\end{gather}

where $f(\cdot; \vw)$ is the neural network parameterized by $\vw$, $t$ is the time at which the forecast is made, $\Delta = \tau - t_0$, $ws$ is the window size which is the number of consequent observations used for inference, and $\mathbb{P}(\cdot)$ is the probability matrix whose each pixel in grid indicates the probability of class among \{`non-rain', `rain', `heavy rain'\}.


\vspace{-5pt}
\paragraph{Algorithm Description.} Here, we describe one inference step (say the $\tau$-th; $\tau \geq t$) of the proposed post-NWP optimization task. Firstly, the subset $\mathbb{X}_{t}$ is reorganized with channel-wise concatenation: $[\vx_{(t, \Delta -ws+1)}; \cdots; \vx_{(t, \Delta - 1)}; \vx_{(t, \Delta)}] \in \mathbb{R}^{(C*ws) \times W \times H}$ where $W, H$ is the width and height of the grid map and $C$ is the number of variables in each pixel--and thus each instance after reorganization is 3D-shaped. We then create batches based on the modified $\mathbb{X}_{t}$ and perform forward and backward propagation. The model is trained over the entire dataset until convergence, or pre-fixed total epochs.

\vspace{-5pt}
\paragraph{Temporal Dependencies on Window Size \& Lead Time.}
The problem formulation of post-NWP optimization elicits two important temporal parameters that influence the learning problem. First, the window size of NWP predictions used for post-processing affects the richness of the input information. While larger window size allows the post-processing model to consider ample information, it may be more difficult to optimize the model. Another important temporal component is lead time. Predictions from different lead times exhibit distinct characteristics, leading to differences in evaluation performance. Taking this into consideration, it may be beneficial to train separate post-processing models for different lead times. For our baseline analysis, we focus on a single model trained to handle all possible lead times, as a representative baseline.



\vspace{-5pt}
\paragraph{Evaluations.} For evaluation, the validity of forecasting algorithms are assessed using diverse statistical metrics, which are commonly used for precipitation forecasting.
Following common practice in the multi-classification settings\,\cite{grandini2020metrics}, such metrics can be calculated based on the number of true positives\,($TP_k$), false positives\,($FP_k$), true negatives\,($TN_k$), and false negatives\,($FN_k$) for some generic class $k$:

\vspace{-5pt}
\begin{itemize}
    \item \textbf{Accuracy\,(ACC)} returns an overall measure of how much the model is correctly predicting on the entire set of data.
    \item \textbf{Probability of Detection\,(POD)} is a recall calculated as $\frac{TP_k}{TP_k + FP_K}$.
    \item \textbf{Critical Success Index\,(CSI)}\,\cite{donaldson1975objective} is a categorical score that considers more aspects of the confusion matrix similar with F1-score having the value as $\frac{TP_k}{TP_k+FN_k+FP_k}$.
    \item \textbf{False Alarm Ratio\,(FAR)}\,\cite{barnes2009corrigendum} is the number of false alarms per the total number of warnings or alarms, also known as the probability of false detection. It is computed as $\frac{FP_k}{TP_k+FP_k}$
    \item \textbf{Bias} is defined as the ratio of the observed frequency of occurrence of a phenomenon to the frequency of the occurrence predicted by the forecast model. $\frac{TP_k + FP_k}{TP_k + FN_k}$. If it has a value greater than 1, it means that the frequency of occurrence predicted by the forecasting model is greater than the frequency of occurrence of the actual phenomenon, and therefore more frequent prediction is made. The more accurate the forecast, the closer this index is to 1.
\end{itemize}
\vspace{-15pt}
\section{Experiment}\label{sec5}
\vspace{-10pt}

\begin{table}[t]
\centering\footnotesize
\addtolength{\tabcolsep}{-3pt}
\caption{Evaluation metrics of KoMet and baseline models for precipitation while 12 variables are utilized for the training. Best performances are marked in bold.}
\begin{tabular}{@{}c|ccccc|ccccc@{}} 
\toprule
          & \multicolumn{5}{c|}{Rain}     & \multicolumn{5}{c}{Heavy Rain} \\ \midrule
          & Acc & POD & CSI & FAR & Bias & Acc  & POD  & CSI & FAR & Bias \\ \midrule
GDAPS-KIM & 0.747 & \textbf{0.633} & 0.263 & 0.690 & 2.042 & 0.985 & \textbf{0.055} & \textbf{0.045} & \textbf{0.795} & 0.266 \\
U-Net     & 0.840 & 0.441 & 0.282 & 0.562 & 1.007 & 0.983 & 0.040 & 0.029 & 0.906 & 0.426 \\
ConvLSTM  & \textbf{0.869} & 0.387 & 0.296 & \textbf{0.444} & 0.696 & \textbf{0.986} & 0.007 & 0.006 & 0.889 & 0.059 \\
MetNet    & 0.854 & 0.468 & \textbf{0.314} & 0.512 & 0.959 & \textbf{0.986} & 0.013 & 0.012 & 0.838 & 0.079 \\ \bottomrule
\end{tabular}
\label{tab:rain_perfor}
\vspace{-10pt}
\end{table}


\begin{figure}[t]
\centering
         \centering
         \includegraphics[width=0.9\textwidth]{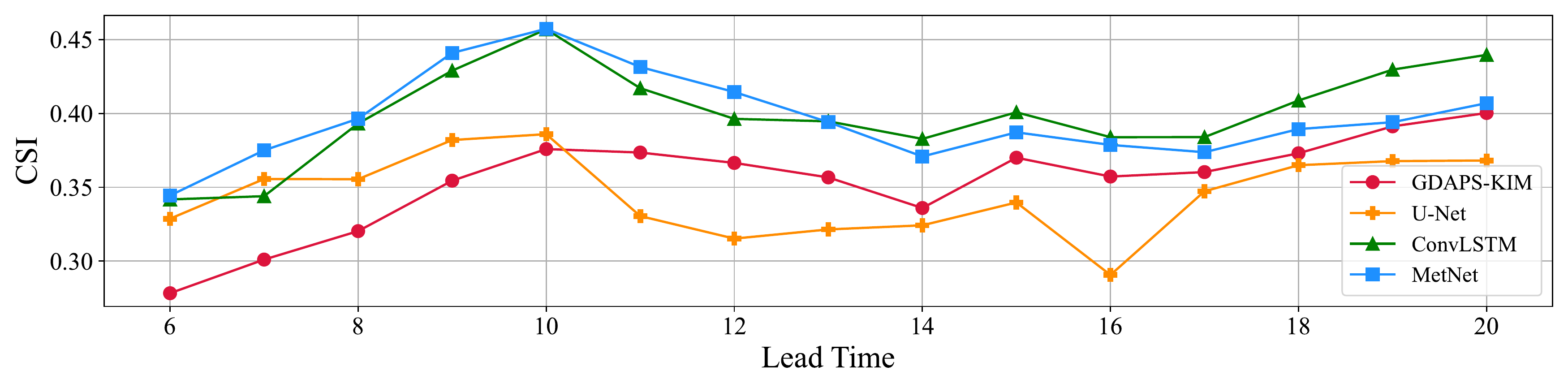}
\vspace{-10pt}
\caption{CSI scores of GDAPS-KIM and baseline models for a class `rain'. Scores are given for predictions corresponding to specific lead times, ranging from 6 to 20 hours.}
\vspace{-15pt}
\label{lead_conditioning}
\end{figure}


\begin{figure}[t]
\begin{center}
   \begin{subfigure}[b]{0.28\textwidth}
         \centering
         \includegraphics[width=\textwidth]{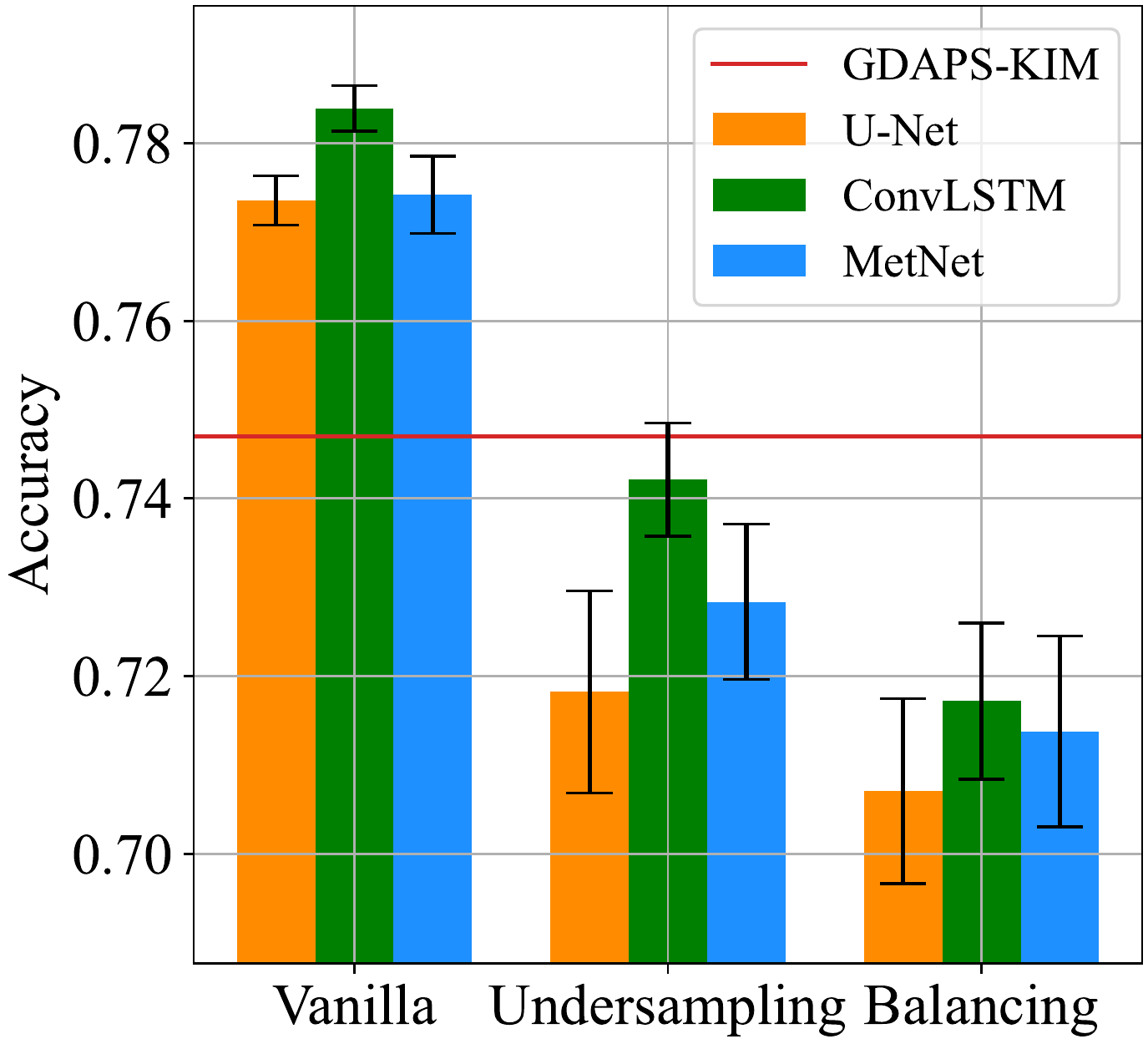}
         \caption{Acc}
         \label{fig:performance_by_method_rain_acc}
     \end{subfigure}
   \begin{subfigure}[b]{0.28\textwidth}
         \centering
         \includegraphics[width=\textwidth]{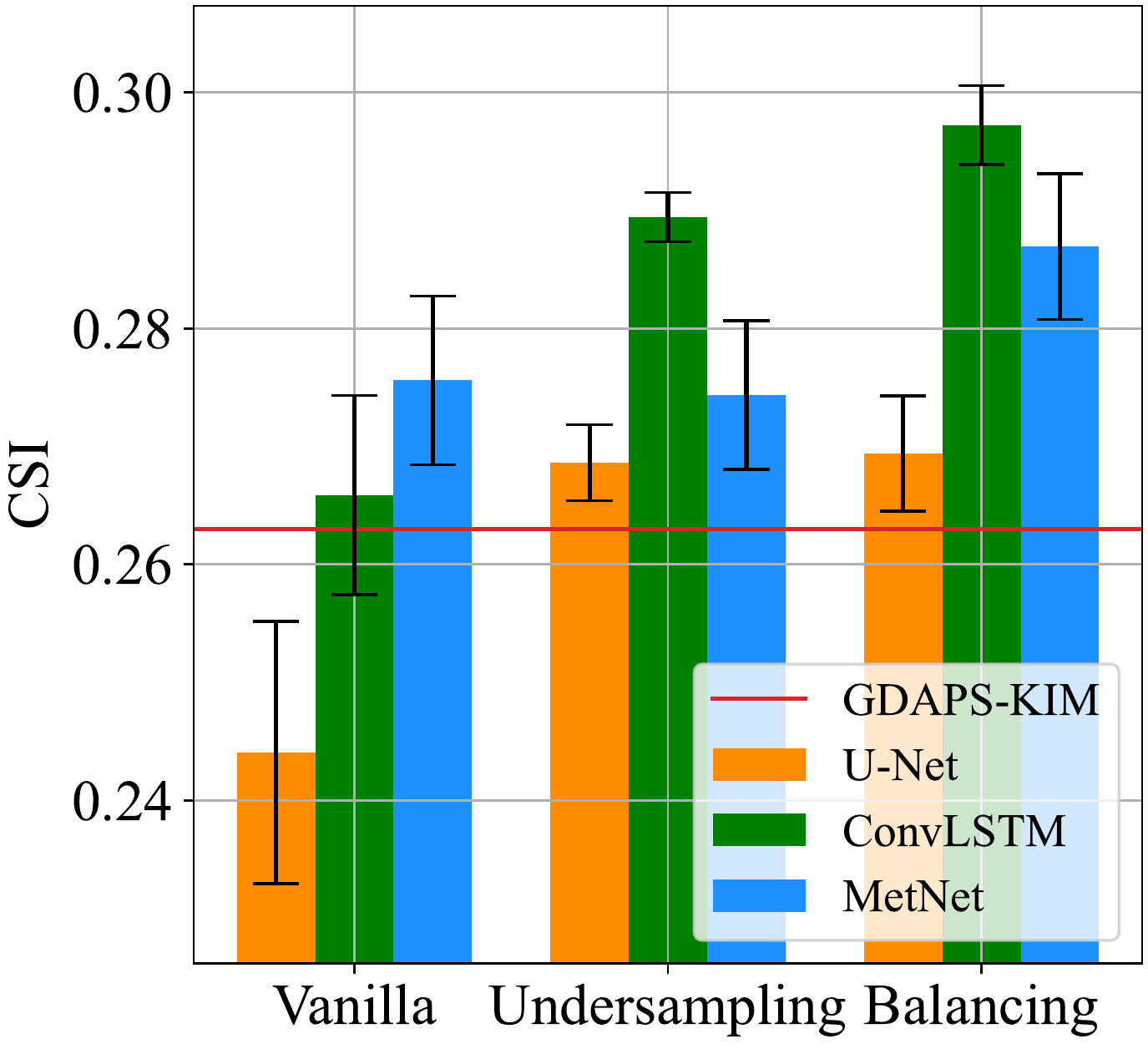}
         \caption{CSI}
         \label{fig:performance_by_method_rain_csi}
     \end{subfigure}
   \begin{subfigure}[b]{0.28\textwidth}
         \centering
         \includegraphics[width=\textwidth]{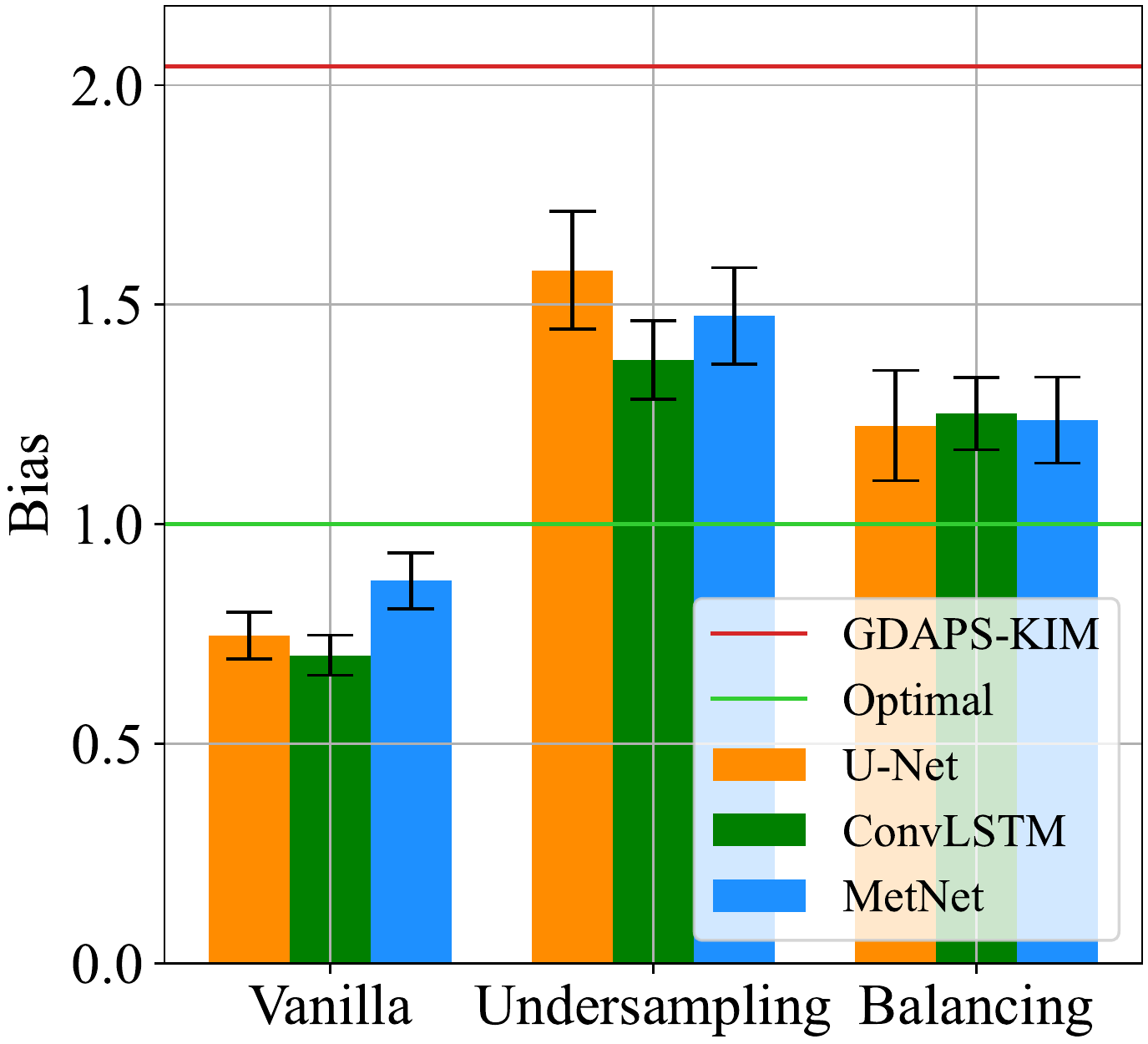}
         \caption{Bias}
         \label{fig:performance_by_method_rain_bias}
    \end{subfigure}
\end{center}
\vspace{-10pt}
\caption{Performance of the models according to the changes of sampling strategies. Vanilla indicates baseline training strategy without resampling of AWS targets. \textcolor{black}{The metrics are averaged over 10 different random seeds and error bars indicate the 95\% confidence intervals.}}
\vspace{-10pt}
\label{sampling}
\end{figure}


\begin{figure}[t]
\begin{center}
   \begin{subfigure}[b]{0.32\textwidth}
         \centering
         \includegraphics[width=\textwidth]{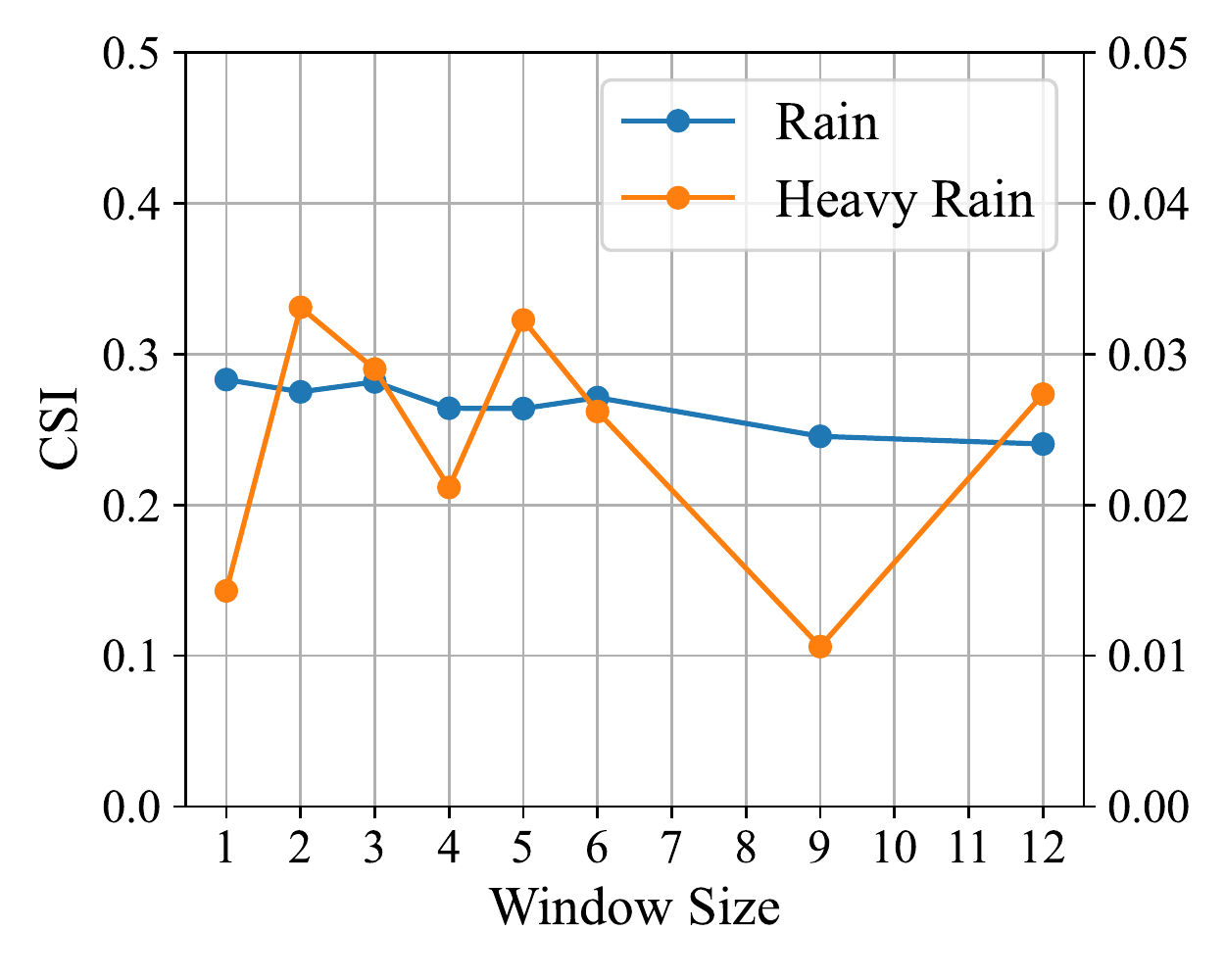}
         \caption{U-Net, window size}
         \label{fig:window_size_ablation_unet}
     \end{subfigure}
   \begin{subfigure}[b]{0.32\textwidth}
         \centering
         \includegraphics[width=\textwidth]{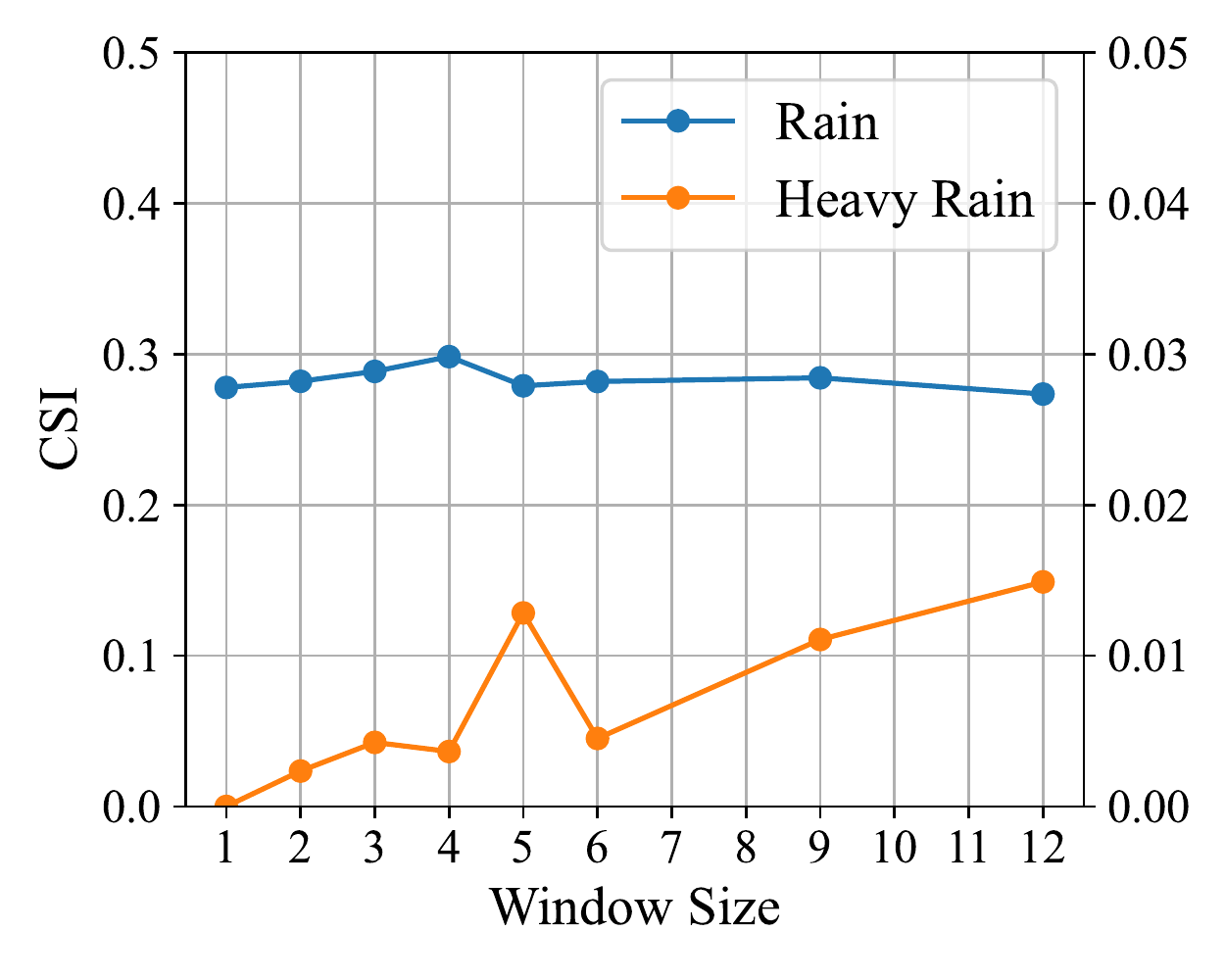}
         \caption{ConvLSTM, window size}
         \label{fig:window_size_ablation_convlstm}
     \end{subfigure}
   \begin{subfigure}[b]{0.32\textwidth}
         \centering
         \includegraphics[width=\textwidth]{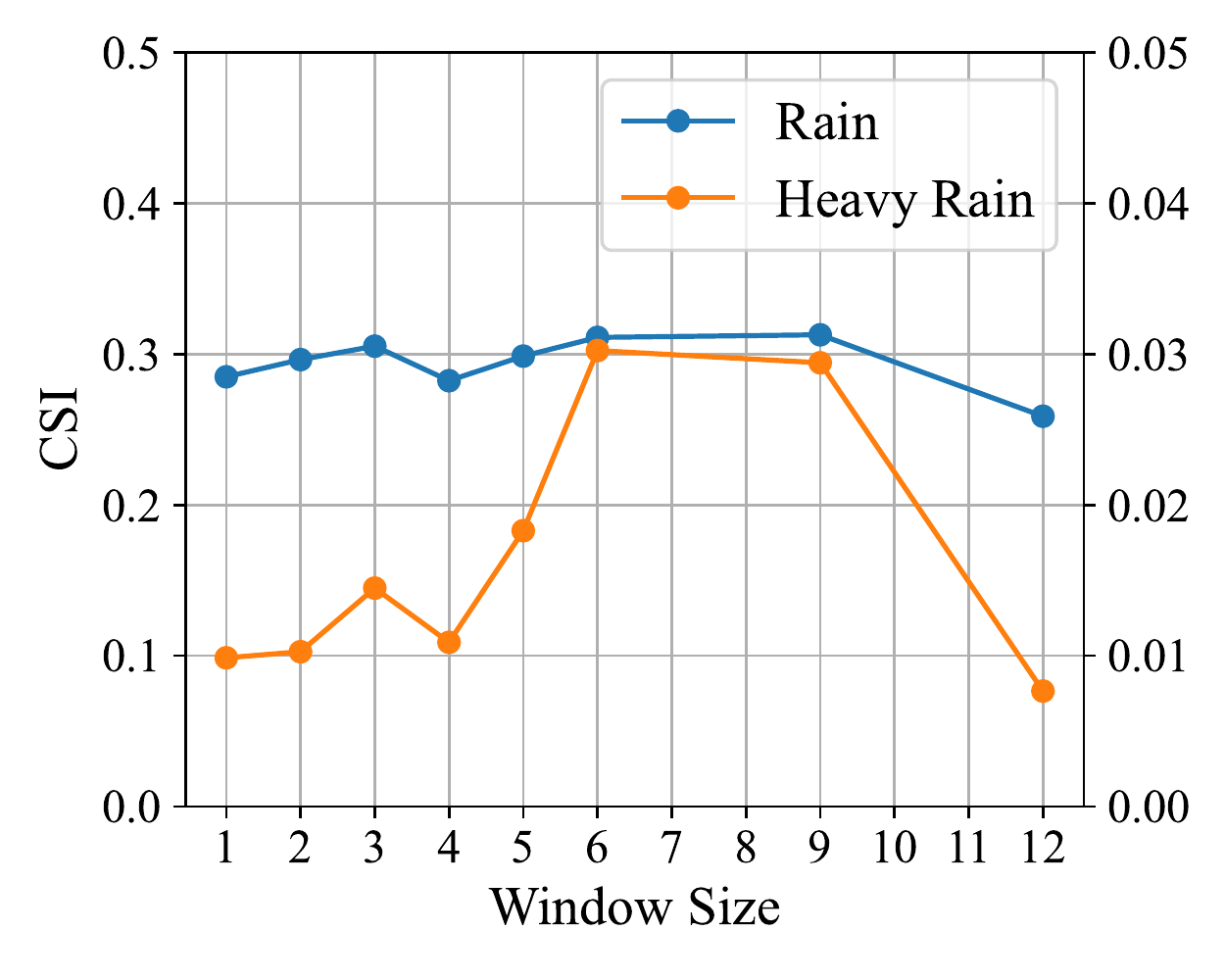}
         \caption{MetNet, window size}
         \label{fig:window_size_ablation_metnet}
     \end{subfigure}
       \begin{subfigure}[b]{0.32\textwidth}
         \centering
         \includegraphics[width=\textwidth]{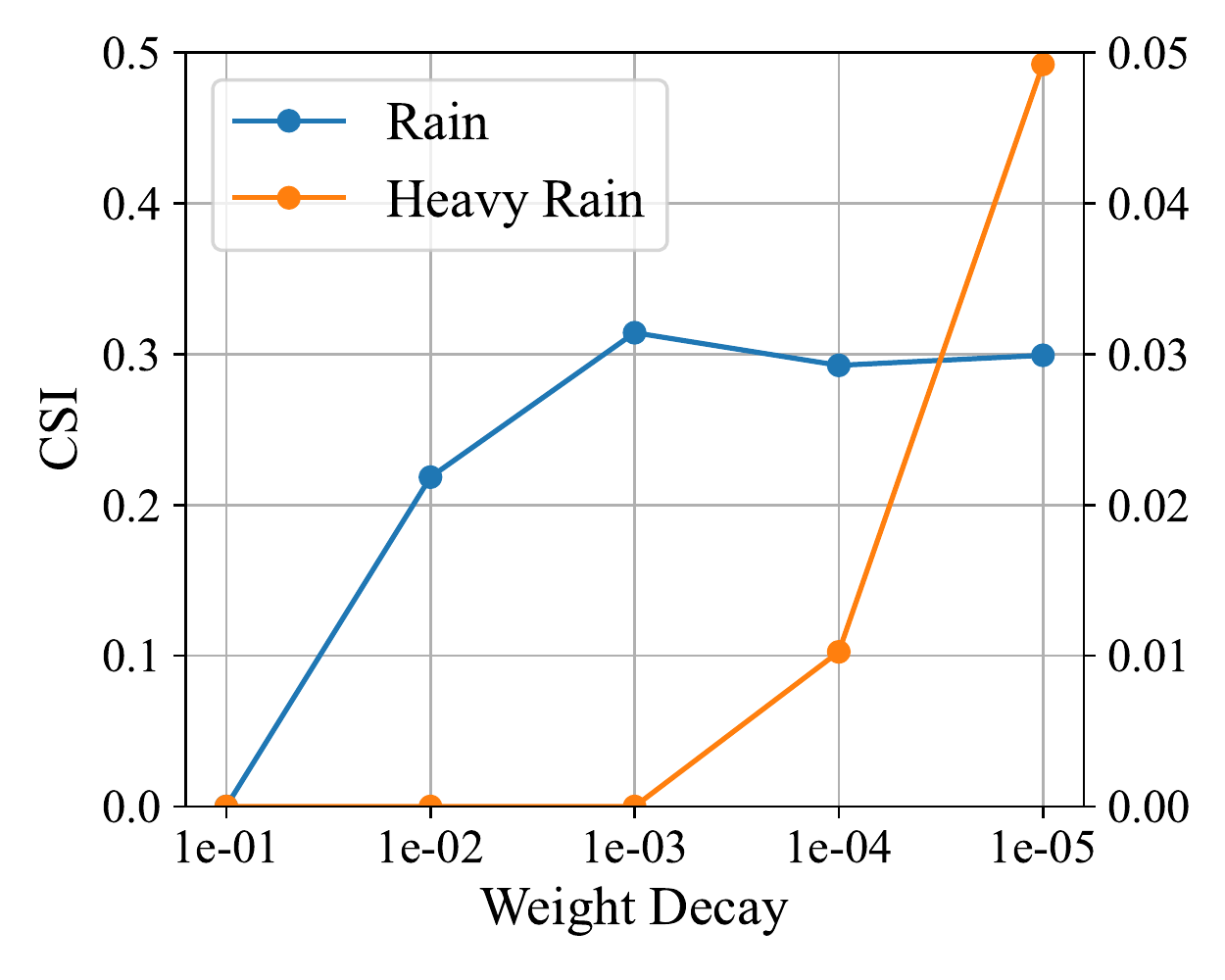}
         \caption{U-Net, weight decay}
         \label{fig:weight_decay_ablation_unet}
     \end{subfigure}
  \begin{subfigure}[b]{0.32\textwidth}
         \centering
         \includegraphics[width=\textwidth]{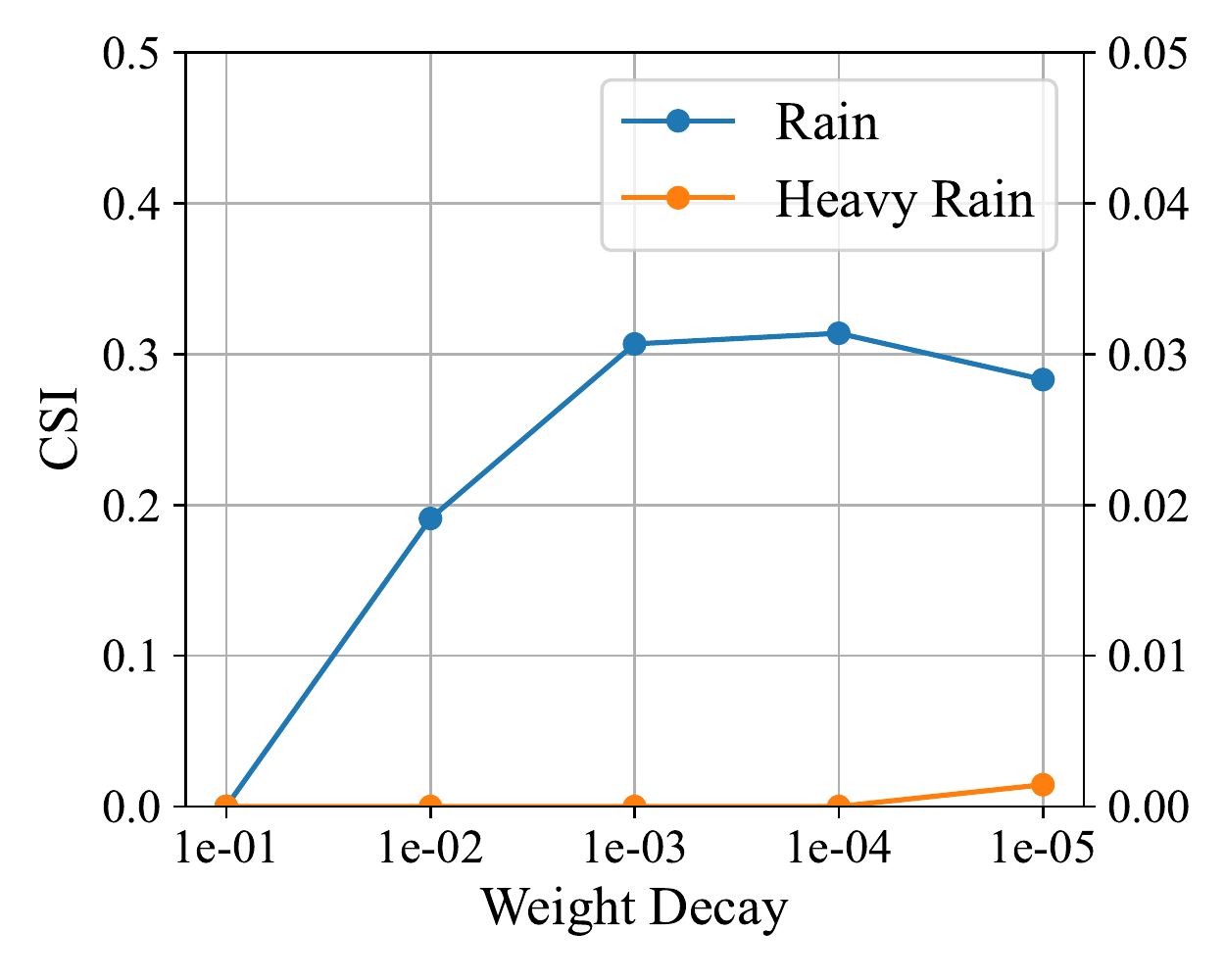}
         \caption{ConvLSTM, weight decay}
         \label{fig:weight_decay_ablation_convlstm}
     \end{subfigure}
  \begin{subfigure}[b]{0.32\textwidth}
         \centering
         \includegraphics[width=\textwidth]{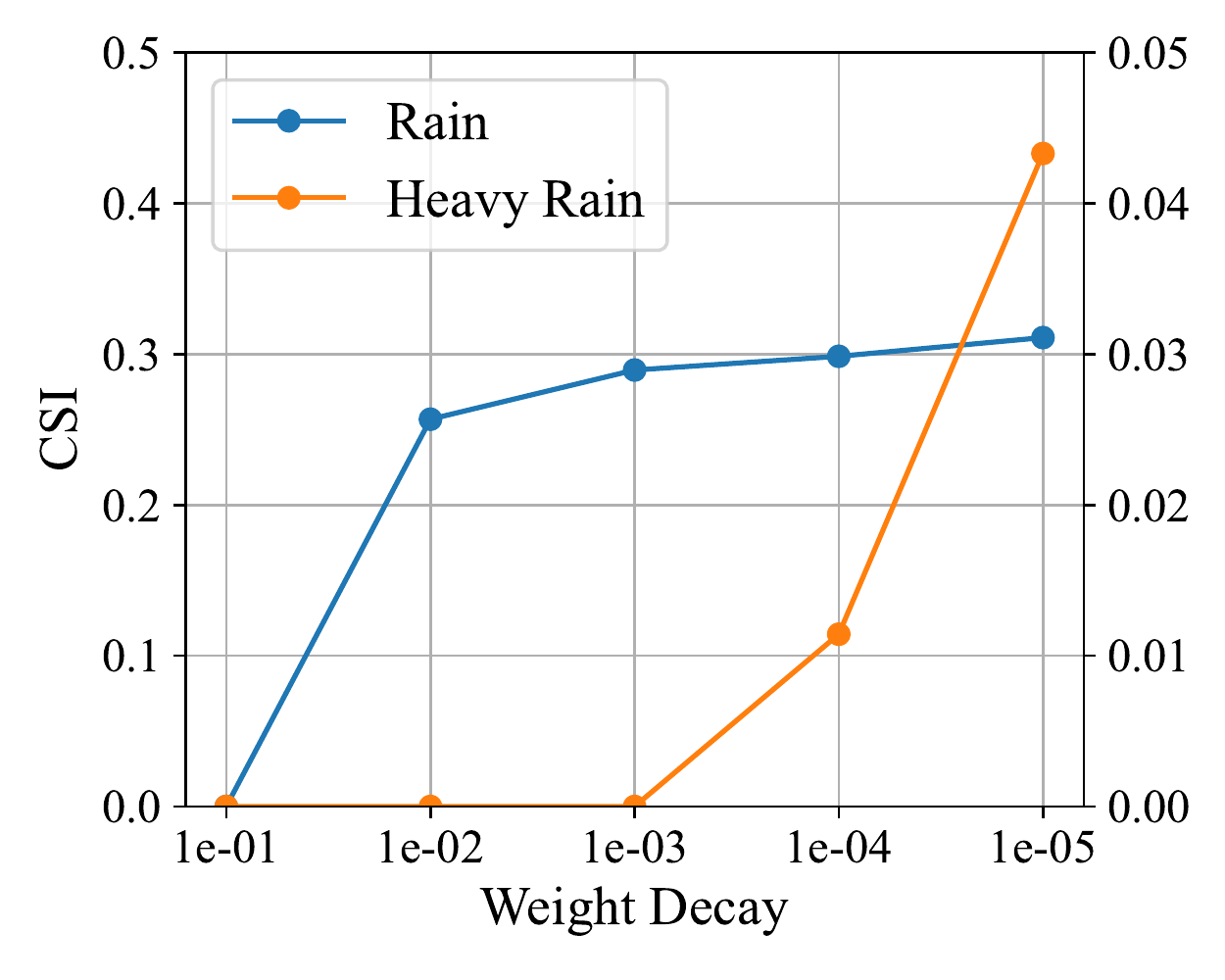}
         \caption{MetNet, weight decay}
         \label{fig:weight_decay_ablation_metnet}
     \end{subfigure}
\end{center}
\vspace{-5pt}
\caption{(a), (b), (c): CSI performance according to changes in window size in hours; (d), (e), (f): CSI performance according to changes in weight decay.}
\vspace{-15pt}
\label{ablation}
\end{figure}


\subsection{Benchmarking neural network architectures}
\vspace{-5pt}
Various models have been adopted or designed to effectively extrapolate future weather status from past and current state. Here, we provide three baseline architectures for precipitation forecasting: U-Net\,\cite{ronneberger2015u}, ConvLSTM\,\cite{ravuri2021skilful}, and MetNet\,\cite{sonderby2020metnet,espeholt2021skillful}\,(detailed explanations are provided in Appendix).

\autoref{tab:rain_perfor} shows the results for various lead times ranging from 6 to 87 hours, according to the changes of the architectures. Compared to the statistics of GDAPS-KIM for predicting class `rain', the qualities of post-processed predictions with the three networks are consistently improved while that of MetNet is the best. In contrast, deep learning-based post-processing rather hinders performance for `heavy rain' prediction. In particular, MetNet faces severe performance degradation. We leave this phenomenon as a challenge to be addressed. \autoref{lead_conditioning} illustrates the results of CSI scores for precipitation of $\geq 0.1$\,(mm/h) over lead time. Here, MetNet also outperforms other baseline models.


\vspace{-5pt}
\subsection{Influential Variable Selection}
\vspace{-5pt}
As listed in \autoref{tab:variable}, GDAPS-KIM contains many variables. Naively using all available variables may impede generalization, therefore it is important to select the most prominent variables. Variable selection experimentation can be done in various ways, and we propose a baseline experiment in which the variables are divided into Pres type and Unis type groups and considered as manipulated variables and control variables, respectively. An example is illustrated in the Appendix, and we observe that selecting certain significant variables performs better than simply using more variables.



\vspace{-5pt}
\subsection{Sampling Strategies for Class-Imbalance Issue}
\label{sec:sampling_strategies}
\vspace{-5pt}
As a solution to the class-imbalance issue, we provide two probabilistic data sampling techniques for target transformation: (1) \textbf{Under-Sampling for `No Rain' Points} that uses only a fraction $p$ of the `no rain' points used for learning, to reduce the no precipitation points, and (2) \textbf{Balancing for `Rain Points'} that maintains the ratio between `no rain' points and `rain' points to $1:p$ by under-sampling `no rain' points, when there are excessive `no rain' points.

\autoref{sampling} depicts the performance of models trained with different sampling strategies. Although the performance change differs depending on the architecture, it can be confirmed that generalization can be further facilitated if the data distribution is adjusted fairly. It is expected that an approach that manipulates the loss function can also be effective\,\cite{lin2017focal, cao2019learning}.

\vspace{-5pt}
\subsection{Sensitivity to Hyperparameter Settings}
\vspace{-5pt}
\paragraph{Window Size.} The first row of \autoref{ablation} shows the CSI scores as the window size changes. In `rain', the CSI score is relatively unchanged, while the `heavy rain' case is shown to be extremely sensitive.
\vspace{-5pt}
\paragraph{Weight Decay.} The second row of \autoref{ablation} shows the CSI scores according to the changes in magnitude of weight decay. When comparing the performance of rain and heavy rain, we see differing trends according to increase in weight decay.

The results above suggest the need for combinatorial optimization among a variety of hyperparameters such as batch size, optimizer, beta of batch normalization, and learning rate.
\vspace{-10pt}
\section{Conclusion \& Future Work}
\label{sec6}
\vspace{-5pt}

In this paper, we present a new benchmark dataset, termed \textbf{KoMet}, for a hybrid NWP-DL workflow that post-processes the rain prediction from South Korea's NWP models under the supervision of surface-level observations from AWS. Our work engages in the solution of precipitation forecasting. We believe that our work opens the door to develop robust and explainable forecasting services that benefit the broad society. Our benchmark dataset will lower the barriers to entry for those in the DL community to contribute to research in precipitation forecasting.

\begin{ack}
This work was supported by the Korea Meteorological Administration Research and Development Program "Development of AI techniques for Weather Forecasting" under Grant (KMA2021-00121) and Institute of Information \& communications Technology Planning \& Evaluation (IITP) grant funded by the Korea government(MSIT) (No.2019-0-00075, Artificial Intelligence Graduate School Program(KAIST)). We thank Jaehoon Oh and Sangmook Kim for discussing the pipelines of precipitation forecasting and Yun Am Seo (Jeju National University) and Min-Gee Hong (National Institute of Meteorological Sciences) for discussing the data pre-processing.
\end{ack}

\bibliography{ref}
\bibliographystyle{plain}

\appendix

\section{Overview of Appendix}

In this supplementary material, we present additional details, results, and experiments that are not included in the main paper due to the space limit. 

\section{Access to Dataset}

\begin{itemize}
    \item URL: \url{https://github.com/osilab-kaist/KoMet-Benchmark-Dataset}
    \item Dataset URL: \url{https://www.dropbox.com/s/qachyygl2ouuy1v/KoMet.v1.0.tar.gz?dl=0}
    \item Hosting and maintenance plan: The dataset is provided via Dropbox. Our GitHub repository provides tools for data access as well as model training and evaluation for our proposed workflow. We will continue to ensure public access to our dataset, and possibly provide the dataset through multiple hosting services, if the need arises.
    \item License: We include the MIT License about the use of public GitHub repository. 
\end{itemize}

\section{Ethics Statement}

To address potential concerns, we describe the ethical aspect with respect to security and the environment.

\paragraph{Security.}
Weather information is an important factor in a country's economic and military security. Nevertheless, general users do not have access to up-to-date NWP prediction data, so if access to live NWP data is well protected, the security problem will be sufficiently prevented.

\paragraph{Environment.} While our approach brings more benefits in generalizing precipitation forecasting, it still relies on NWP simulations which require vast amounts of computation, and thus energy consumption. The same is true for training and running deep learning models for post-processing. However, we believe that the positive impact of accurate weather forecasts on energy consumption and planning far outweighs these costs and will lead to more environmentally conscious energy policies.

\section{Limitations and Future Directions}
In this section, we describe the limitations of our work and future directions for further development.

\paragraph{Limitations.}

Although we illustrate the advantages of our benchmark dataset, the bottleneck lies in the accuracy of GDAPS-KIM as well as AWS. We only consider the summer season of South Korea, specifically, July and August. We do not investigate other NWP datasets. We also do not consider the precipitation nowcasting task, with lead times of less than 3 hours, as this would limit the deep model's window size to less than 3, possibly hindering model performance. 

\paragraph{Future Directions.}
In future work, we aim to explore more robust training strategies under the data scarcity scenario. Furthermore, thorough investigation is needed for the selection of important variables in \autoref{tab:variable}. In addition, we intend to develop our methods in other seasons such as Spring, Fall, and Winter. Lastly, we plan to explore methods to tackle precipitation nowcasting, using a limited window size.

\section{Additional Detailed Explanations}

\begin{itemize}
    \item \textcolor{black}{\textbf{Thresholds for Rain and Heavy Rain.} KMA defines that rain occurs when the rainfall is over 0.1mmh$^{-1}$ and heavy rain occurs when the rainfall is over 10mmh$^{-1}$. Literature based on the precipitation forecasting of the Korea Peninsula follows this standard\,\cite{sohn2005statistical, song2019evaluation}.}

    \item \textbf{U-Net}\,\cite{ronneberger2015u} is a model designed to solve the image segmentation problem in biomedical images. 
    During the propagation of the encoder part, important features can be captured in a low-dimensional form. 
    Applying this to the task of NWP post-processing, U-Net is used to extract the meteorological features from GDAPS-KIM and decode them into a feature map containing precipitation predictions, similar to\,\cite{ravuri2021skilful}. Note that\,\cite{ravuri2021skilful} used radar observations as inputs, rather than NWP predictions.

    
    \item \textbf{Convolutional Long Short-Term Memory Neural Network\,(Conv-LSTM)}\,\cite{shi2015convolutional, shi2017deep} 
    is a model that combines LSTM and convolutional operations, each designed to model temporal and spatial relationships, respectively, within sequences of images.
    This enables the model to identify relationships between sequential NWP prediction maps to derive a refined predictions for precipitation.
    The model structure consists of encoding, decoding and forecasting modules comprised of stacks of ConvLSTM layers.
    
    \item \textbf{MetNet.}\,\cite{sonderby2020metnet,espeholt2021skillful}
    In MetNet, the spatial downsampler module first reduces the size by passing the input through several convolutional layers. Next, the ConvLSTM structure is used in the temporal encoder to create an output tensor with spatial and temporal information for each pixel. Lastly, such propagated feature map goes through the self-attention block of the Spatial Aggregator, collects information in the global range, and finally passes through a classifier to output the precipitation amount for each pixel as a probability distribution.
\end{itemize}

\section{Implementation Details}

The exact implementation details and reproduction instructions for our experiments are available on our GitHub repository. We explain important details in the following paragraphs.

\begin{figure}[h]
\begin{center}
   \begin{subfigure}[b]{0.49\textwidth}
         \centering
         \includegraphics[width=\textwidth]{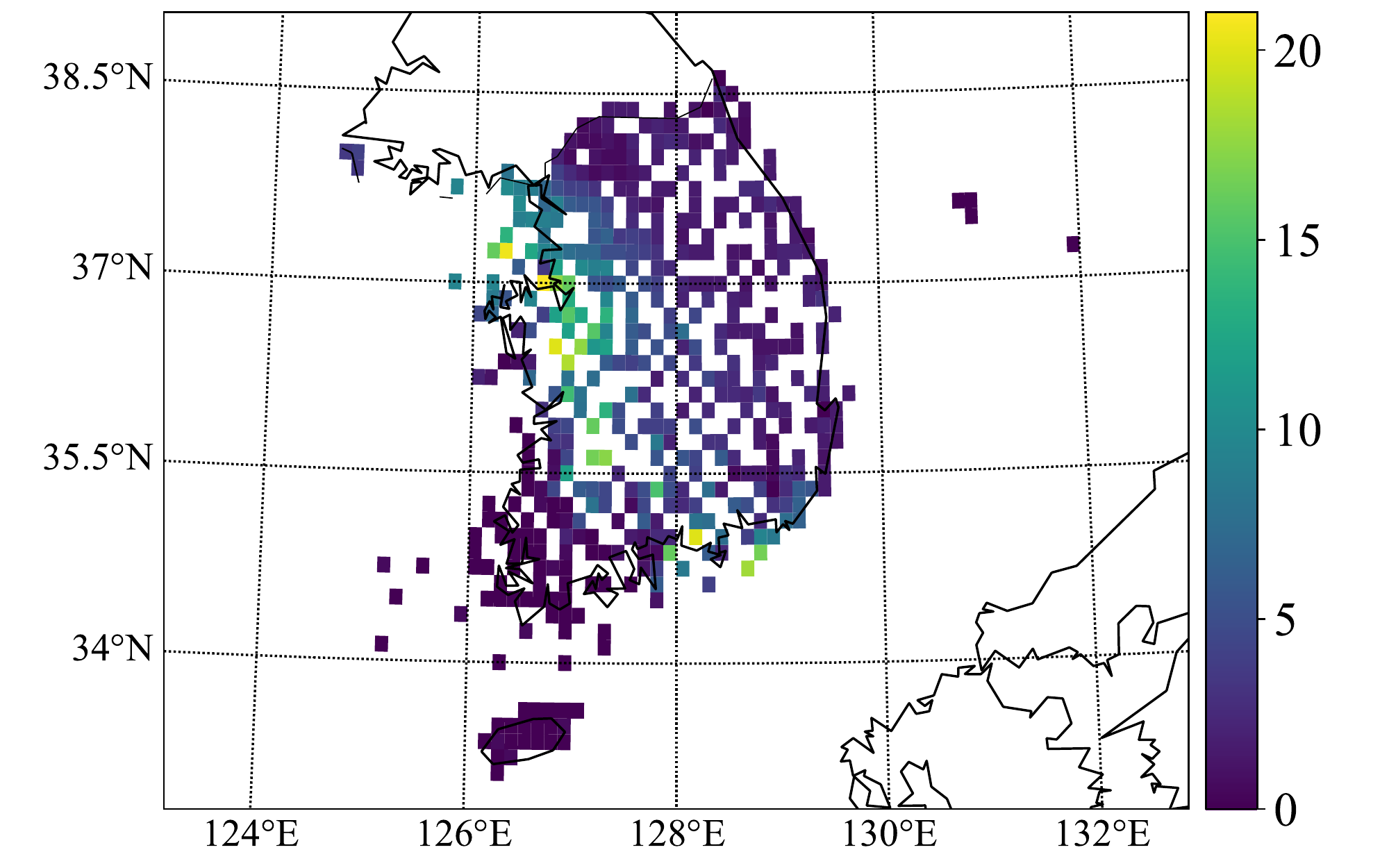}
         \caption{Original}
         \label{fig:interpolation_before}
     \end{subfigure}
   \begin{subfigure}[b]{0.49\textwidth}
         \centering
         \includegraphics[width=\textwidth]{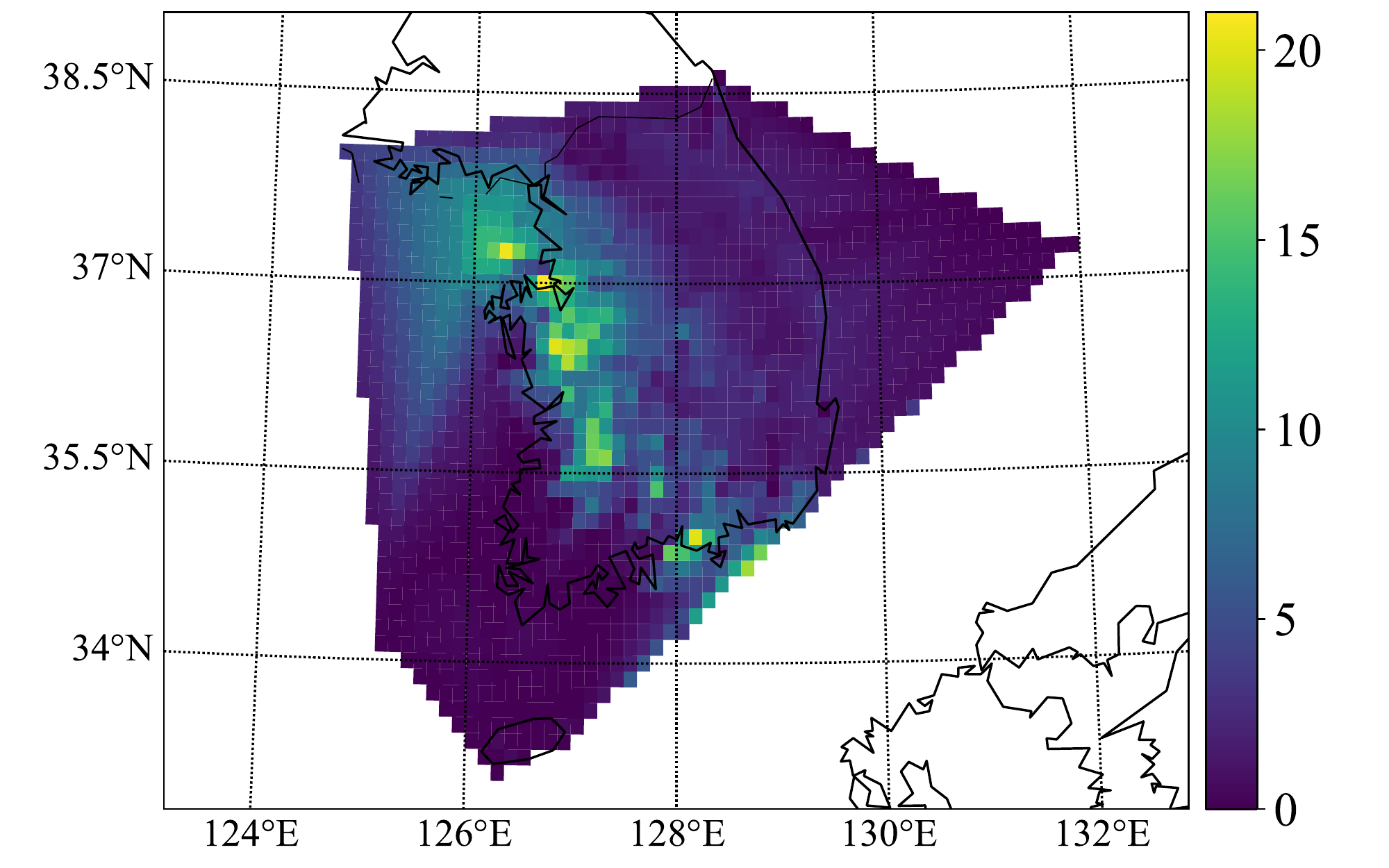}
         \caption{Interpolated}
         \label{fig:interpolation_after}
     \end{subfigure}
\end{center}
\caption{Visualization of AWS observations of precipitation before and after interpolation, on April 23, 2020 06:00 UTC. The unit of precipitation is mm/h.}
\label{fig:interpolation}
\end{figure}

\paragraph{Data Preprocessing.} We refer to the settings in \cite{espeholt2021skillful}. For the transformation of GDAPS-KIM, we only apply Min-Max normalization to each variable. We employ various transformations on the ground-truth AWS targets to mitigate the sparsity of observations. In all of our experiments, we apply linear interpolation between available observations to fill in the non-learning points, using functions provided by the \textsc{scipy} package. This is illustrated in \autoref{fig:interpolation}. To tackle the class imbalance issue, we study two sampling strategies in \autoref{sec:sampling_strategies}. Details of sampling strategies are explained in the following paragraphs.

\paragraph{Under-Sampling for `No Rain' Points.} As the ratio of `no rain' is significantly higher than that of `rain' and `heavy rain', we consider a simple sampling strategy of randomly sampling a subset of `no rain` points within each sample, at a rate of $p$. The remaining `no rain' points are discarded during training, i.e., set to nan. We use a sampling ratio of $p=0.20$.

\paragraph{Balancing for `Rain' Points.} This technique considers the ratio of `rain' and `heavy rain' to `no rain' within each feature map. Specifically, let $R$ be the number of points with `rain' or `heavy rain', and $N$ be the number of `no rain' points. Given the target rate $p$, we randomly sample `no rain' points to match $p = N/R$ as closely as possible. We also use a sampling ratio of $p=0.20$ for this method. The specific strategy applied to each sample is as follows:

\begin{itemize}
    \item Case 1. $R = 0$: The sample is not used for training.
    \item Case 2. $(R × p) < N$: $(N - R × p)$ `no rain' points are discarded from the sample.
    \item Case 3. $(R × p) >= N$: The sample is used as is.
\end{itemize}

\paragraph{Training Settings.}

We use the Adam optimizer with the default learning rate of 0.001, and beta parameters of 0.9 and 0.999. We train the models for 20 epochs and select the best epoch based on the highest CSI performance on the validation set. Reported performances are based on the test set, unless otherwise specified. We use a window size of 3 and lead times ranging from 6 to 87 hours. We use two Pres type variables (\textit{T}, \textit{rh\_liq}) at three isobaric planes,\{500hPa, 700hPa, 850hPa\}; and six variables of Unis type (\textit{rain, q2m, rh2m, t2m, tsfc, ps}). These baseline settings are used in all experiments, unless otherwise stated.

\paragraph{Resources.} We run our experiments on NVIDIA GeForce RTX 3090 Ti GPUs. Each training run takes approximately 30 minutes.

\section{Additional Experiments}

\subsection{Influential Variables}
 
We conduct an ablation study to investigate the influence of each individual variable. We start with a baseline set of variables, curated under the guidance of weather forecasters in South Korea. This includes all baseline variables used in our other experiments, as well as one additional Unis variable: \textit{pbltype}. We then measure the performance of our baseline U-Net model when adding or subtracting individual variables used for training. We consider the \textit{u}, \textit{v} and \textit{u10m}, \textit{v10m} variables jointly, as they both pertain to wind. For Pres variables, we use values corresponding to three isobraic surfaces at \{500hPa, 700hPa, 850hPa\}. We report our results in \autoref{tab:variable_ablation} and find that while some metrics such as accuracy are relatively unaffected, others such as CSI and bias vary depending on the choice of variables.

\begin{table}[t]
\label{tab:variable_ablation}
\centering\footnotesize
\caption{Ablation study on the binary classification performance of our baseline U-Net model for `rain' and `heavy rain' depending on the selection of variables. Each row represents an addition (+) or a subtraction (-) of a single variable from our curated set.}
\begin{tabular}{ll|ccccc|ccccc}
\toprule
     \multirow{2}{*}{Type} & \multirow{2}{*}{Ablation} & \multicolumn{5}{c}{Rain} & \multicolumn{5}{c}{Heavy Rain} \\
      & & Acc &    POD &    CSI &    FAR &   Bias &        Acc &    POD &    CSI &    FAR &   Bias \\
\midrule
    Curated & & 0.840 & 0.441 & 0.282 & 0.562 & 1.007 & 0.983 & 0.040 & 0.029 & 0.906 & 0.426 \\
\midrule
    \multirow{4}{*}{Pres} & +uv &  0.859 &  0.385 &  0.280 &  0.493 &  0.759 &      0.983 &  0.042 &  0.031 &  0.896 &  0.407 \\
    & -T &  0.857 &  0.388 &  0.279 &  0.503 &  0.782 &      0.984 &  0.037 &  0.030 &  0.874 &  0.296 \\
    & -rh\_liq &  0.854 &  0.444 &  0.302 &  0.515 &  0.915 &      0.984 &  0.018 &  0.015 &  0.930 &  0.264 \\
    & +hgt &  0.851 &  0.392 &  0.272 &  0.530 &  0.834 &      0.983 &  0.046 &  0.035 &  0.880 &  0.388 \\
\midrule
    \multirow{10}{*}{Unis} & +hpbl &  0.851 &  0.378 &  0.266 &  0.527 &  0.799 &      0.984 &  0.032 &  0.025 &  0.889 &  0.287 \\
    & -pbltype &  0.840 &  0.483 &  0.300 &  0.558 &  1.093 &      0.983 &  0.041 &  0.031 &  0.891 &  0.378 \\
    & +psl &  0.862 &  0.361 &  0.271 &  0.480 &  0.695 &      0.985 &  0.035 &  0.028 &  0.870 &  0.268 \\
    & -q2m &  0.848 &  0.457 &  0.300 &  0.534 &  0.982 &      0.983 &  0.052 &  0.037 &  0.885 &  0.449 \\
    & -rh2m &  0.838 &  0.489 &  0.301 &  0.561 &  1.112 &      0.983 &  0.065 &  0.046 &  0.860 &  0.462 \\
    & -t2m &  0.846 &  0.460 &  0.299 &  0.540 &  1.000 &      0.983 &  0.044 &  0.032 &  0.892 &  0.409 \\
    & -tsfc &  0.844 &  0.510 &  0.317 &  0.544 &  1.119 &      0.983 &  0.067 &  0.048 &  0.850 &  0.445 \\
    & +uv10m &  0.843 &  0.447 &  0.288 &  0.553 &  0.999 &      0.984 &  0.049 &  0.039 &  0.837 &  0.299 \\
    & +topo &  0.855 &  0.411 &  0.288 &  0.510 &  0.839 &      0.984 &  0.057 &  0.044 &  0.842 &  0.361 \\
    & +ps &  0.852 &  0.445 &  0.300 &  0.521 &  0.927 &      0.985 &  0.026 &  0.022 &  0.884 &  0.227 \\
\bottomrule
\end{tabular}
\label{tab:variable_ablation}
\end{table}

\subsection{Lead Time on Heavy Rain Prediction}
\begin{figure}[t]
\begin{center}
   \begin{subfigure}[b]{\textwidth}
         \centering
         \includegraphics[width=\textwidth]{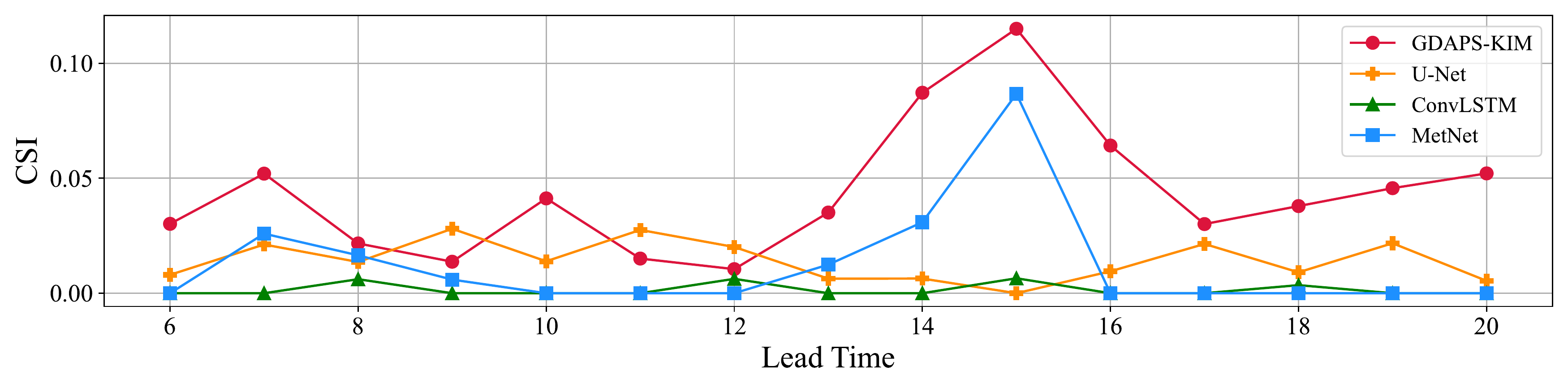}
         \caption{Heavy Rain}
         \label{fig:vanilla_performance_by_lead_time_heavy}
     \end{subfigure}
\end{center}
\vspace{-5pt}
\caption{CSI scores of GDAPS-KIM and baseline models for binary heavy rain classification. Scores are given for predictions corresponding to specific lead times, ranging from 6 to 20 hours.}
\vspace{-10pt}
\label{vanilla_performance_by_lead_time_appx}
\end{figure}

\autoref{vanilla_performance_by_lead_time_appx} illustrates the results of CSI scores for precipitation of $\geq 10.0$\,(mm/h) over lead time. Unlike the trends in rain prediction, interestingly GDAPS-KIM outperforms other baseline DL models. It seems that heavy rain prediction is a more ardous challenge to be addressed.

\subsection{Learning Rate}

\begin{figure}[h]
\begin{center}
   \begin{subfigure}[b]{0.32\textwidth}
         \centering
         \includegraphics[width=\textwidth]{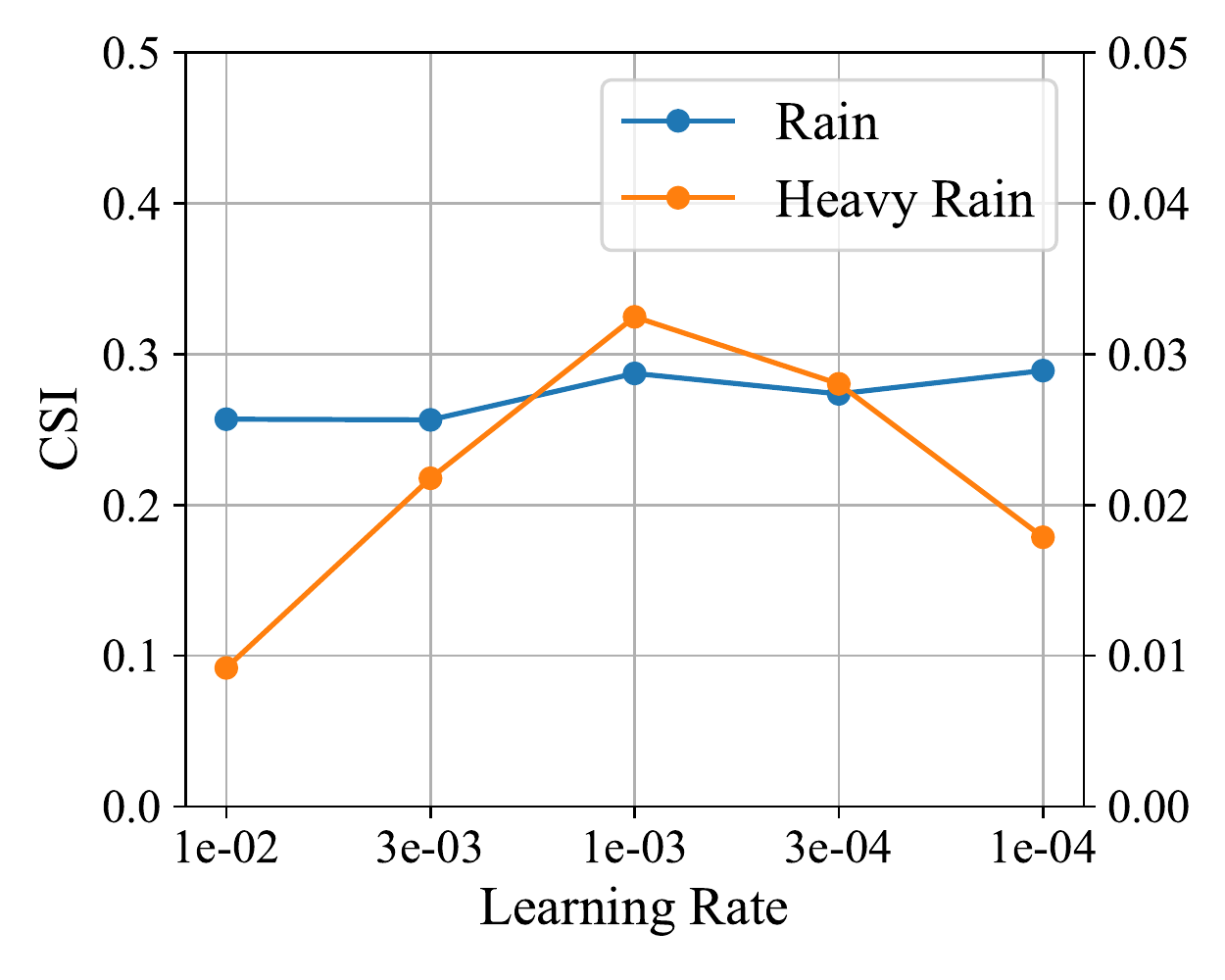}
         \caption{U-Net}
         \label{fig:learning_rate_ablation_unet}
     \end{subfigure}
   \begin{subfigure}[b]{0.32\textwidth}
         \centering
         \includegraphics[width=\textwidth]{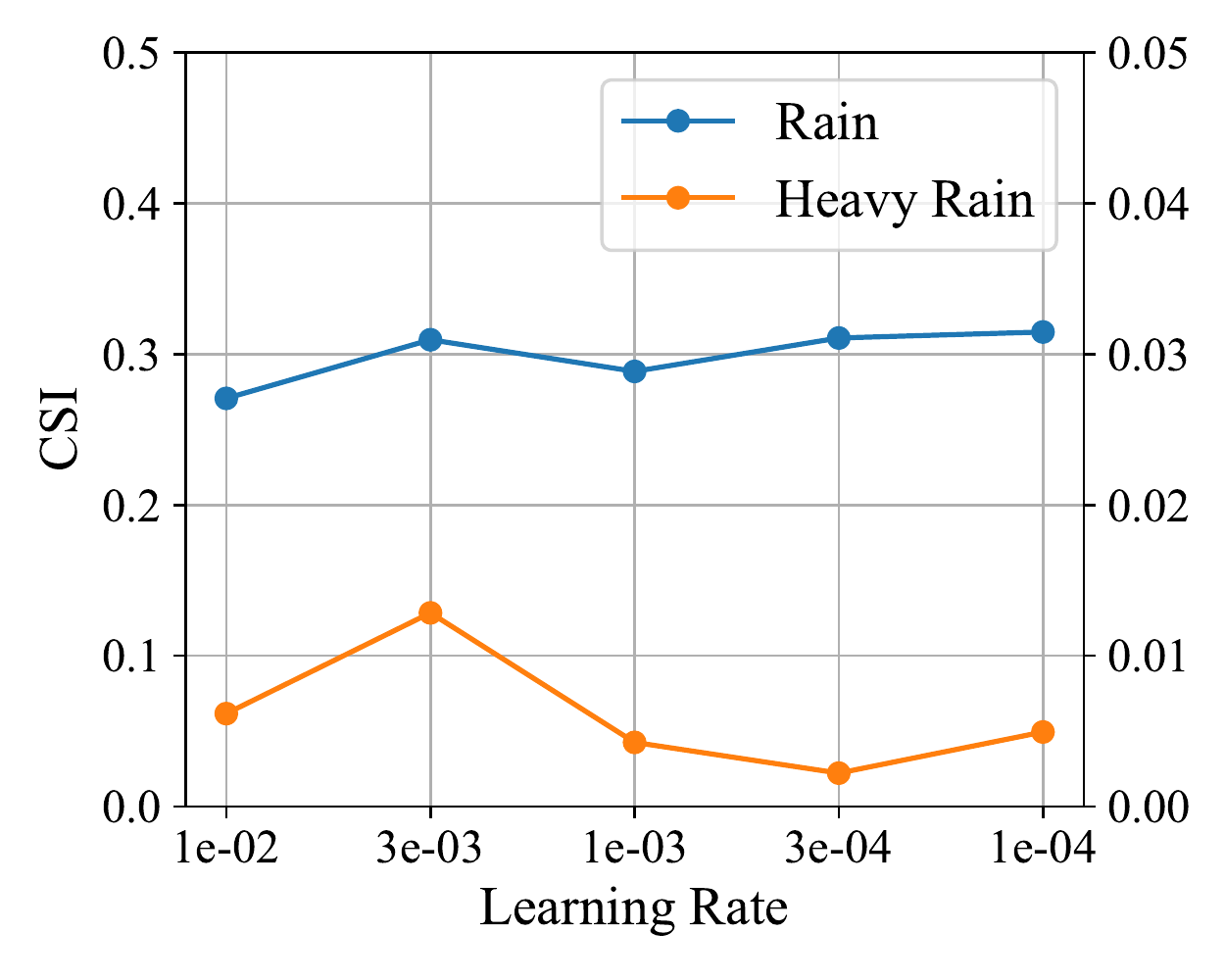}
         \caption{ConvLSTM}
         \label{fig:learning_rate_ablation_convlstm}
     \end{subfigure}
   \begin{subfigure}[b]{0.32\textwidth}
         \centering
         \includegraphics[width=\textwidth]{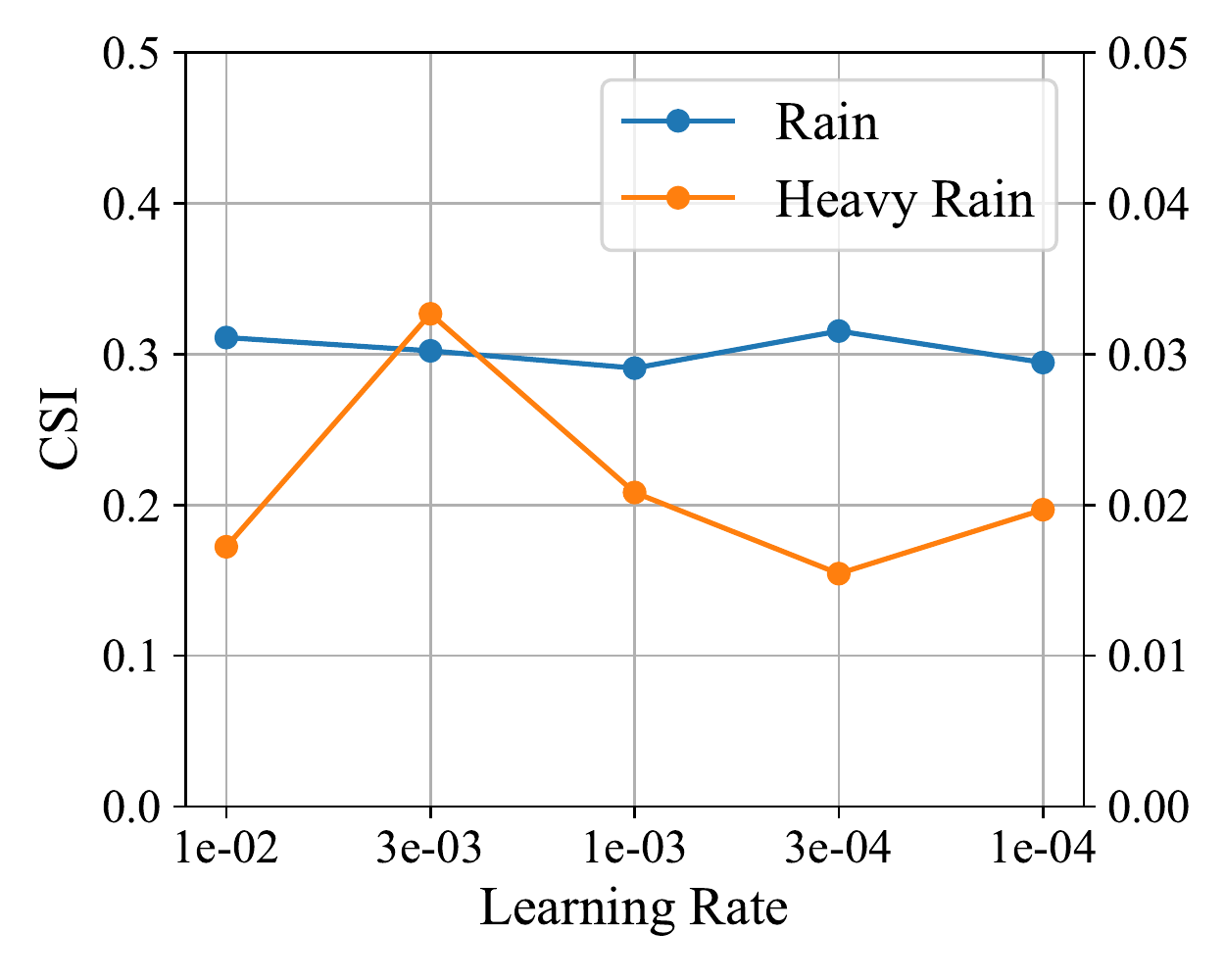}
         \caption{MetNet}
         \label{fig:learning_rate_ablation_metnet}
     \end{subfigure}
\end{center}
\caption{CSI performance according to changes in learning rate.}
\label{learning_rate_ablation}
\end{figure}

\autoref{learning_rate_ablation} shows the CSI scores according to changes in learning rate. When comparing the performance of `rain' and `heavy rain', we see different trends based on increase in learning rate.

\subsection{Learning Curve}
\begin{figure}[h]
\begin{center}
     \begin{subfigure}[b]{0.32\textwidth}
         \centering
         \includegraphics[width=\textwidth]{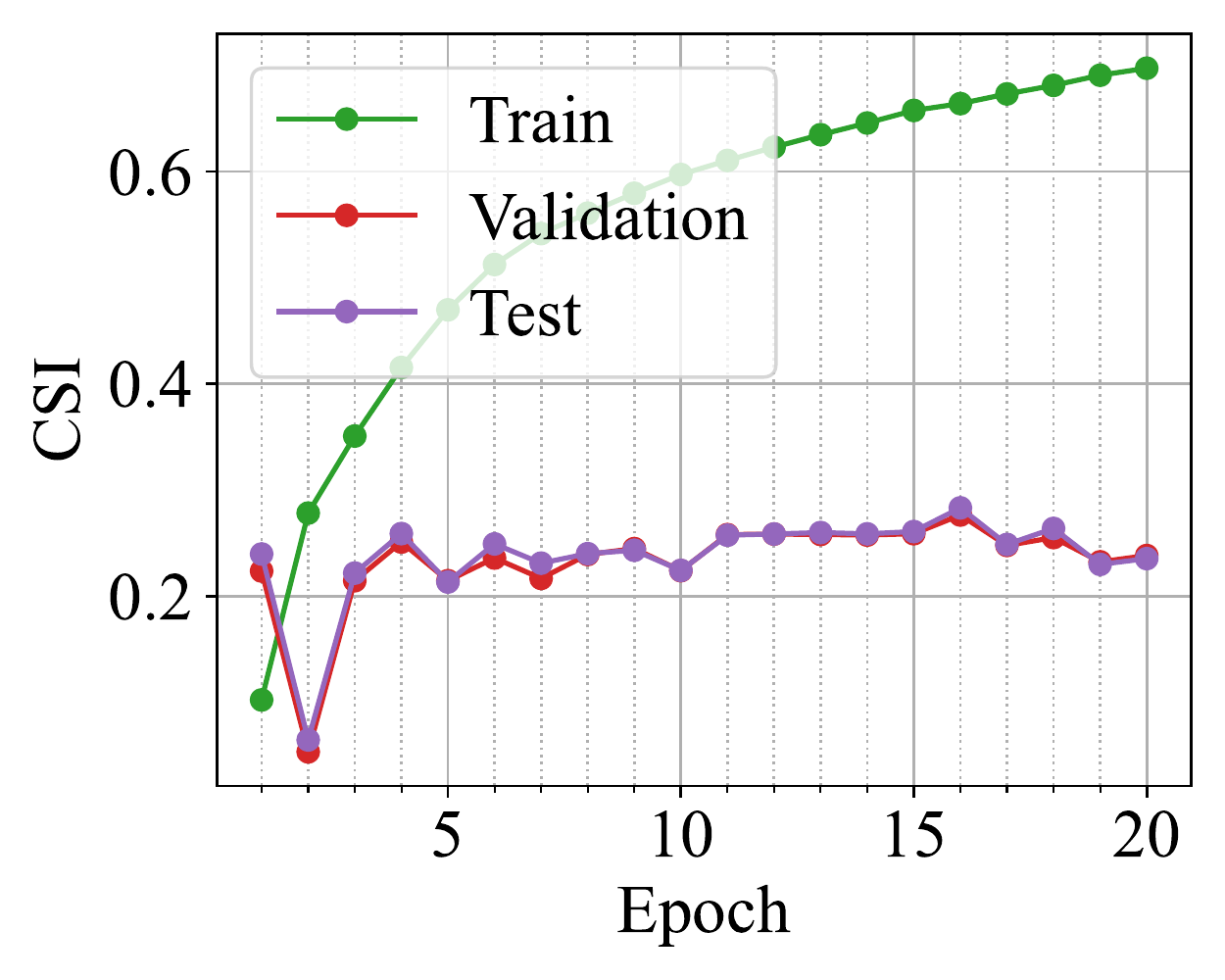}
         \caption{U-Net, window size=1}
         \label{fig:ws_1_unet}
     \end{subfigure}
  \begin{subfigure}[b]{0.32\textwidth}
         \centering
         \includegraphics[width=\textwidth]{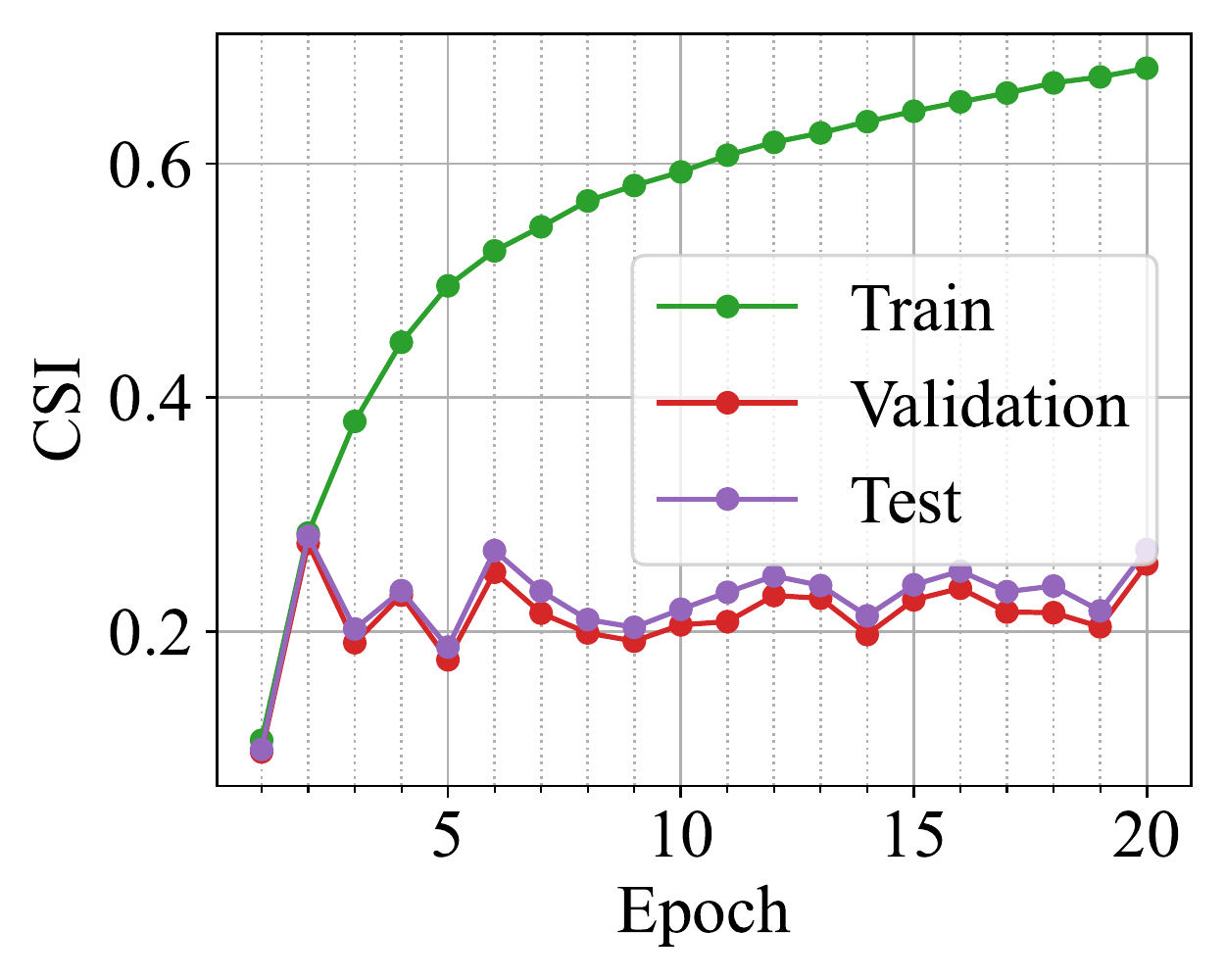}
         \caption{U-Net, window size=3}
         \label{fig:ws_3_unet}
     \end{subfigure}
  \begin{subfigure}[b]{0.32\textwidth}
         \centering
         \includegraphics[width=\textwidth]{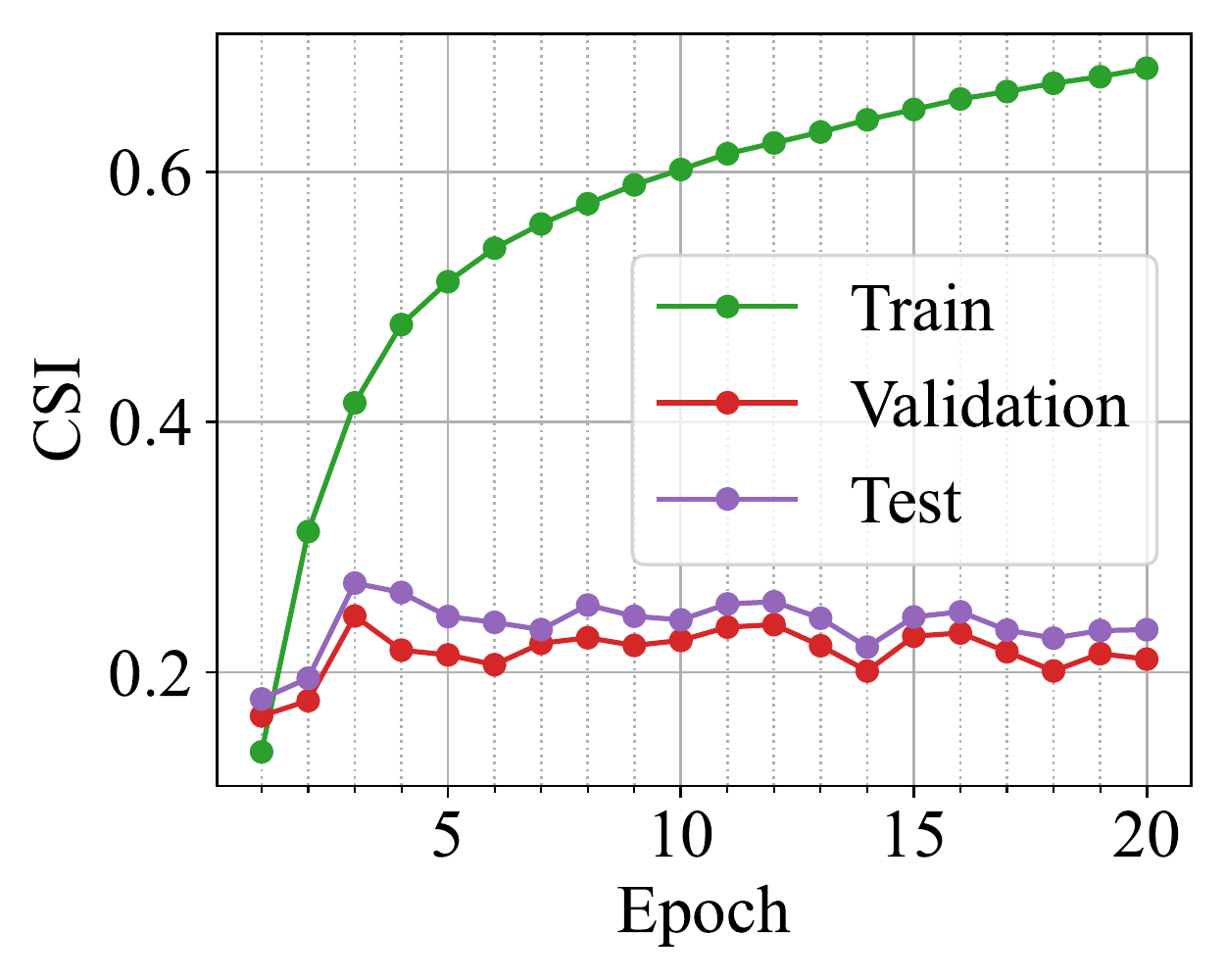}
         \caption{U-Net, window size=6}
         \label{fig:ws_6_unet}
     \end{subfigure}
\end{center}
\caption{U-Net: CSI learning curve according to window size.}
\label{unet_window_size}
\end{figure}

\begin{figure}[h]
\begin{center}
     \begin{subfigure}[b]{0.32\textwidth}
         \centering
         \includegraphics[width=\textwidth]{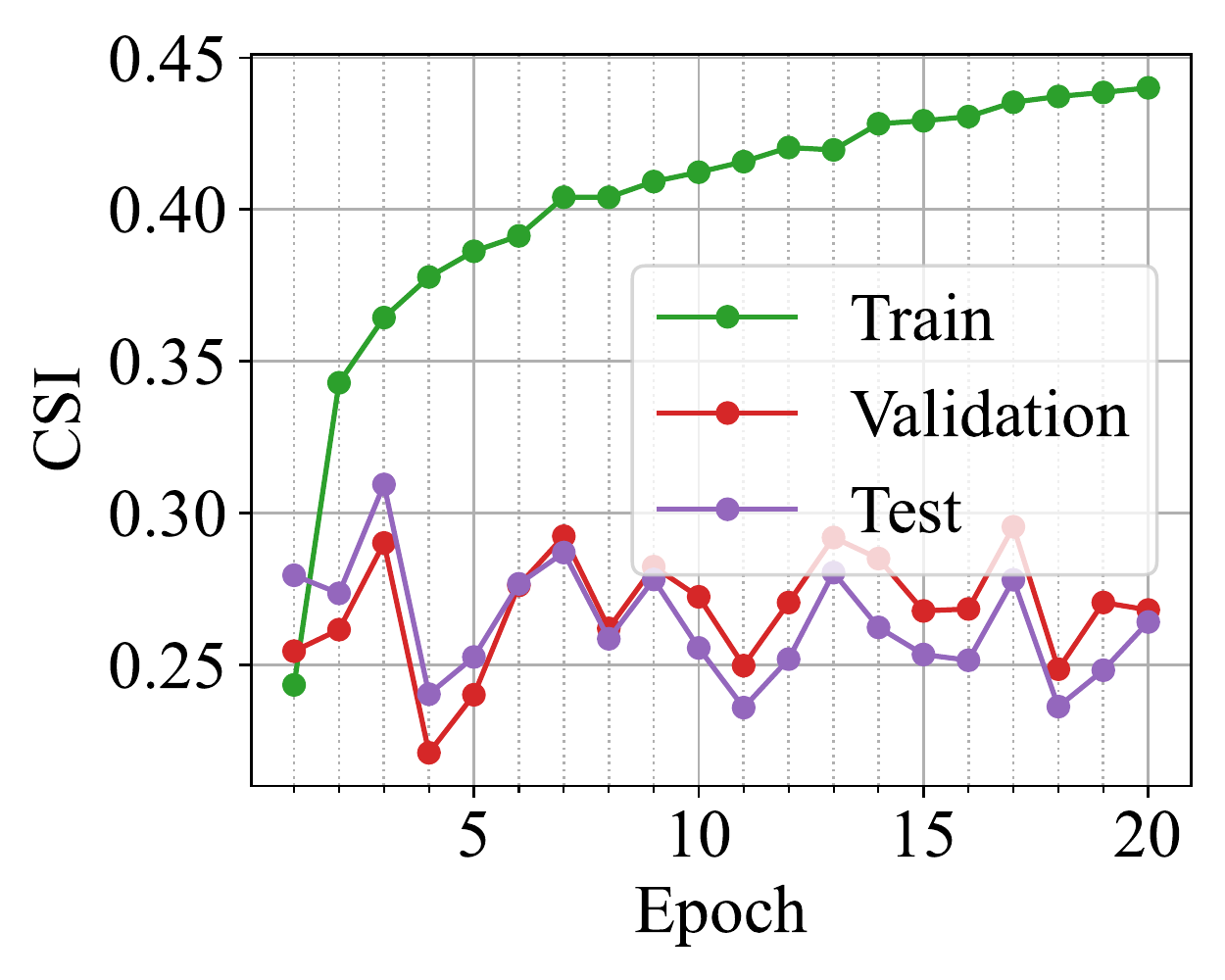}
         \caption{Conv-LSTM, window size=1}
         \label{fig:ws_1_convlstm}
     \end{subfigure}
  \begin{subfigure}[b]{0.32\textwidth}
         \centering
         \includegraphics[width=\textwidth]{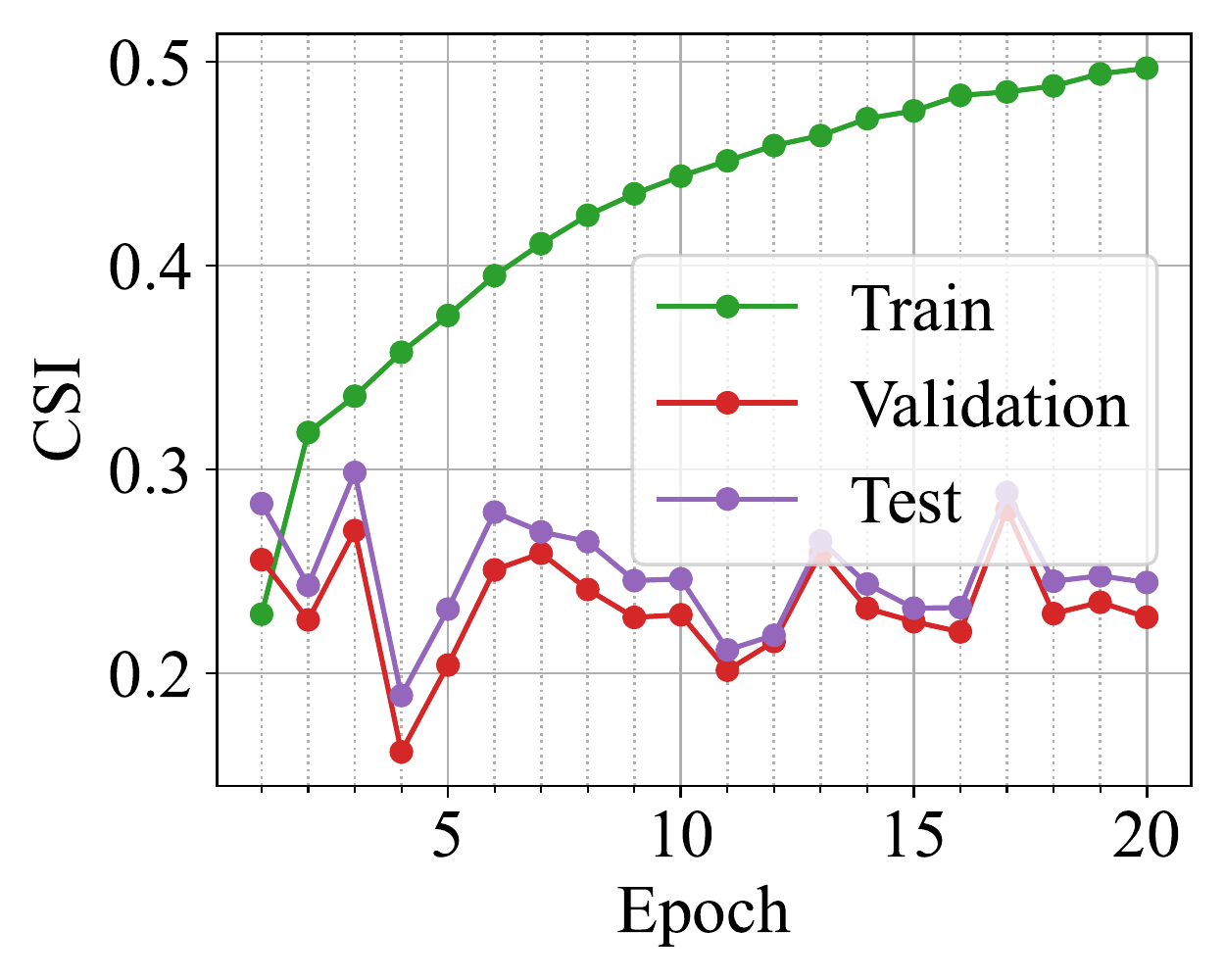}
         \caption{Conv-LSTM, window size=3}
         \label{fig:ws_3_convlstm}
     \end{subfigure}
  \begin{subfigure}[b]{0.32\textwidth}
         \centering
         \includegraphics[width=\textwidth]{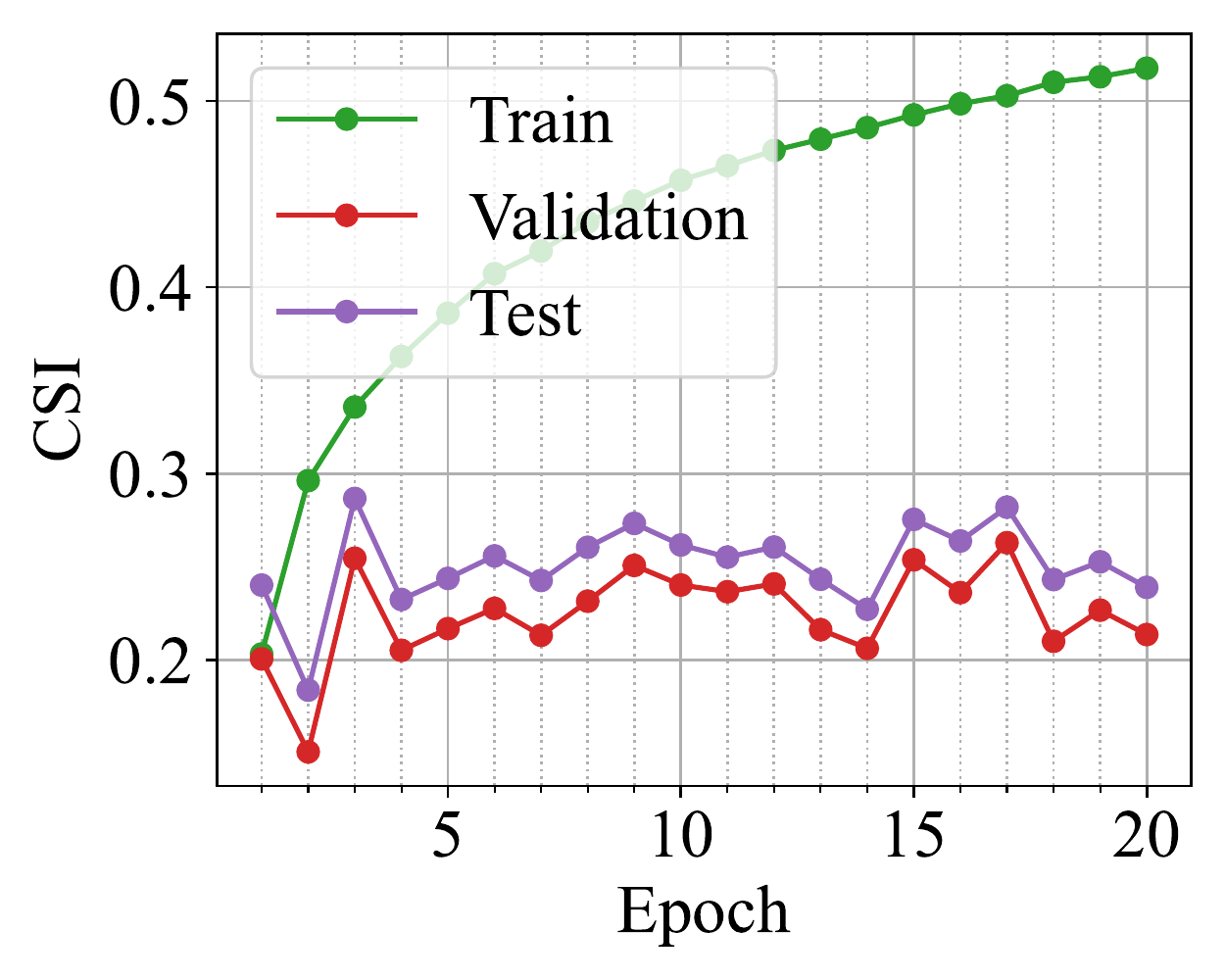}
         \caption{Conv-LSTM, window size=6}
         \label{fig:ws_6_convlstm}
     \end{subfigure}
\end{center}
\caption{Conv-LSTM: CSI learning curve according to window size.}
\label{convlstm_window_size}
\end{figure}

\begin{figure}[h]
\begin{center}
     \begin{subfigure}[b]{0.32\textwidth}
         \centering
         \includegraphics[width=\textwidth]{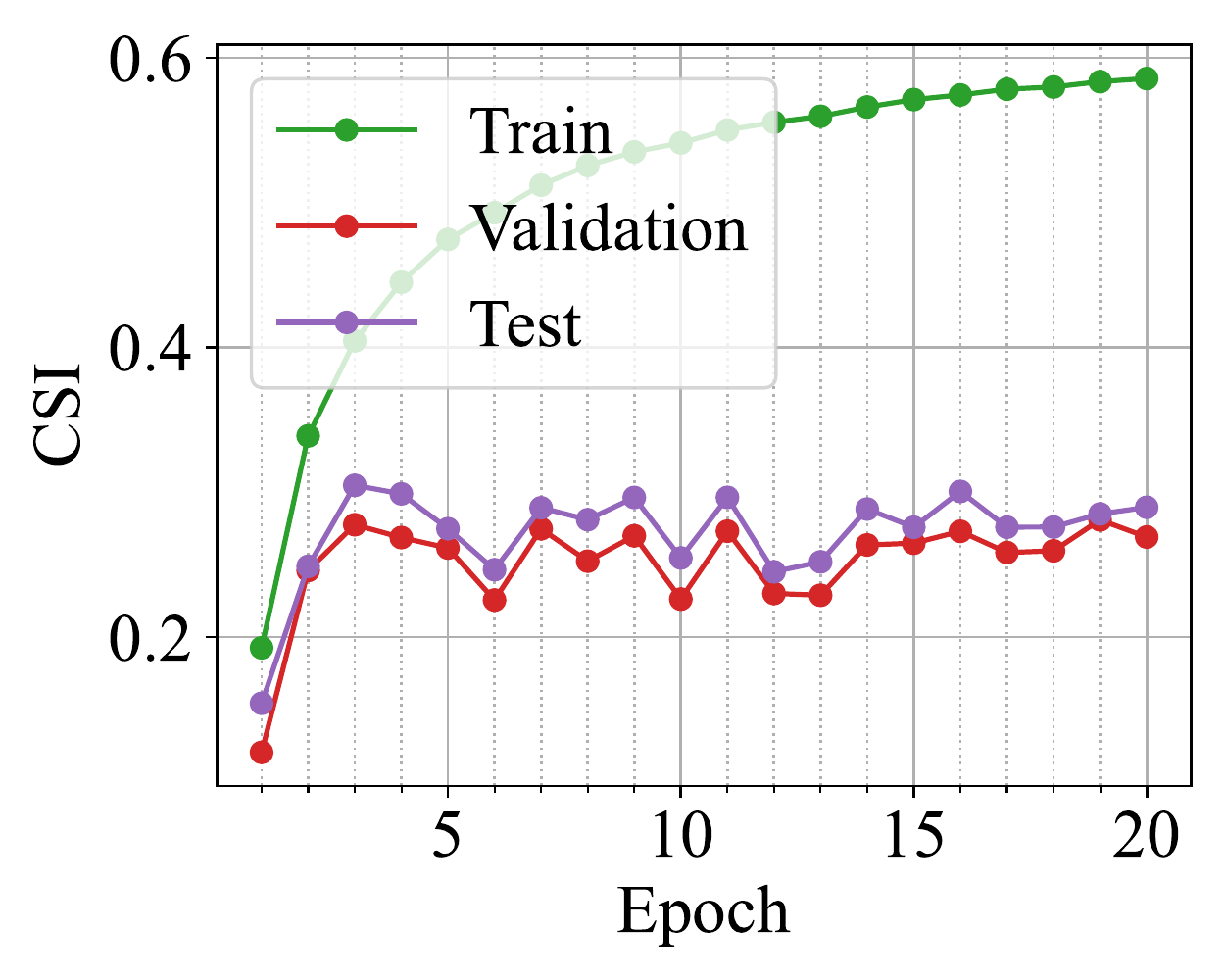}
         \caption{MetNet, window size=1}
         \label{fig:ws_1_metnet}
     \end{subfigure}
  \begin{subfigure}[b]{0.32\textwidth}
         \centering
         \includegraphics[width=\textwidth]{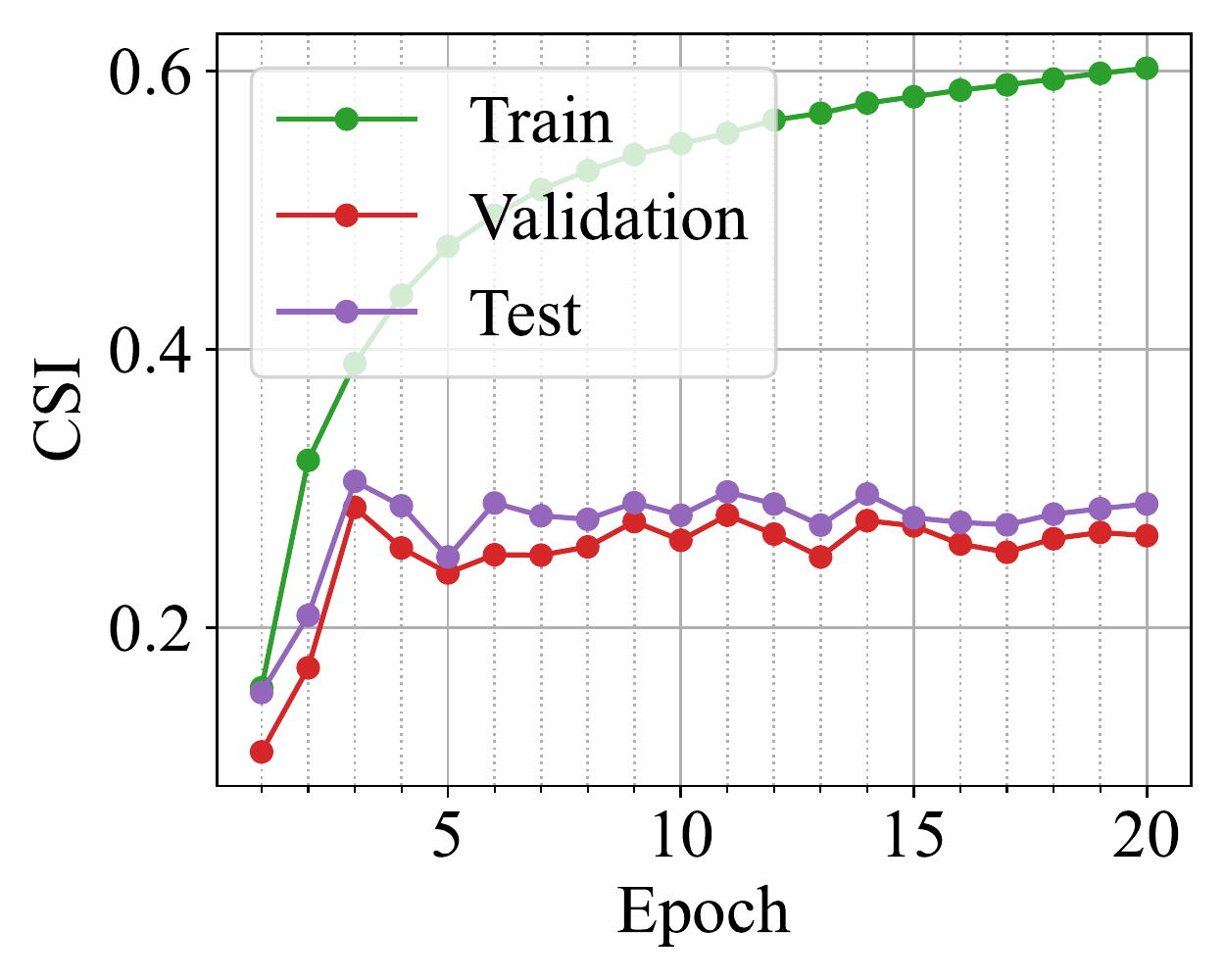}
         \caption{MetNet, window size=3}
         \label{fig:ws_3_metnet}
     \end{subfigure}
  \begin{subfigure}[b]{0.32\textwidth}
         \centering
         \includegraphics[width=\textwidth]{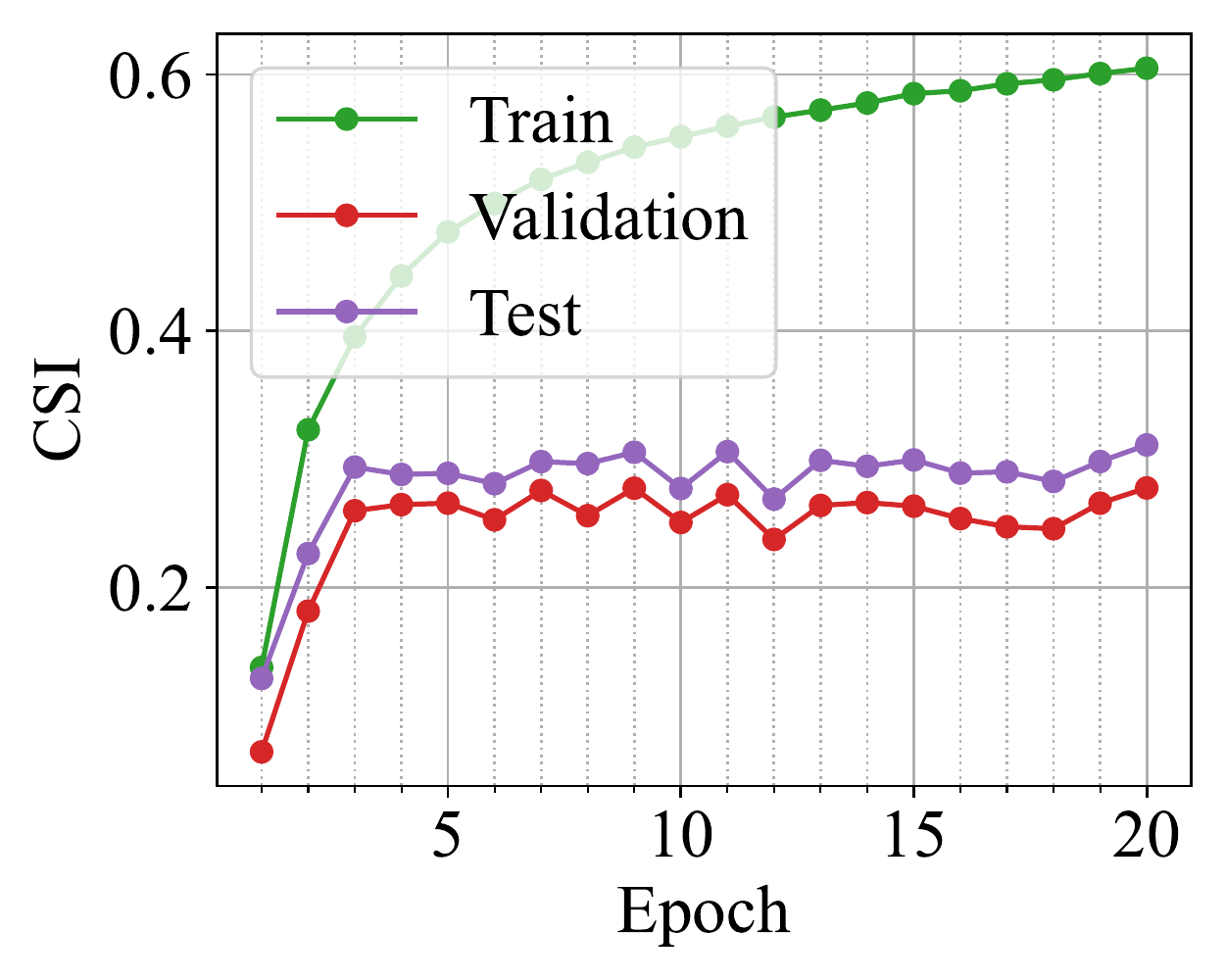}
         \caption{MetNet, window size=6}
         \label{fig:ws_6_metnet}
     \end{subfigure}
\end{center}
\caption{MetNet: CSI learning curve according to window size.}
\label{metnet_window_size}
\end{figure}

We visualize the learning curves of the networks according to changes in window size\,(\autoref{unet_window_size}, \autoref{convlstm_window_size}, \autoref{metnet_window_size}).

\clearpage

\subsection{Generalization for Different Times of the Month}

\textcolor{black}{We evaluate the deep models trained on our benchmark for different times of the year, ranging from July 2020 to July 2021. Our benchmark dataset covers only summer for 2020 and 2021, and thus it harbors concerns about the generalization of different seasons. While the data of other months is not provided owing the limit of license, it is available for the analysis of CSI distribution according to the lead time as well as month, as a substitute of raw data.}

\begin{figure}[h]
\begin{center}
   \begin{subfigure}[b]{0.48\textwidth}
         \centering
         \includegraphics[width=\textwidth]{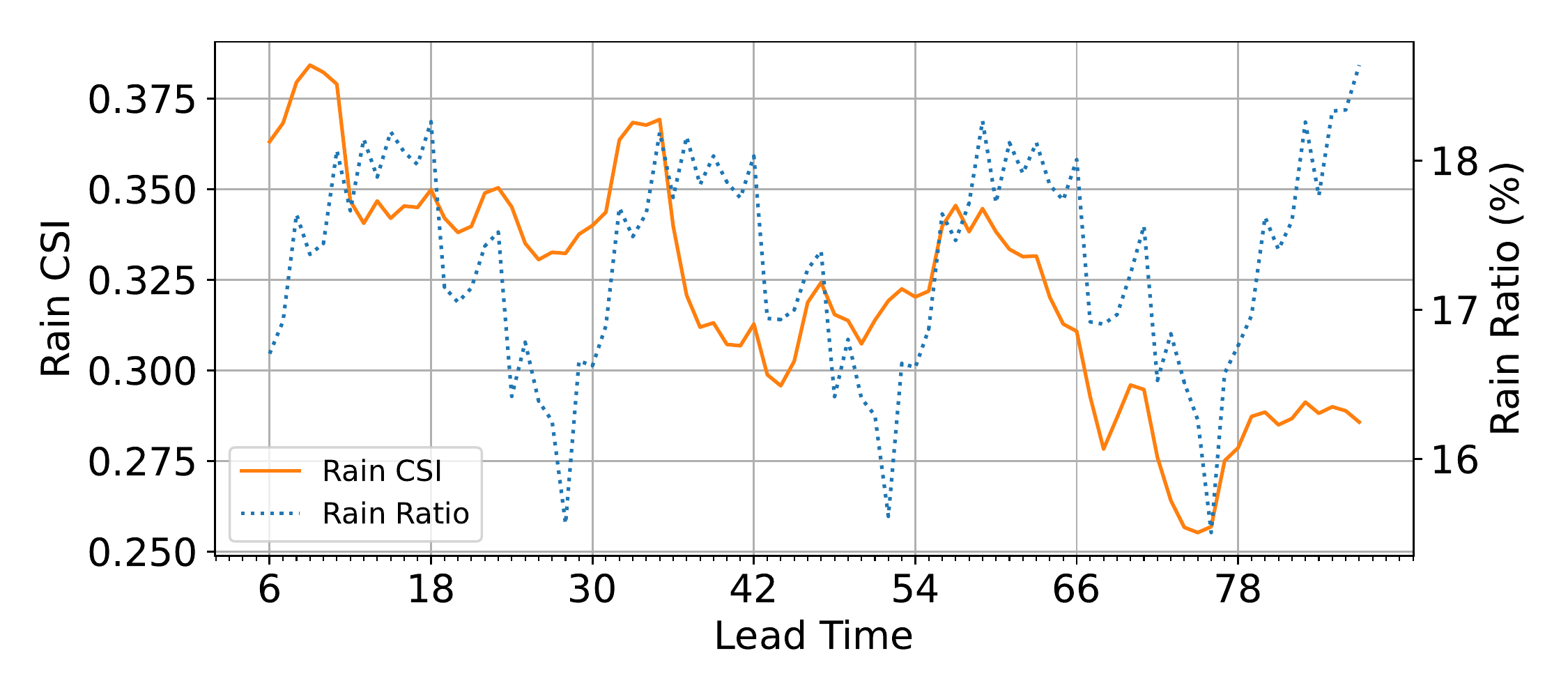}
         \vspace{-20pt}
         \caption{Rain CSI by lead time}
         \label{fig:convlstm_rain_hour}
     \end{subfigure}
   \begin{subfigure}[b]{0.48\textwidth}
         \centering
         \includegraphics[width=\textwidth]{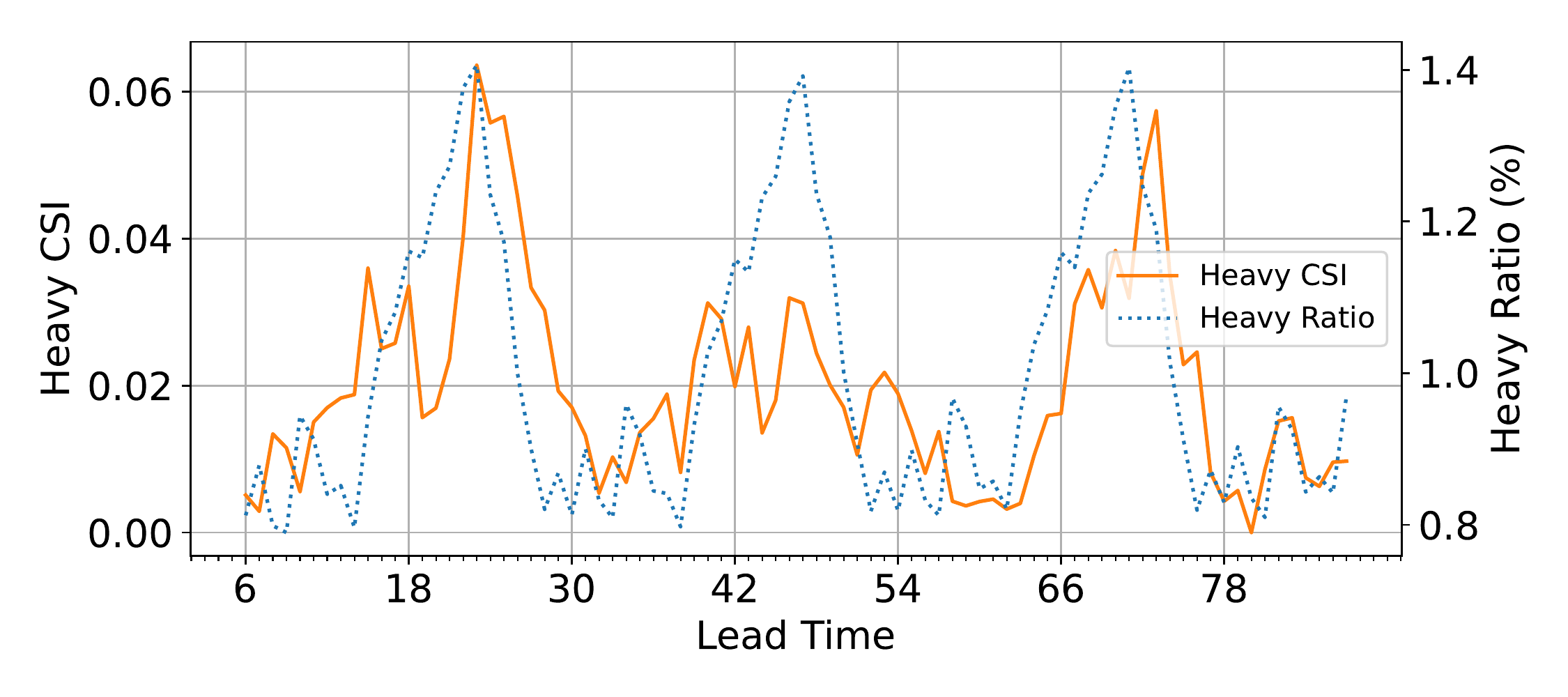}
         \vspace{-20pt}
         \caption{Heavy rain CSI by lead time}
         \label{fig:convlstm_heavy_hour}
     \end{subfigure}
   \begin{subfigure}[b]{0.48\textwidth}
         \centering
         \includegraphics[width=\textwidth]{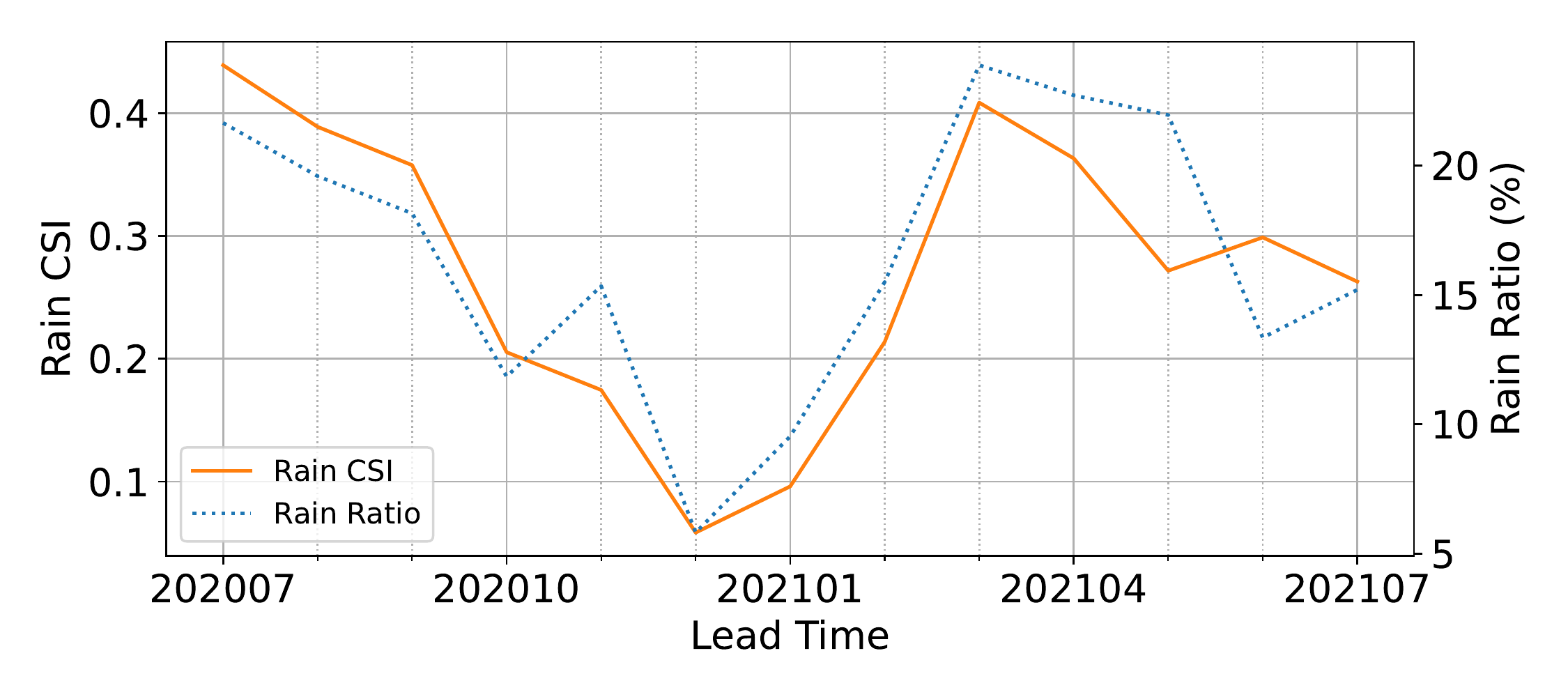}
         \vspace{-20pt}
         \caption{Rain CSI by month}
         \label{fig:convlstm_rain_month}
     \end{subfigure}
   \begin{subfigure}[b]{0.48\textwidth}
         \centering
         \includegraphics[width=\textwidth]{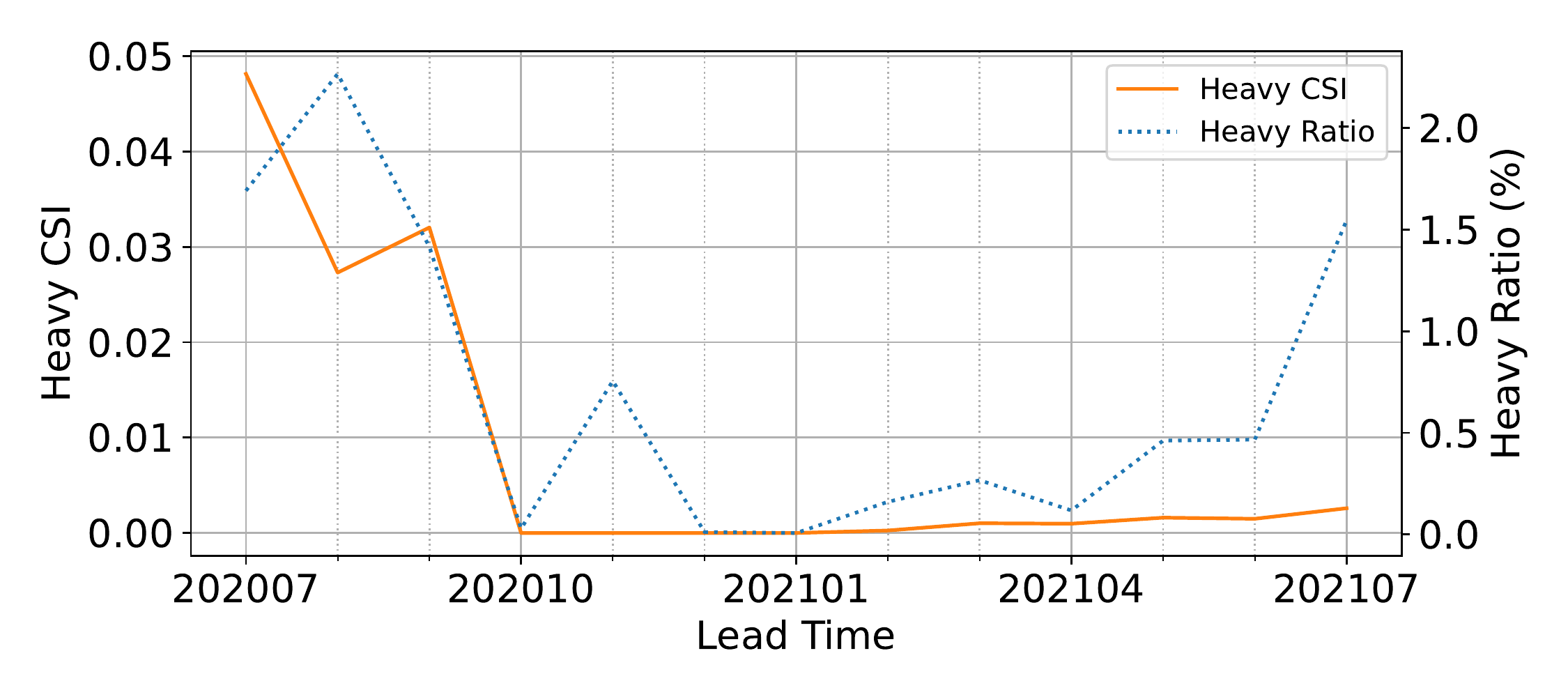}
         \vspace{-20pt}
         \caption{Heavy rain CSI by month}
         \label{fig:convlstm_heavy_month}
     \end{subfigure}
\end{center}
\vspace{-5pt}
\caption{\textcolor{black}{CSI distribution of ConvLSTM outputs from July 2020 to July 2021. Orange line indicates the rain CSI of the model and the blue dot line is the distribution of observed rain ratio from AWS.}}
\label{csi_convlstm}
\vspace{-15pt}
\end{figure}

\begin{figure}[h]
\begin{center}
   \begin{subfigure}[b]{0.48\textwidth}
         \centering
         \includegraphics[width=\textwidth]{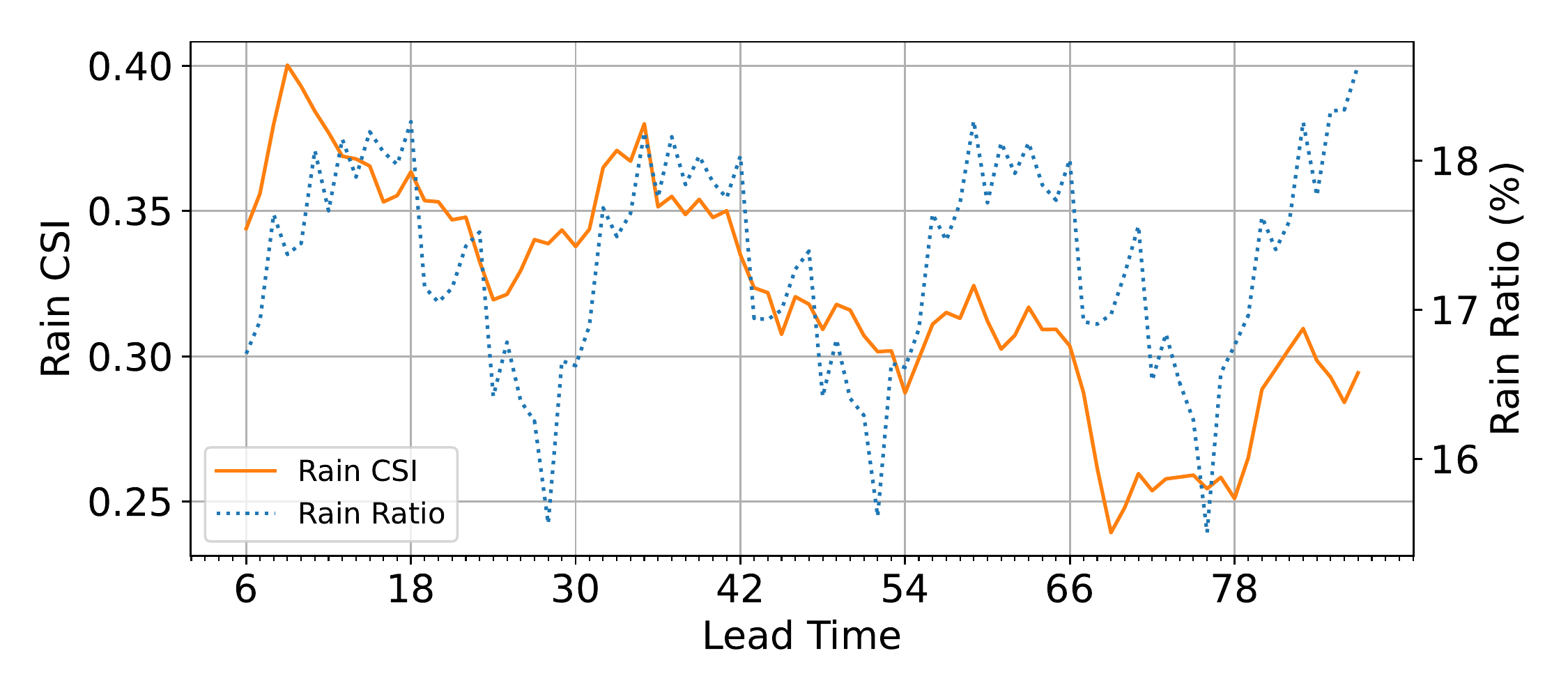}
         \vspace{-20pt}
         \caption{Rain CSI by lead time}
         \label{fig:convlstm_rain_hour}
     \end{subfigure}
   \begin{subfigure}[b]{0.48\textwidth}
         \centering
         \includegraphics[width=\textwidth]{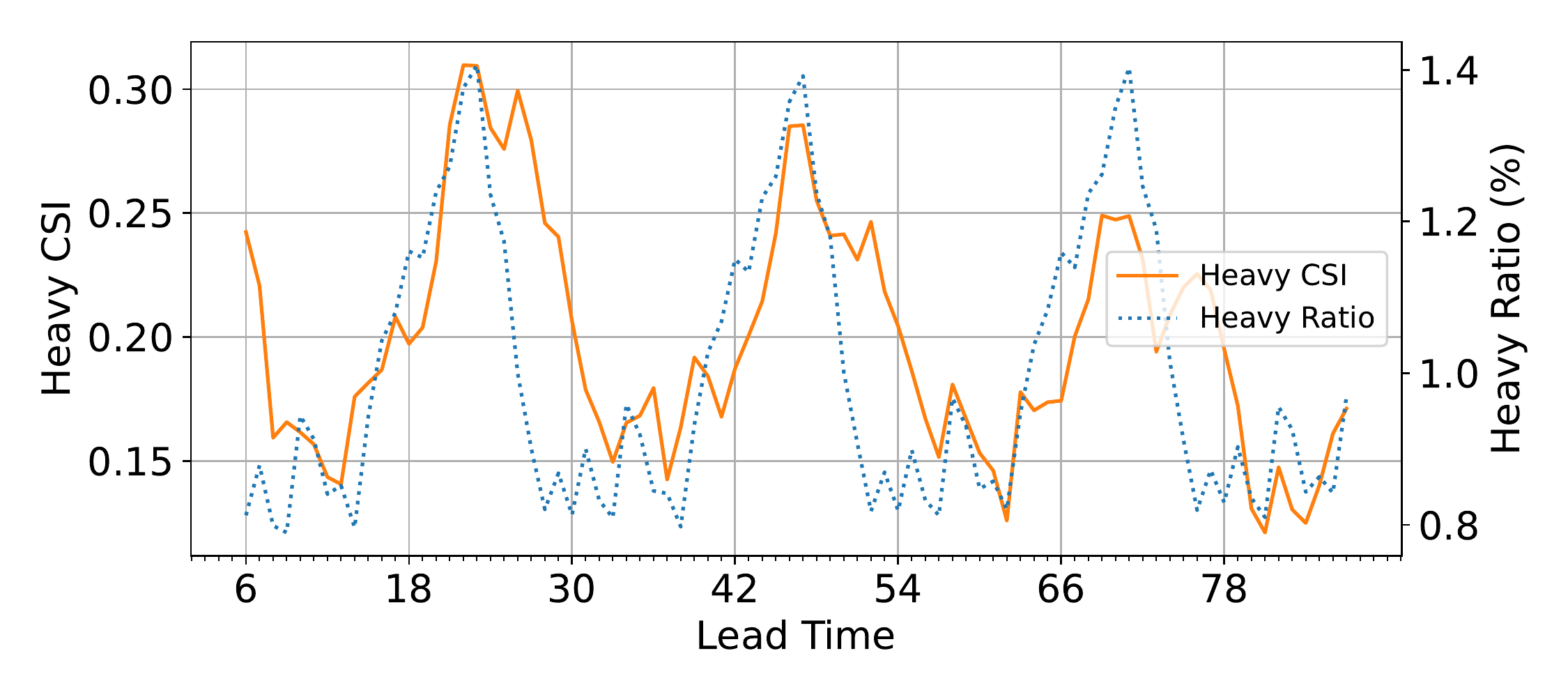}
         \vspace{-20pt}
         \caption{Heavy rain CSI by lead time}
         \label{fig:convlstm_heavy_hour}
     \end{subfigure}
   \begin{subfigure}[b]{0.48\textwidth}
         \centering
         \includegraphics[width=\textwidth]{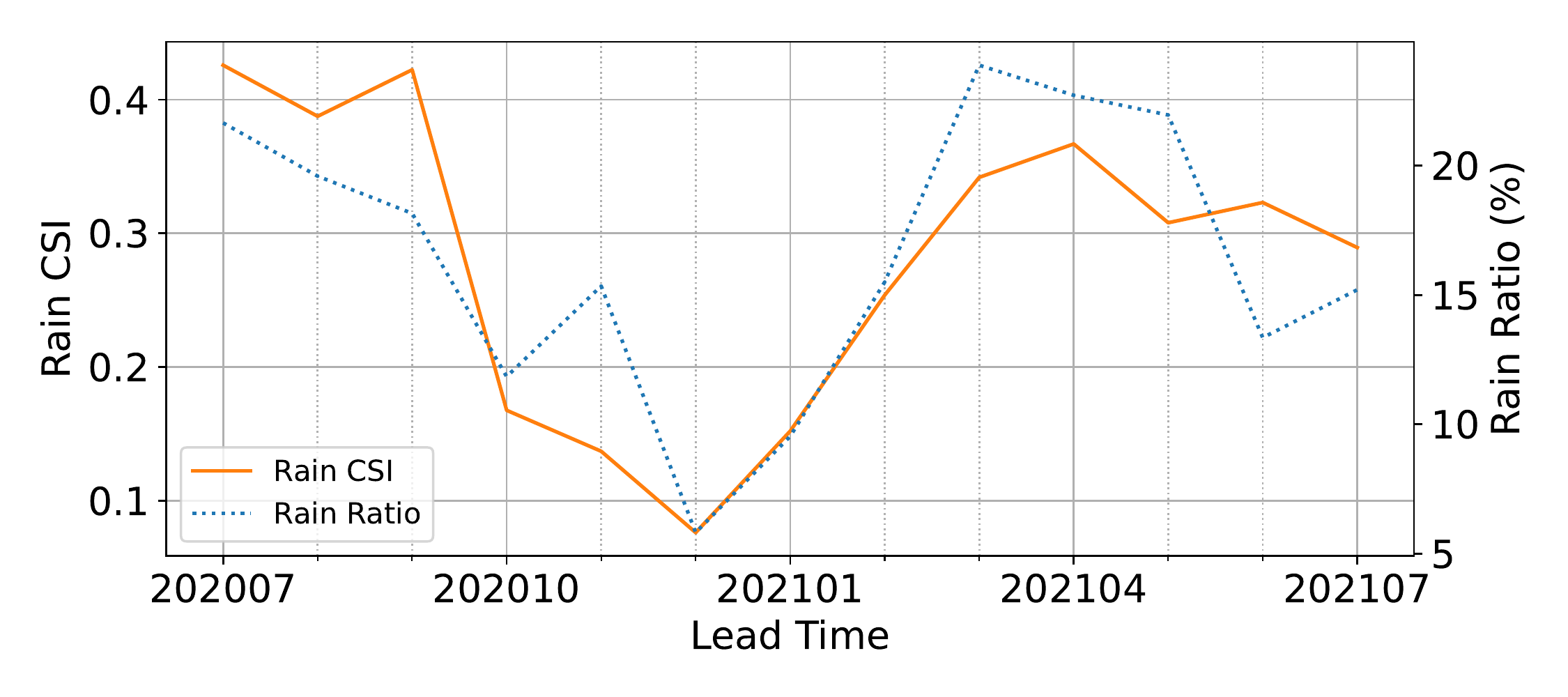}
         \vspace{-20pt}
         \caption{Rain CSI by month}
         \label{fig:convlstm_rain_month}
     \end{subfigure}
   \begin{subfigure}[b]{0.48\textwidth}
         \centering
         \includegraphics[width=\textwidth]{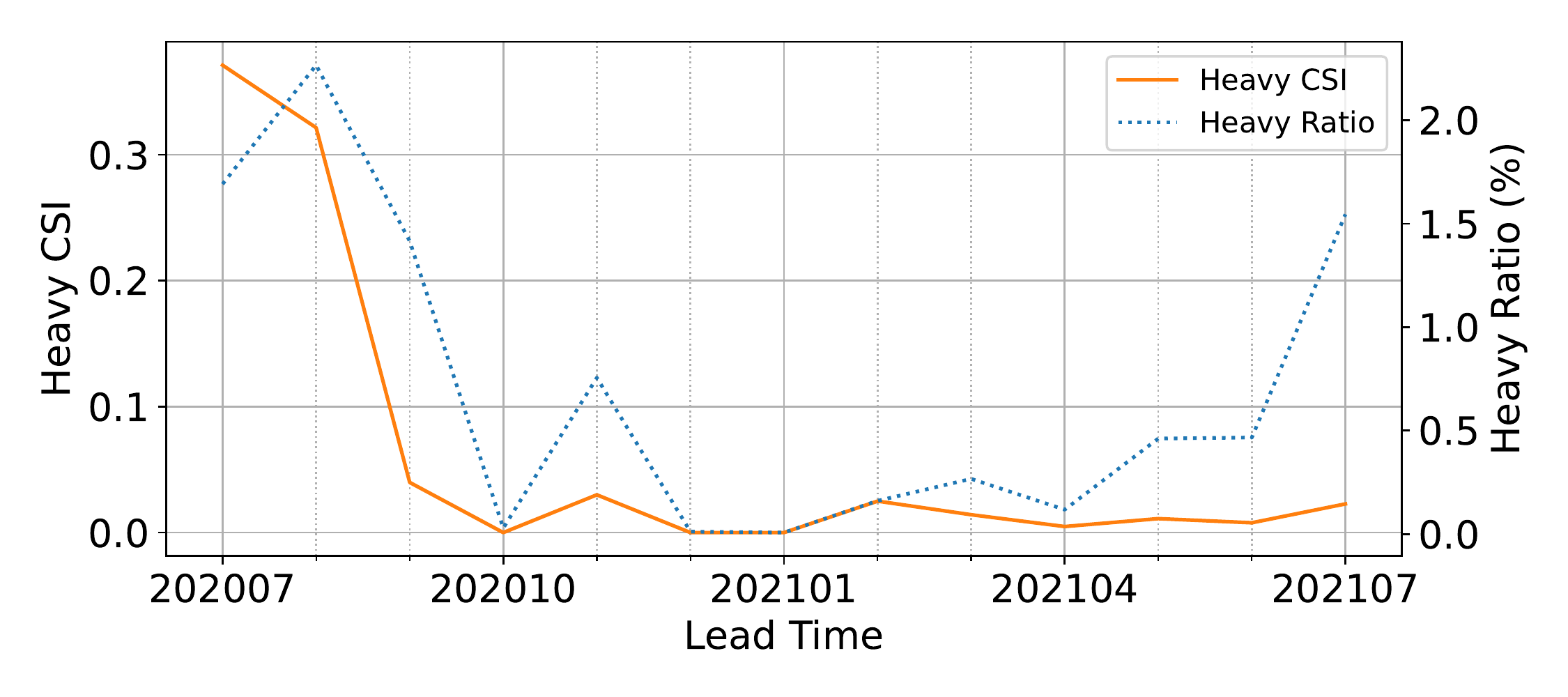}
         \vspace{-20pt}
         \caption{Heavy rain CSI by month}
         \label{fig:convlstm_heavy_month}
     \end{subfigure}
\end{center}
\vspace{-5pt}
\caption{\textcolor{black}{CSI distribution of U-Net outputs from July 2020 to July 2021. Orange line indicates the rain CSI of the model and the blue dot line is the distribution of observed rain ratio from AWS.}}
\label{csi_unet}
\vspace{-15pt}
\end{figure}

\textcolor{black}{\autoref{csi_convlstm} and \autoref{csi_unet} show the CSI distribution by lead time and month from ConvLSTM and U-Net, respectively. For the lead time, both models has larger CSI values of early lead time than those of latter and the overall curve has inflection points at similar time compared to the curve of rain ratio. On the other hand, the curve of heavy rain CSI is significantly similar with the graph of heavy rain ratio distribution on lead time. For the month, it can be seen that the annual average rain CSI is near 0.3, but the performance is extremely low in winter in both models. This seems to be the case, however, because the amount of rain in winter is close to zero. This phenomenon is more pronounced in heavy rain.}

\clearpage

\subsection{Training on Other Multiple Categorical Classes}
\textcolor{black}{Our framework allows users to specify categorical classes in training as they wish if 0.1 and 10 are included in the threshold list. By referring to the other regions\,\cite{ravuri2021skilful, espeholt2021skillful, shi2017deep}, we conduct an experiments with different categorical classes while the evaluation is verified with the standard threshold for rain and heavy rain used in our framework. Followings are the threshold lists with the number of classes:
\begin{itemize}
    \item V1: [0.1, 2.0, 10.0], 4 classes
    \item V2: [0.1, 5.0, 10.0], 4 classes
    \item V3: [0.1, 0.5, 5.0, 10.0], 5 classes
    \item V4: [0.1, 0.5, 5.0, 10.0, 30.0], 6 classes
\end{itemize}}

\textcolor{black}{\autoref{tab:multiple_classes} shows the performance according to the changes of the threshold list. As the class of middle rain is newly created, it can be seen that the CSI can be changed. Furthermore, when the number of classes in the middle rain is further increased, it can be seen that the csi of the heavy rain increases significantly with some degradation in rain CSI.}



\begin{table}[h]
\caption{\textcolor{black}{Evaluation metrics of KoMet with the MetNet for precipitation while 12 variables are utilized for the training, according to the changes of the number of categorical classes during training.}}
\centering
\begin{tabular}{@{}c|c|c|c|c|c|c@{}}
\toprule
   & \multicolumn{3}{c|}{Rain} & \multicolumn{3}{c}{Heavy Rain} \\ \midrule
   & Acc    & CSI    & Bias   & Acc      & CSI      & Bias     \\ \midrule
V1 & 0.862  & 0.293  & 0.769  & 0.986    & 0.011    & 0.079    \\
V2 & 0.858  & 0.322  & 0.946  & 0.986    & 0.016    & 0.097    \\
V3 & 0.871  & 0.270  & 0.581  & 0.987    & 0.048    & 0.210    \\ 
V4 & 0.862  & 0.274  & 0.705  & 0.984    & 0.032    & 0.286    \\ \bottomrule
\end{tabular}
\label{tab:multiple_classes}
\end{table}

\subsection{Pointwise Architecture without the Use of Spatial Information}

\textcolor{black}{We conduct an experiment with methods that do not use spatial information for NWP correction. While our benchmark architectures include both spatial encoder and temporal encoder, it is important to check the needs of spatial neural processing. \autoref{tab:point_wise} shows the results for various lead times from 6 to 87 hours, according to the presence of spatial kernel operator in neural networks. Here, we design pointwise architecture having no spatial kernel. Pointwise architecture with LSTM indicates the modified version of ConvLSTM whose spatial encoder is replaced with pointwise architecture. For the NWP correction, as \autoref{tab:point_wise} shows, using only pointwise operator lead to worse optima than our benchmark models. In other words, it can be seen that the methodology considering the spatial information is more effective.}

\begin{table}[h]
\centering
\caption{\textcolor{black}{Evaluation metrics of KoMet and pointwise architectures for precipitation while 12 variables are utilized for the training.}}
\begin{tabular}{@{}c|c|c|c|c|c|c@{}}
\toprule
                              & \multicolumn{3}{c|}{Rain} & \multicolumn{3}{c}{Heavy Rain} \\ \midrule
                              & Acc    & CSI    & Bias   & Acc      & CSI      & Bias     \\ \midrule
Pointwise Architecture        & 0.871  & 0.195  & 0.346  & 0.987    & 0.001    & 0.009    \\
Pointwise Architecture + LSTM & 0.866  & 0.253  & 0.575  & 0.987    & 0.006    & 0.060    \\
ConvLSTM                      & 0.869  & 0.296  & 0.696  & 0.986    & 0.006    & 0.059    \\ \bottomrule
\end{tabular}
\label{tab:point_wise}
\end{table}

\end{document}